\pdfoutput=1
\documentclass{article}

    \usepackage[final,nonatbib]{neurips_2022}

\usepackage[utf8]{inputenc} 
\usepackage[T1]{fontenc}    
\usepackage{hyperref}       
\usepackage{url}            
\usepackage{booktabs}       
\usepackage{amsfonts}       
\usepackage{nicefrac}       
\usepackage{microtype}      
\usepackage{xcolor}         
\usepackage{soul}
\usepackage{amsmath}
\usepackage{amssymb}
\usepackage{amsthm}
\usepackage{cite}
\usepackage{nicefrac}       
\usepackage{microtype}      
\usepackage{lipsum}
\usepackage{graphicx}
\usepackage{subfigure}
\graphicspath{ {./images/} }
\usepackage{bbm}
\usepackage{enumitem}
\usepackage{bm}
\usepackage[absolute,overlay]{textpos}
\newtheorem{theorem}{Theorem}

\newtheorem{remark}{Remark}

\newtheorem{lemmaproofx}{Lemma Proof}

\newtheorem{lemmaproofC}{Lemma Proof}

\newtheorem{lemmaproofF}{Lemma Proof}

\newtheorem{lemmaproofI}{Lemma Proof}

\newtheorem{exmp}{Example}
\newtheorem{assumption}{Assumption}
\newtheorem{lemma}{Lemma}

\newtheorem{proposition}{Proposition}
\newtheorem{lemmax}{Lemma}

\newtheorem{lemmaC}{Lemma}

\newtheorem{lemmaF}{Lemma}

\newtheorem{lemmaI}{Lemma}

\newtheorem{propositionC}{Proposition}

\newcommand{\RR}{\mathbb{R}}

\DeclareMathOperator*{\argmax}{argmax}
\usepackage{mathtools}
\usepackage{graphicx,wrapfig}

\title{Understanding Non-linearity in Graph Neural Networks from the Perspective of Bayesian Inference}

\author{
  Rongzhe Wei$^{1}$, Haoteng Yin$^{1}$, Junteng Jia$^{2}$, Austin R. Benson$^{2}$, Pan Li$^{1}$
  \\
  $^1$ Department of Computer Science, Purdue University \\
  $^2$ Department of Computer Science, Cornell University\\
  \texttt{\{wei397, yinht, panli\}@purdue.edu, jj585@cornell.edu, arbenson@gmail.com}
}

\begin{document}

\maketitle

\vspace{-3mm}
\begin{abstract}
Graph neural networks (GNNs) have shown superiority in many prediction tasks over graphs due to their impressive capability of capturing nonlinear relations in graph-structured data. However, for node classification tasks, often, only marginal improvement of GNNs over their linear counterparts has been observed. Previous works provide very few understandings of this phenomenon. In this work, we resort to Bayesian learning to deeply investigate the functions of non-linearity in GNNs for node classification tasks. Given a graph generated from the statistical model CSBM, we observe that the max-a-posterior estimation of a node label given its own and neighbors' attributes consists of two types of non-linearity, a possibly non-linear transformation of node attributes and a ReLU-activated feature aggregation from neighbors. The latter surprisingly matches the type of non-linearity used in many GNN models. By further imposing a Gaussian assumption on node attributes, we prove that the superiority of those ReLU activations is only significant when the node attributes are far more informative than the graph structure, which nicely matches many previous empirical observations. A similar argument can be achieved when there is a distribution shift of node attributes between the training and testing datasets. Finally, we verify our theory on both synthetic and real-world networks. Our code is available at \url{https://github.com/Graph-COM/Bayesian_inference_based_GNN.git}.
\end{abstract}

\vspace{-3mm}
\section{Introduction}  
\label{sec:into}\vspace{-1mm}
Learning on graphs (LoG) has been widely used in the applications with graph-structured data~\cite{zhu2005semi,GRLbook}. Node classification, as one of the most crucial tasks in LoG, asks to predict the labels of nodes in a graph, which has been used in many applications such as community detection~\cite{fortunato2010community,lancichinetti2009community,chen2020supervised,liu2021deep}, anomaly detection~\cite{ma2021comprehensive,wang2021bipartite},  biological pathway analysis~\cite{aittokallio2006graph,scott2006efficient} and so on. 

Recently, graph neural networks (GNNs) have become the de-facto standard used in many LoG tasks due to their super empirical performance~\cite{hamilton2017inductive,kipf2016semi}. Researchers often attribute such success to  non-linearity in GNNs which associates them with great expressive power~\cite{xu2018powerful,morris2019weisfeiler}. GNNs can approximate a wide range of functions defined over graphs~\cite{keriven2019universal,chen2019equivalence,azizian2021expressive} and thus excel in predicting, e.g., the free energies of molecules~\cite{gilmer2017neural}, which are by nature non-linear solutions of some quantum-mechanical equations. However, for node classification tasks, many studies have shown that non-linearity to control the exchange of features among neighbors seems not that crucial. For example, many works use linear propagation of node attributes over graphs~\cite{wu2019simplifying,he2020lightgcn}, and others recommend adding non-linearity while only to the transformation of initial node attributes~\cite{klicpera2018predict,DBLP:conf/nips/KlicperaWG19,huang2020combining}. Both cases achieve comparable or even better performance than other models with complex nonlinear propagation, such as using neighbor-attention mechanism~\cite{velivckovic2018graph}. Recently, even in the complicated heterophilic setting where nodes with same labels are not directly connected, linear propagation still achieves competitive performance~\cite{chien2021adaptive,wang2022powerful}, compared with the models with nonlinear and deep architectures~\cite{zhu2020beyond,zhu2021graph}. 

Although empirical studies on GNNs are extensive till now and many practical observations as above have been made, there have been very few works attempting to characterize GNNs in theory, especially to understand the effect of non-linearity by comparing with the linear counterparts for node classification tasks. The only work on this topic to the best of our knowledge still focuses on comparing the expressive power of the two methods to distinguish nodes with different local structures~\cite{chen2020graph}. However, the achieved statement that non-linear propagation improves expressiveness may not necessarily reveal the above phenomenon that non-linear and linear methods have close empirical performance while with subtle difference. Moreover, more expressiveness is often at the cost of model generalization and thus may not necessarily yield more accurate prediction~\cite{cong2021provable,li2022expressive}.

In this work, we expect to give a more precise characterization of the values of  non-linearity in GNNs from a statistical perspective, based on Bayesian inference specifically. We resort to contextual stochastic block models (CSBM)~\cite{binkiewicz2017covariate,deshpande2018contextual}. We make a significant observation that given a graph generated by CSBM, the max-a-posterior (MAP) estimation of a node label given its own and neighbors' features surprisingly corresponds to a graph convolution layer with \emph{ReLU} as the activation combined with an initial node-attribute transformation. Such a transformation of node attributes is generally nonlinear unless they are generated from the natural exponential family~\cite{morris1982natural}. Since the MAP estimator is known to be Bayesian optimal~\cite{theodoridis2015machine}, the above observation means that ReLU-based propagation has the potential to outperform linear propagation. To precisely characterize such benefit, we further assume that the node attributes are generated from a label-conditioned Gaussian model, and analyze and compare the node mis-classification errors of linear and nonlinear models. We have achieved the following conclusions (note that we only provide informal statements here and the formal statements are left in the theorems).\vspace{-1.5mm}
\begin{itemize}[leftmargin=3mm]
    \item When the node attributes are less informative compared to the structural information, non-linear propagation and linear propagation have almost the same mis-classification error (case I in Thm.~\ref{thm:2}).\vspace{-0.5mm}
    \item When the node attributes are more informative, non-linear propagation shows advantages. The mis-classification error of non-linear propagation can be  significantly smaller than that of linear propagation with sufficiently informative node attributes (case II in Thm.~\ref{thm:2}).  \vspace{-0.5mm}
    \item  When there is a distribution shift of the node attributes between the training and testing datasets, non-linearity provides better transferability in the regime of informative node attributes (Thm.~\ref{thm:3}).   \vspace{-1.5mm}
\end{itemize}
 Given that practical node attributes are often not that informative, the advantages of non-linear propagation over linear propagation for node classification is limited albeit observable. Our analysis and conclusion apply to both homophilic and heterophilic settings, i.e., when nodes with same labels tend to be connected (homophily) or disconnected (heterophily), respectively~\cite{chien2021adaptive,zhu2020beyond,zhu2021graph,suresh2021breaking,ma2021homophily}. 
 



\begin{figure}[t]
\label{fig.ova-gaussian}
\centering
\includegraphics[width=0.3\textwidth]{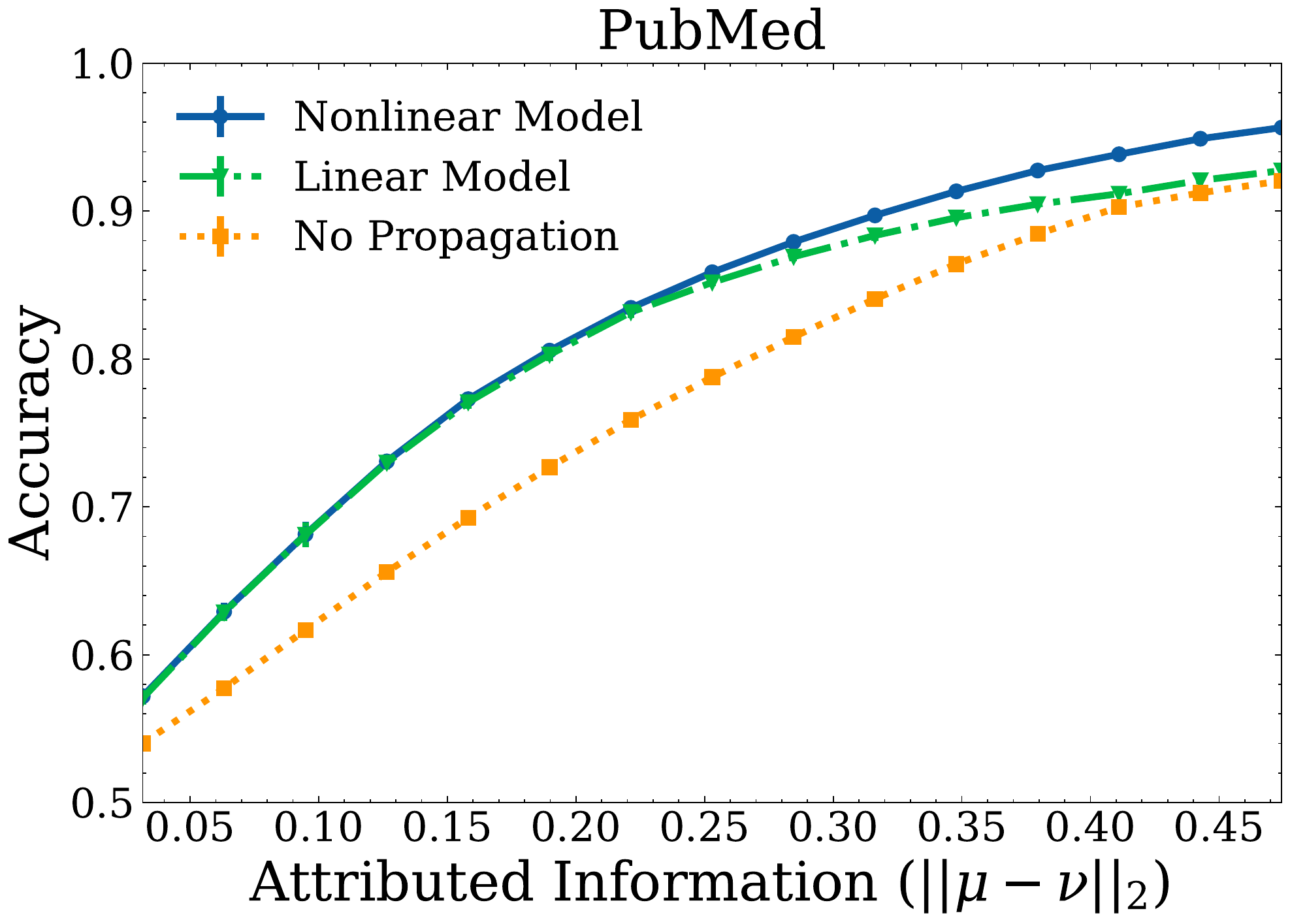}
\includegraphics[width=0.3\textwidth]{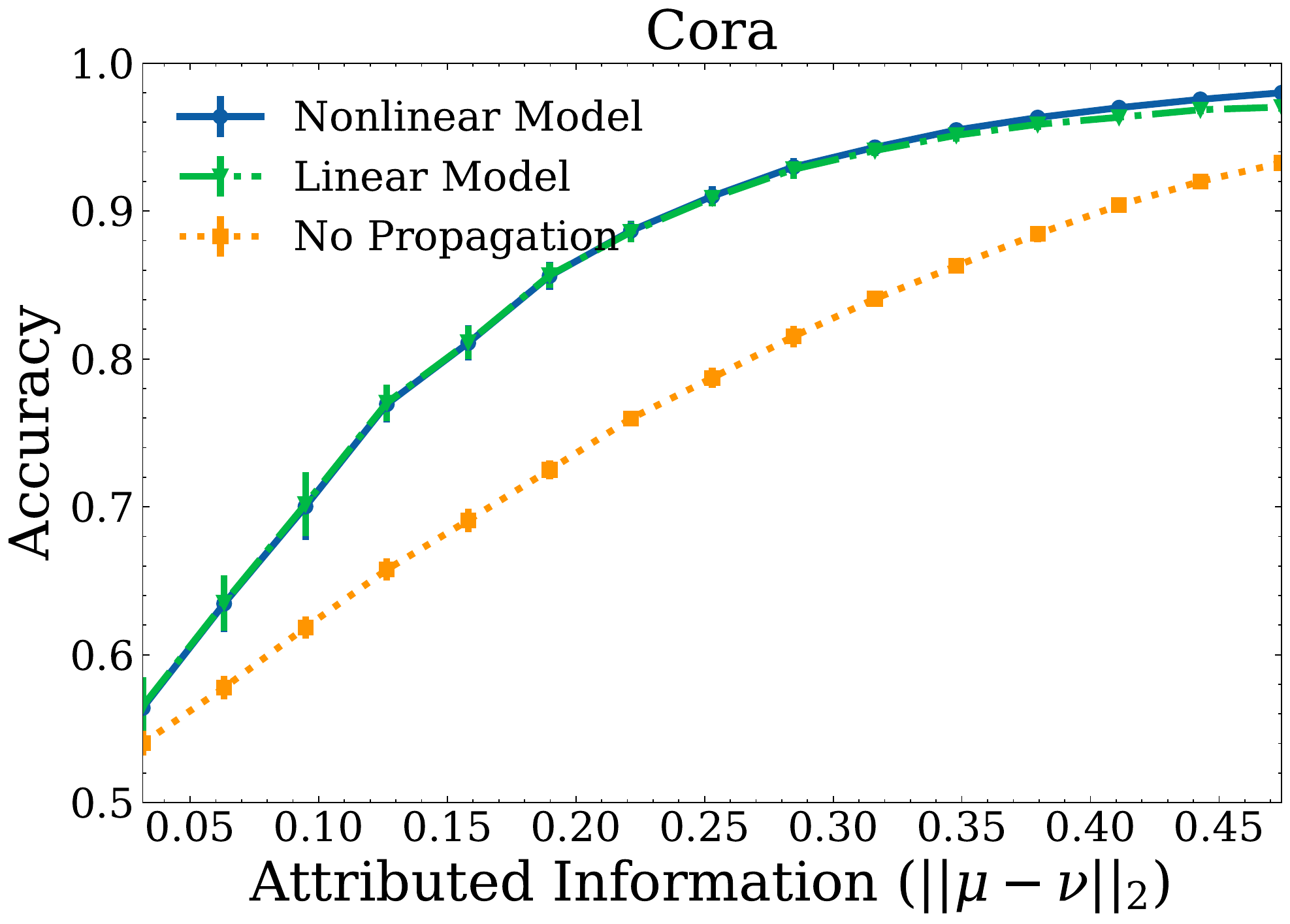}
\includegraphics[width=0.3\textwidth]{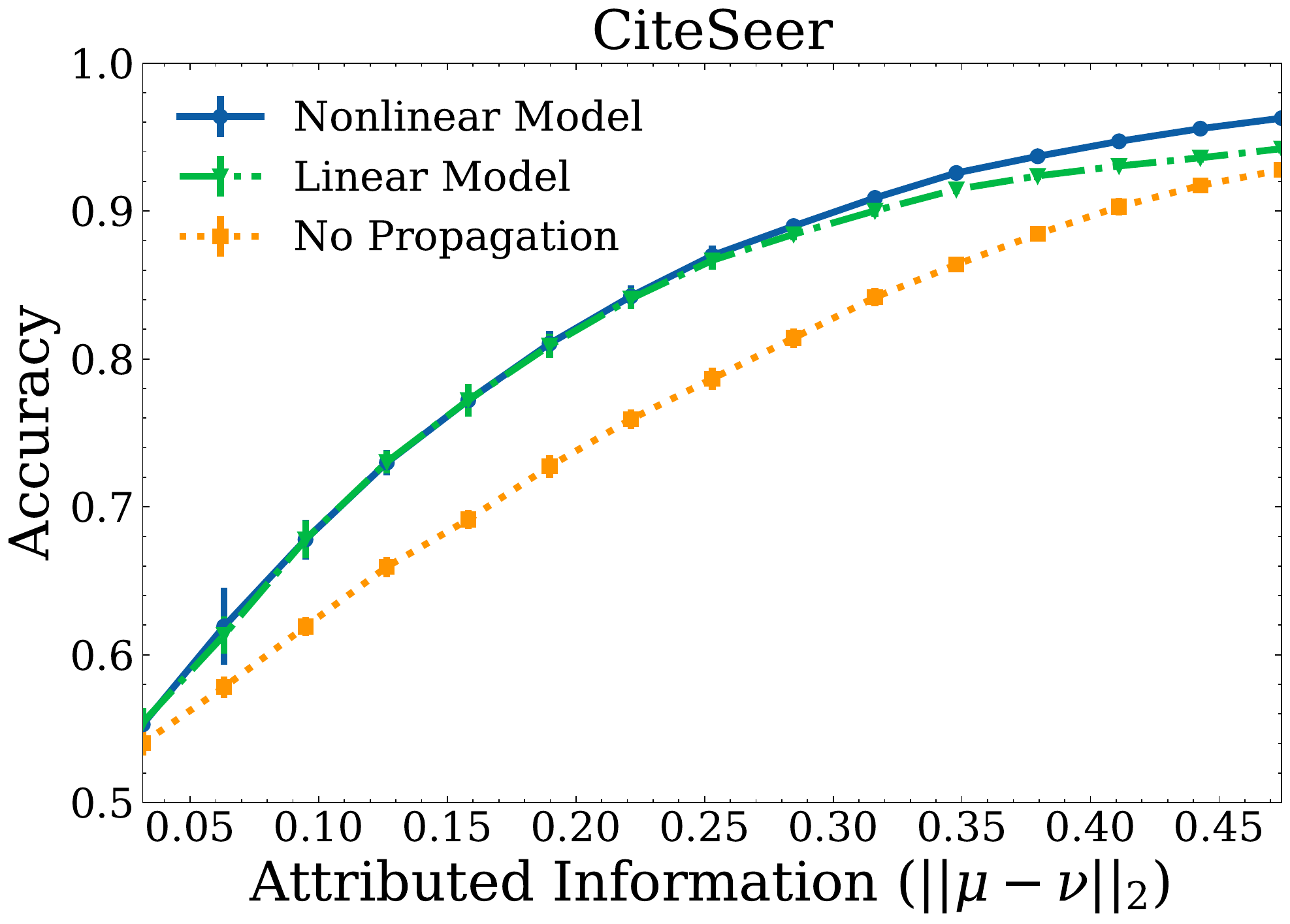}
\small{\caption{Averaged one-vs-all Classification Accuracy on Citation Networks of Nonlinear Models v.s. Linear Models. Node attributes in or out of the one class are generated from Gaussian distributions $\mathcal{N}(\mu, \frac{I}{m})$ and $\mathcal{N}(\nu, \frac{I}{m})$, $\mu,\nu\in\mathbb{R}^{m}$, respectively. The detailed settings are introduced in Sec.~\ref{sec:exp-real}.}}
\vspace{-2mm}
\end{figure}

Extensive evaluation on both synthetic and real datasets demonstrates our theory. Specifically, the node mis-classification errors of three citation networks with different levels of attributed information (Gaussian attributes) are shown in Fig.~\ref{fig.ova-gaussian}, which precisely matches the above conclusions.   
\vspace{-1mm}
\subsection{More Related Works}
\vspace{-1mm}
GNNs have achieved great empirical success while theoretical understanding of GNNs, their non-linearity in particular, is still limited. There are many works studying the expressive power of GNNs~\cite{maron2019provably,balcilar2021breaking,azizian2020expressive,murphy2019relational,sato2021random,abboud2021surprising,bodnar2021weisfeiler,chen2019equivalence,li2020distance,loukas2020graph,vignac2020building,zhang2021nested}, while they often assume arbitrarily complex non-linearity with limited quantitative results. Only a few works provide characterizations on the needed width or depth of GNN layers~\cite{li2020distance,vignac2020building,loukas2020graph,zhang2021nested}. More quantitative arguments on GNN analysis often depend on linear or Lipschitz continuous assumptions to enable graph spectral analysis, such as feature oversmoothing~\cite{oono2019graph,li2018deeper} and over-squashing~\cite{alon2020bottleneck,topping2021understanding}, the failure to process heterophilic graphs~\cite{zhu2020beyond,yan2021two,chien2021adaptive} and the limited spectral representation~\cite{balcilar2020analyzing,bianchi2021graph}. Some works also study the generalization bounds~\cite{du2019graph,garg2020generalization,liao2020pac} and the stability of GNNs~\cite{gama2019diffusion,gama2019stability,levie2021transferability,gama2020stability}. However, the obtained results may not reveal a direct comparison between non-linearity and linearity of the model, and their analytic techniques avoid tackling the specific forms of non-linear activations by using a Lipschitz continuous bound which is too loose in our case. 

Stochastic block models (SBM) and its contextual counterparts have been widely used to study the node classification problems~\cite{abbe2017community,abbe2018proof,massoulie2014community,bordenave2018nonbacktracking,montanari2016semidefinite,binkiewicz2017covariate,deshpande2018contextual}, while these studies focus on the fundamental limits. Recently, (C)SBM and its large-graph limitation also have been used to study the transferrability and expressive power of GNN models~\cite{keriven2020convergence,keriven2021universality,ruiz2020graphon} and GNNs on line graphs~\cite{chen2020supervised}, while these works did not compare non-linear and linear propagation. CSBM has also been used to show the advantage of linear convolution over no convolution for node classification~\cite{baranwal2021graph}. 
A very recent result shows that attention-based propagation~\cite{velivckovic2018graph} may be much worse than linear propagation given low-quality node attributes under CSBM~\cite{fountoulakis2022graph}. Our results imply that ReLU is the de facto optimal non-linearity instead of attention and may at most marginally outperform the linear model when with low-quality node attributes. Some previous works also use Bayesian inference to inspire GNN architectures~\cite{jin2019graph,qu2019gmnn,kuck2020belief,jia2022unifying,satorras2021neural,jia2021graph,qu2021neural,mehta2019stochastic}, while these works focus on empirical evaluation instead of theoretical analysis.

\vspace{-2mm}
\section{Preliminaries}\label{sec:prelim} 
\vspace{-1.5mm}
In this section, we introduce preliminaries and notations for our later discussion.

\textbf{Maximum-a-posteriori (MAP) estimation.} 
Suppose there are a set of finite classes $\mathcal{C}$. A class label $Y\in \mathcal{C}$ is generated with probability $\pi_Y$, where $\sum_{Y\in\mathcal{C}} \pi_Y = 1$. Given $Y$, the corresponding feature $X$ in the space $\mathcal{X}$ is generated from the distribution  $X\sim \mathbb{P}_{Y}$. A classifier is a decision $f:\mathcal{X}\rightarrow \mathcal{C}$ and the \emph{Bayesian mis-classification error} can be characterized as $\xi(f)= \sum_{Y\in \mathcal{C}} \pi_Y \int 1_{f(X)\neq Y} \mathbb{P}_{Y}(X)$, where and later $1_{S}$ indicates 1 if $S$ is true and 0 otherwise. The \emph{MAP estimation} of $Y$ given $X$ is the classifier $f^*(X) \triangleq \arg\max_{Y\in\mathcal{C}} \pi_Y \mathbb{P}_{Y}(X)$ that can minimize $\xi(f)$~\cite{theodoridis2015machine}. Later, we denote the minimal Bayesian mis-classification error $\xi(f)$ as $\xi^* = \xi(f^*)$.

\textbf{Signal-to-Noise Ratio (SNR).} Detection of a signal from the background essentially corresponds to a binary classification problem. SNR is widely used to measure the potential detection performance before specifying the classifier~\cite{poor2013introduction}. 
In particular, if we have two equiprobable classes $\mathcal{C} = \{-1,1\}$ and the features follows 1-d Gaussian distributions $\mathbb{P}_{Y} = \mathcal{N}(\mu_{Y},\sigma^2)$, $Y\in\mathcal{C}$. The SNR $\rho$ defined as follows precisely characterizes the minimal Bayesian mis-classification error.  \vspace{-0.5mm}
\begin{align}\label{eq:SNR}
    \textbf{SNR:} \quad \rho = \frac{\text{mean difference}^2}{\text{variance}}= \frac{(\mu_{1} - \mu_{-1})^2}{\sigma^2}.
\end{align}
In this case, the MAP estimation $f^*(X) = 2*1_{|X - \mu_1|\geq |X-\mu_{-1}|}-1$ and the minimal Bayesian mis-classification error is  $\Phi(-\sqrt{\rho}/2)$ where $\Phi$ denotes the cumulative standard Gaussian distribution function. For more general cases where
the two classes are associated with sub-Guassian distributions $\mathbb{P}_{Y}$, s.t. $\mathbb{P}_{Y}(|X| > t) \in [c_1\exp(- c_2t^2),C_1\exp(- C_2t^2)]$, for some non-negative constants $c_1,c_2,C_1,C_2$, a similar connection between $\xi^*$ and $\rho$ can be shown by leveraging sharp sub-Gaussian lower bounds~\cite{zhang2020non}. 
We will specify the connection to SNR in our case in Sec.~\ref{sec:main} and the SNR $\rho$ will be used as the main bridge to compare the mis-classification errors of  non-linear v.s. linear models. 

\textbf{Contextual Stochastic Block Model (CSBM).} Random graph models have been widely used to study the performance of algorithms on graphs~\cite{ding2021efficient,li2019optimizing}. For node classification problems, CSBM is often used~\cite{keriven2020convergence,keriven2021universality,ruiz2020graphon},  as it well combines the models of network structure and node attributes. \vspace{-1mm}

We study the case that nodes are in two equi-probable classes $\mathcal{C}=\{-1,1\}$, where $\pi_{Y}=\frac{1}{2}, Y\in\mathcal{C}$. Our analysis can be generalized. An attributed network $\mathcal{G}=(\mathcal{V},\mathcal{E},\mathbf{X})$ is sampled from CSBM with parameters $(n, p, q, \mathbb{P}_{1}, \mathbb{P}_{-1})$ as follows. Suppose there are $n$ nodes, $\mathcal{V}=[n]$. For each node $v$, the label $Y_v\in\mathcal{C}$ is sampled from Rademacher distribution. Given $Y_v$, the node attribute $X_v$ is sampled from $\mathbb{P}_{Y_v}$. For two nodes $u,v$, if $Y_u=Y_v$, there is an edge $e_{uv}\in\mathcal{E}$ connecting them with probability $p$. If $Y_u\neq Y_v$, there is an edge $e_{uv}\in\mathcal{E}$ connecting them with probability $q$. All node attributes $\mathbf{X}$ and edges $\mathcal{E}$ are independent given the node labels $\mathbf{Y}=\{Y_v|v\in\mathcal{V}\}$. 

Note that $p>q$ indicates the nodes with the same labels tend to be directly connected, which corresponds to the \emph{homophilic} case, while $p<q$ corresponds to the \emph{heterophilic} case. 

The gap $|p-q|$, representing probabilities difference of a node connects to nodes from the same class or the different class, reflects \emph{structural information}  and the gap between $\mathbb{P}_{1},\,\mathbb{P}_{-1}$ reflects \emph{attributed information}, e.g., Jensen-Shannon distance $\text{JS}(\mathbb{P}_{1}, \mathbb{P}_{-1})$ that is well connected to Bayesian mis-classification error~\cite{lin1991divergence}. Graph learning allows combining these two types of information. In Sec.~\ref{sec:main}, we give more specific definitions of these two types of information and their regime for our analysis.

\vspace{-2mm}
\section{Bayesian Inference and Nonlinearity in Graph Neural Networks}  
\vspace{-1mm}
\label{sec:bayes}
In the previous section, we discuss that given conditioned feature distributions $X\sim\mathbb{P}_{Y}$, $Y\in\mathcal{C}$, the MAP estimation $f^*(X)$ can minimize mis-classification error. For node classification in an attributed network, the estimation of a node label should depend on not only one's own attributes but also its neighbors'. For example, in a homophilic network, nodes with same labels tend to be directly connected. Intuitively, using the averaged neighbor attributes may provide better estimation of the label, which gives us graph convolution. In a heterophilic network, nodes with different labels tend to be directed connected. So, intuitively, checking the difference between one's attributes and the neighbors' may provide better estimation. However, what could be the optimal form to combine one's own attribute with the neighbors' attributes? We resort to the MAP estimation. That is, given the attributes of a node $v\in\mathcal{V}$ and its neighbors $\mathcal{N}_v$, we consider the MAP estimation as follows.
\begin{align*}
    f^*(X_v,\{X_{u}\}_{u\in\mathcal{N}_v}) = \argmax_{Y_v\in\mathcal{C}}\max_{Y_u\in\mathcal{C},\forall u\in\mathcal{N}_v}  \pi_{Y_v,\{Y_{u}\}_{u\in\mathcal{N}_v}}\mathbb{P}\left(X_v,\{X_{u}\}_{u\in\mathcal{N}_v},\mathcal{N}_v|Y_v,\{Y_{u}\}_{u\in\mathcal{N}_v}\right),
\end{align*}
where $\pi_{Y_v,\{Y_{u}\}_{u\in\mathcal{N}_v}}$ denotes their prior distributions of node labels. Note that here we simplify the problem and consider only 1-hop neighbors by following the setting~\cite{baranwal2021graph}. In practice, most GNN models can only work on local networks due to the scalability constraints~\cite{hamilton2017inductive,zeng2019graphsaint,yin2022algorithm}. 
Even with the above simplification, the above MAP estimation is generally intractable.

Therefore, we consider the CSBM with parameters $(n, p, q, \mathbb{P}_{1}, \mathbb{P}_{-1})$. In this case, the prior distribution follows $\pi_{Y_v,\{Y_{u}\}_{u\in\mathcal{N}_v}} = 2^{-|\mathcal{N}_v|-1}$, which is a constant given $\mathcal{N}_v$. The rest term follows
\begin{align}
    \mathbb{P}\left(X_v,\{X_{u}\}_{u\in\mathcal{N}_v},\mathcal{N}_v|Y_v,\{Y_{u}\}_{u\in\mathcal{N}_v}\right) &= \mathbb{P}\left(X_v,\{X_{u}\}_{u\in\mathcal{N}_v}|Y_v,\{Y_{u}\}_{u\in\mathcal{N}_v}\right)\mathbb{P}\left(\mathcal{N}_v|Y_v,\{Y_{u}\}_{u\in\mathcal{N}_v}\right) \nonumber \\
    &=\prod_{u\in\mathcal{N}_v\cup\{v\}}\mathbb{P}_{Y_u}\left(X_u\right)\prod_{u\in\mathcal{N}_v} p^{(1+Y_vY_u)/2}q^{(1-Y_vY_u)/2}\label{eq:potential}
\end{align}
Therefore, the MAP estimation $f^*(X_v,\{X_{u}\}_{u\in\mathcal{N}_v})$ is to solve 
\begin{align}\label{eq:MAP}
f^*(X_v,\{X_{u}\}_{u\in\mathcal{N}_v}) = \argmax_{Y_v\in\mathcal{C}} \; \mathbb{P}_{Y_v}\left(X_v\right)\prod_{u\in\mathcal{N}_v} \max_{Y_u\in\mathcal{C}}\;\mathbb{P}_{Y_u}\left(X_u\right)p^{(1+Y_vY_u)/2}q^{(1-Y_vY_u)/2}
\end{align}
This can be solved via the max-product algorithm~\cite{weiss2001optimality}. To establish the connection to GNNs, we rewrite the RHS of Eq.~\ref{eq:MAP} in the logarithmic form and use the fact that $\mathcal{C}=\{-1,1\}$. And, we achieve
\begin{align}
& f^*(X_v,\{X_{u}\}_{u\in\mathcal{N}_v}) = \text{sgn}\left(\log \frac{\mathbb{P}_{1}\left(X_v\right)}{\mathbb{P}_{-1}\left(X_v\right)} 
+ \sum_{u\in\mathcal{N}_v} \mathcal{M}(X_u,p,q)\right), \quad \text{where}\nonumber\\
& \mathcal{M}(X_u,p,q)= \text{ReLU}\left(\log\frac{\mathbb{P}_{1}\left(X_v\right)}{\mathbb{P}_{-1}\left(X_v\right)} + \log \frac{p}{q}\right) - \text{ReLU}\left(\log\frac{\mathbb{P}_{1}\left(X_v\right)}{\mathbb{P}_{-1}\left(X_v\right)} + \log \frac{q}{p}\right) + \log \frac{q}{p}. \nonumber
\end{align}
We leave the derivation in Appendix~\ref{app:pro}. Amazingly, activation ReLUs in the message $\mathcal{M}$ well connect to the activations commonly-used in GNN models, e.g., graph convolution networks~\cite{kipf2016semi}. Given the optimality of the MAP estimation, we summarize this observation in  Proposition~\ref{prop:nonlinear}.
\begin{proposition}[\textbf{Optimal Nonlinear Propagation}]\label{prop:nonlinear}
\label{pro.onp}
Consider a network $\mathcal{G}\sim$ CSBM$(n, p, q, \mathbb{P}_{1}, \mathbb{P}_{-1})$. To classify a node $v$, the optimal nonlinear propagation (derived by the MAP estimation) given the attributes of $v$ and its neighbors follows: 
\begin{align}\label{eq:prop}
    \mathcal{P}_v = \psi\left(X_v;\mathbb{P}_1,\mathbb{P}_{-1}\right) + \sum_{u \in \mathcal{N}_v}\phi\left(\psi\left(X_u;\mathbb{P}_1,\mathbb{P}_{-1}\right); \log(p/q)\right)
\end{align}
where $\psi\left(a;\mathbb{P}_1,\mathbb{P}_{-1}\right) = \log\frac{\mathbb{P}_{1}\left(a\right)}{\mathbb{P}_{-1}\left(a\right)}$ and $\phi(a;\log\frac{p}{q}) = \text{ReLU}(a+\log\frac{p}{q}) - \text{ReLU}(a-\log\frac{p}{q})-\log\frac{p}{q}.$
\end{proposition}
\begin{wrapfigure}{r}{4.3cm}
\vspace{-5mm}
\centering
\includegraphics[height=2.0cm]{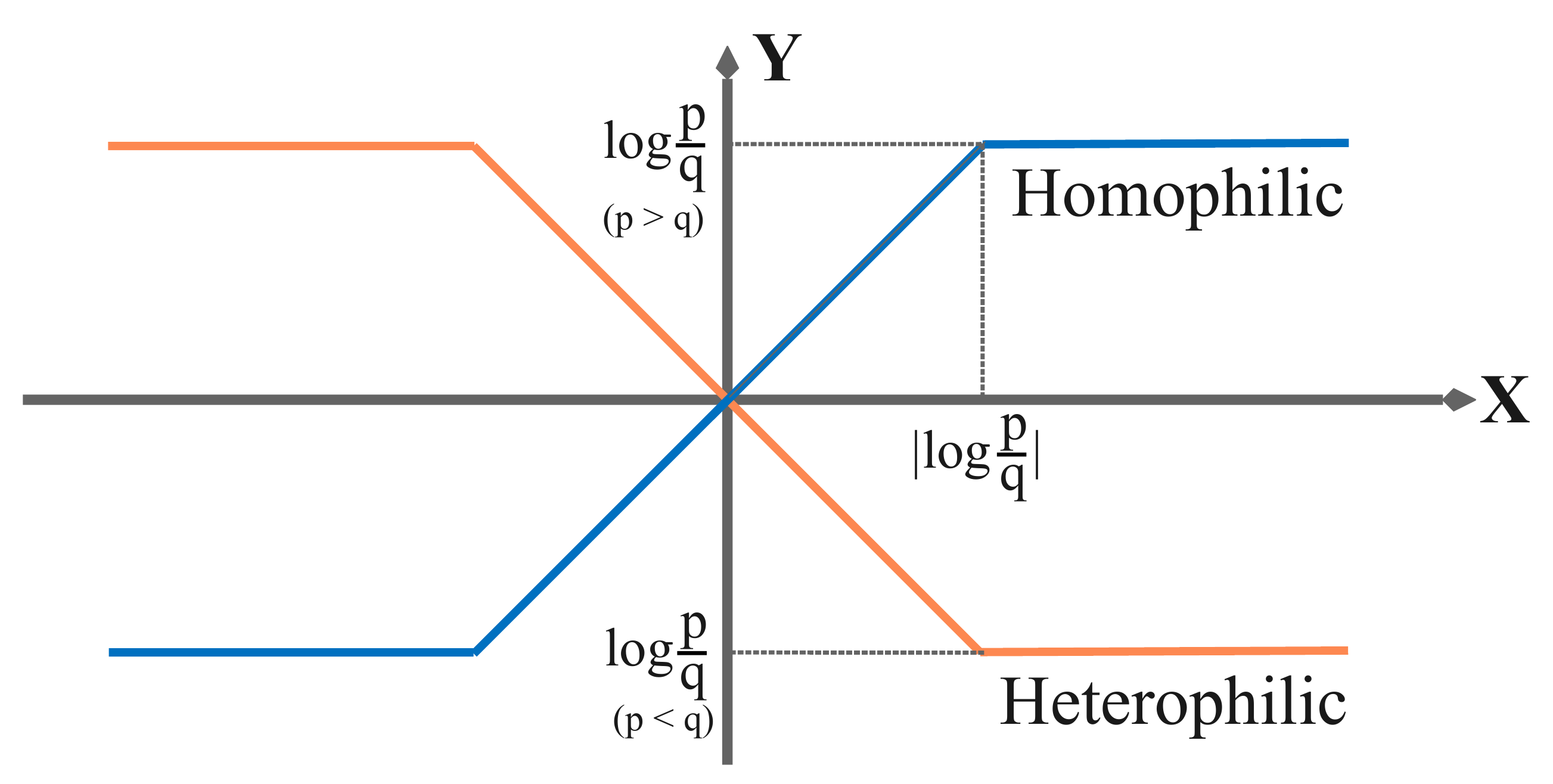}
\vspace{-3mm}
\small{\caption{Function $\phi(x;\log\frac{p}{q})$.\label{fig:nl-agg}}}
\vspace{-5mm}
\end{wrapfigure} 
The optimal nonlinear propagation in Eq.~\eqref{eq:prop} may contain two types of non-linear functions: (1) $\psi$ is to measure the likelihood ratio between two classes given the node attributes; (2) $\phi$ is to propagate the likelihood ratios of the neighbors. ReLUs in $\phi$ avoid the overuse of the likelihood ratios from neighbors, as $\phi$ essentially provides a bounded function (See Fig.~\ref{fig:nl-agg}). One observation of the direct benefit of this non-linear propagation is as follows.  

\begin{remark}\label{remark:no-struc}
When there is no structural information, i.e., $p=q$, $\phi(x;0)=0$, $\forall x\in\mathbb{R}$, the propagation is deactivated, which avoids potential contamination from the attributes of the neighbors. 
\end{remark} 
In the equiprobable case, the MAP estimation also gives the maximum likelihood estimation (MLE) of $Y$ if we view the labels as the fixed parameters. When the classes are unbalanced $\pi_{Y}\neq\frac{1}{2}$, similar results can be obtained while additional terms  $\log\frac{\pi_1}{\pi_{-1}}$ may appear as bias in Eq.~\eqref{eq:prop}. Later, our analysis focuses on the equiprobable case while empirical results in Sec.~\ref{sec:exp} show more general cases.

Moreover, if one is to infer the posterior distribution of $Y$, one may replace the max-product algorithm to solve Eq.~\eqref{eq:MAP} with the sum-product algorithm~\cite{pearl1982reverend}. Then, the obtained non-linearity in $\phi$ will turn into Tanh functions. As ReLUs are more used in practical GNNs, we focus on the case with ReLUs.


\textbf{Discussion on the Non-linearity.} Next, we discuss more insights into the non-linearity of $\psi$ and $\phi$. 

The function $\psi$ essentially corresponds to \emph{a node-attribute transformation}, which depends on the distributions $\mathbb{P}_{\pm1}$. As these distributions are unknown in practice, a NN model to learn $\psi$ is suggested, such as the one in the model APPNP~\cite{klicpera2018predict} and GPR-GNN~\cite{chien2021adaptive}. Due to the expressivity of NNs~\cite{cybenko1989approximation,hornik1989multilayer}, a wide range of $\psi$ can be modeled. One insightful example is that when $\mathbb{P}_{\pm1}$ are Laplace distributions, $\psi$ is a bounded function (same as $\phi$) to control the heavy-tailed attributes generated by Laplace distributions.   


\begin{exmp}[Laplace Assumption]
\label{eg.lap}
    When node attributes follow $m$-dimensional independent Laplace distribution, i.e., $\mathbb{P}_{Y_v}=\frac{1}{(2b)^m}\exp(-\|X_v - Y_v \mu\|_1/b)$ for $\mu\in\mathbb{R}^m$, $b>0$ and $Y_v \in \{-1, 1\}$. According to Eq.~\eqref{eq:prop}, the function $\psi(\cdot;\mathbb{P}_{1},\mathbb{P}_{-1})$ can be specified as 
    \begin{align} \nonumber
        \psi_{\text{lap}}(X_v;\mathbb{P}_{1},\mathbb{P}_{-1}) = \mathbbm{1}^T\phi(X_v; 2\mu/b),\;\text{where}\; \phi\;\text{as defined in Eq.~\eqref{eq:prop} works in an entry-wise manner.}
    \end{align}
\end{exmp}

As node-attribute distributions may vary a lot, $\psi$ is better to be modeled via a general NN in practice. More interesting findings may come from  $\phi$ in Eq.~\eqref{eq:prop} as it has a fixed form and well matches the most commonly-used GNN architecture. Specifically, besides the extreme case stated in Remark~\ref{remark:no-struc}, we are to investigate \emph{how} non-linearity induced by the ReLUs in $\phi$ may benefit the model. We expect the findings to provide the explanations to some previous empirical experiences on using GNNs. 

To simplify our discussion, when analyzing $\phi$, we focus on the case with a linear node-attribute transformation $\psi=\psi_{\text{Gau}}$ in Eq.~\eqref{eq:gau} by assuming label-dependent Gaussian node attributes. This follows the assumptions in previous studies~\cite{baranwal2021graph,fountoulakis2022graph}. 
In fact, there are a class of distributions named natural exponential family (NEF)~\cite{morris1982natural} which if the node attributes satisfy, the induced $\phi$ is linear. We conjecture that our later results in Sec.~\ref{sec:main} are applied to the general NEF since the only difference is the bias term by comparing Eq.~\eqref{eq:nef} and Eq.~\eqref{eq:gau}. 

\begin{exmp}[Natural Exponential Family Assumption] 
    When node features follow $m$-dimensional natural exponential family distributions $\mathbb{P}_{Y_v}(X) = h(X_v) \cdot \exp(\theta_{Y_v}^TX_v - M(\theta_{Y_v}))$ for $\theta_{Y_v}\in\mathbb{R}^m$ and $Y_v \in \{-1, 1\}$ where $M(\theta_{Y_v})$ is a parameter function. The function $\psi(\cdot;\mathbb{P}_{1},\mathbb{P}_{-1})$ is specified as:
    \begin{align}
        \psi_{\text{nef}}(X_v;\theta_1,\theta_{-1}) = (\theta_1 - \theta_{-1})^TX_v - (M(\theta_1) - M(\theta_{-1})).
        \label{eq:nef}
    \end{align}
    In particular, when $\mathbb{P}_{1}=\mathcal{N}(\mu, I/m)$, $\mathbb{P}_{-1}=\mathcal{N}(\nu, I/m)$ for $\mu,\nu\in\mathbb{R}^m$, 
    \begin{align} \label{eq:gau}
        \psi_{\text{Gau}}(X_v;\mu,\nu)= m\left[(\mu-\nu)^TX_v - (\|\mu\|_2^2-\|\nu\|_2^2)/2\right].
    \end{align}
\end{exmp}\vspace{-1mm}

More generally, our optimal nonlinear propagation Eq.~\eqref{eq:prop} can be well generalized to other settings as long as the model satisfies \textit{edge-independent assumption}, where edges random variables are mutually independent conditioned on the labels of nodes. When this assumption is satisfied, the MAP estimation will result in graph convolution with ReLU activation.

We summarize our main theoretical findings regarding the nonlinearity of $\phi$ in the next section.

 \vspace{-1mm}
\section{Main Results on ReLU-based Nonlinear Propagation}  \label{sec:main}
\vspace{-1mm}
In this section, we summarize our analytical results on $\phi$ in the optimal nonlinear propagation (Eq.~\eqref{eq:prop}). 
Our study assumes an attributed network generated from  CSBM$(n,p,q,\mathcal{N}(\mu, I/m),\mathcal{N}(\nu, I/m))$ where $\mu,\nu\in\mathbb{R}^m$. We use CSBM-G$(n,p,q,\mu,\nu)$ later to denote this model for simplicity. 
We are interested in the asymptotic behavior when $n\rightarrow \infty$. Note that all parameters $\mu,\nu,p,q,m$ may implicitly depend on $n$. We are to compare the non-linear propagation model $\mathcal{P}_v$ suggested by Eq.~\eqref{eq:prop} where $\psi=\psi_{\text{Gau}}$ with the following linear counterpart $\mathcal{P}_v^l$.\vspace{-1mm}
\begin{align}\label{eq:linear}
    \textbf{Baseline linear model:} \quad \mathcal{P}_v^{l}(w)&= \psi_{\text{Gau}}(X_v;\mu,\nu)  + w\sum_{u\in\mathcal{N}_v}\psi_{\text{Gau}}(X_u;\mu,\nu), \;\text{for all $v\in\mathcal{V}$.}
\end{align}
where $w\in\mathbb{R}$ is an extra parameter to be tuned. Note that this linear model can be claimed as an optimal linear model up-to a choice of $w$ because the distributions of both the center node attribute $X_v$ and the linear aggregation from the neighbors  $\sum_{u\in\mathcal{N}_v}X_u$ are Gaussian and symmetric w.r.t. the hyperplane $\{Z\in\mathbb{R}^m|(\mu-\nu)^TZ=(\|\mu\|_2^2-\|\nu\|_2^2)/2\}$ for the two classes. 
We are to compare their classification errors $\xi^r = \xi(\text{sgn}(\mathcal{P}_v))$ and $\xi^l(w) = \xi(\text{sgn}(\mathcal{P}_v^l(w)))$. By following~\cite{baranwal2021graph}, we also discuss separability of all nodes in the network, i.e., $\mathbb{P}(\forall v \in\mathcal{V}, \mathcal{P}_v \cdot Y_v > 0)$ in Theorem 1. 

To begin with, we introduce several quantities for the convenience of further statements. The SNRs
\[\rho_r = \frac{\left(\mathbb{E}[\mathcal{P}_{v}|Y_{v}=1] - \mathbb{E}[\mathcal{P}_{v}|Y_{v}=-1]\right)^2}{ \text{var}(\mathcal{P}_{v}|Y_{v}=1)},\rho_l(w) = \frac{\left(\mathbb{E}[\mathcal{P}_{v}^l(w)|Y_{v}=1] - \mathbb{E}[\mathcal{P}_{v}^l(w)|Y_{v}=-1]\right)^2}{ \text{var}(\mathcal{P}_{v}^l(w)|Y_{v}=1)}\] 
 are important quantities to later characterize different types of propagation. Also, we characterize structural information by  $\mathcal{S}(p, q) = (p - q)^2 / (p + q)$ and attributed information by $\sqrt{m}\|\mu - \nu\|_2$.
\begin{assumption}[Moderate Structural Information] $\mathcal{S}(p, q) = \omega_n(\frac{(\log n)^2}{n})$ and $\frac{\mathcal{S}(p, q)}{|p-q|}  \nrightarrow 1 $.
\label{as.1} 
\end{assumption}

\begin{assumption}[Bounded Attributed Information]
\label{as.2}
$\sqrt{m}\|\mu - \nu\|_2 = o_n(\log n)$.
\end{assumption}

Assumption~\ref{as.1} states that structural information should be neither too weak nor too strong. $\mathcal{S}(p, q) = \omega_n(\frac{(\log n)^2}{n})$ excludes the extremely weak case discussed in Remark~\ref{remark:no-struc}. Moreover, the graph structure should not be too sparse, so the aggregated information from neighbors dominates the propagation. $\frac{\mathcal{S}(p, q)}{|p-q|}  \nrightarrow 1 $ means neither $p = \omega_n (q)$ nor $q = \omega_n (p)$, which avoids extremely strong structural information.  
This assumption is more general than some concurrent works on CSBM-based GNN analysis~\cite{baranwal2021graph,fountoulakis2022graph} as we include the cases with less structural information $|p-q|=o_n(p+q)$ and with heterophily $p<q$. Assumption~\ref{as.2} is to avoid too strong attributed information: when $\sqrt{m}\|\mu - \nu\|_2=\Omega_n(\log n)$, all nodes in CSBM can be accurately classified in the asymptotic sense without structural information, i.e. $\mathbb{P}(\forall v \in \mathcal{V},\,\psi_{\text{Gau}}(X_v;\mu,\nu)\cdot Y_{v} >0) = 1 - o_n(1)$. Now, we present our first lemma which links the mis-classification errors $\xi^r, \xi^l$ to with the SNRs $\rho_r,\rho_l$:
\begin{lemma}
\label{lm.1}
   Suppose $(p, q)$ satisfies Assumption~\ref{as.1}, for any $\mathcal{G} \sim$ CSBM-G$(n,p,q,\mu,\nu)$, 
   \begin{align}
       \xi^{r} \in [C_1\exp(-C_2\rho_r / 2), \exp(-\rho_r / 2)],\quad\xi^{l}(w) \to \exp(-\rho_l(w)(1 + o_n(1)) /2 )
   \end{align}
\end{lemma}

where $C_2$ is asymptotically a constant, and the notation $a(n)\to b(n)$ denotes $a(n) / b(n) \to 1$. Lemma~\ref{lm.1} claims that the classification errors under both nonlinear and linear model can be controlled by their SNRs. By leveraging Lemma~\ref{lm.1}, we can further illustrate the separability of all nodes in the network, which is presented in the following theorem.

\begin{theorem}[Separability]
\label{thm.1}
    Suppose that $(p, q)$ satisfies Assumption~\ref{as.1}, for $\mathcal{G} \sim$ CSBM-G$(n,p,q,\mu,\nu)$, if $\sqrt{m}\|\mu - \nu\|_2 = \omega_n(\sqrt{\log n / S(p, q)n})$, then
\begin{align}
            &\mathbb{P}(\forall v \in\mathcal{V}, \mathcal{P}_v \cdot Y_v > 0) = 1-\mathcal{O}_n(n\exp(-\rho_r / 2))=1-o_n(1),\\
            &\mathbb{P}(\forall v \in\mathcal{V}, \mathcal{P}_v^l(w) \cdot Y_v > 0)  = 1- \mathcal{O}_n(n\exp(-\rho_l(w) / 2)) = 1- o_n(1). \label{eq.linear_separability}
\end{align}
Here, assume $|w| > c$ for some positive constant $c$ and $\text{sgn}(w)=\text{sgn}(p-q)$ in the linear model.

\end{theorem}
Theorem~\ref{thm.1} applies to both homophilic ($p>q$) and heterophilic $(p<q)$ scenarios. Even for just the linear case, compared to~\cite{baranwal2021graph} which  needs $\sqrt{m}\|\mu - \nu\|_2 = \omega_n(\log n/\sqrt{S(p, q)n})$ to achieve separability, we have $\sqrt{\log n}$ improvement due to a tight analysis. 

As shown in Lemma~\ref{lm.1} and Theorem~\ref{thm.1}, the errors are mainly determined by SNRs. Large SNR implies a fast decay rate of the errors of a single node and the entire graph, which motivates us to further explore SNRs to illustrate a comparison between  non-linear and linear models. We consider comparing with the optimal linear model, i.e., $\rho_l^*= \rho_l(w^*)$, where  $w^*=\arg\min_{w\in\mathbb{R}}\xi^l(w)$. 

\begin{theorem}
\label{thm:2}
    Suppose that $(p, q)$ satisfies Assumption~\ref{as.1}, for $\mathcal{G} \sim$ CSBM-G$(n,p,q,\mu,\nu)$, under the separable condition in Theorem~\ref{thm.1} $\sqrt{m}\|\mu - \nu\|_2 = \omega_n(\sqrt{\log n / S(p, q)n})$, we further have\vspace{-1mm}
        \begin{itemize}[leftmargin=5mm]
            \item \textbf{I. Limited Attributed Information}: When $\sqrt{m}\|\mu - \nu\|_2 = \mathcal{O}_n(1)$, \vspace{-1mm}
                \begin{align}\label{eq:lim-regime}
                    \rho_r = \Theta_n(\rho_l^*),
                \end{align}
                Further, if $\sqrt{m}\|\mu - \nu\|_2 = o_n(|\log(p/q)|)$, $\rho_r / \rho_l^* \to 1$;\vspace{-2mm}
            \item \textbf{II. Sufficient Attributed Information}: When $\sqrt{m}\|\mu - \nu\|_2  = \omega_n(1)$ and satisfies Assumption~\ref{as.2}, 
            \begin{align}\label{eq:suff-regime}
                \rho_r =& \omega_n(\min\{\exp(m\|\mu - \nu\|_2^2/3), nS(p, q) m^{-1}\|\mu - \nu \|_2^{-2}\} \cdot \rho_l^*) = \omega_n(\rho_l^*).
            \end{align}
        \end{itemize}\vspace{-2mm}
\end{theorem}

Theorem~\ref{thm:2} also works both homophilic ($p>q$) and heterophilic $(p<q)$ scenarios. Theorem~\ref{thm:2} implies that when attributed information is limited,  nonlinear propagation behaves similar to the linear model as their SNRs are in the same order. Particularly, when attributed information is very limited, $\sqrt{m}\|\mu - \nu\|_2 = o_n(|\log(p/q)|)$, the SNRs of two models are asymptotically the same. 
In the regime of sufficient attributed information, nonlinear propagation brings order-level superiority compared with the linear model. The intuition is that in this regime, when the attributes are very informative, the bounds of $\phi$ in Eq.~\eqref{eq:prop} help with avoiding overconfidence given by the node attributes. The coefficient before $\rho_l^*$ in Eq.~\eqref{eq:suff-regime} shows the trade-off between structural information and attributed information on controlling the superiority of nonlinear propagation. 


Next, we analyze whether nonlinearity makes model more transferable or not when there often exists a distribution shift between the training and testing datasets, which is also practically useful. 

We consider the following setting. We assume using a large enough network generated by CSBM-G$(n,p,q,\mu,\nu)$ for training so that the optimal parameters as in $\mathcal{P}_v$ and $\mathcal{P}_v^l(w^*)$ for this CSBM-G have been learnt. We consider their mis-classification errors over another CSBM-G with parameters $(n,p',q',\mu',\nu')$. We keep the amounts of attributed information and structural information unchanged by setting $p=p'$, $q=q'$, $\mu' = (\mu + \nu) / 2 + \bm{R}(\mu - \nu) / 2, \nu' = \mu + \nu /2 + \bm{R}(\nu - \mu ) / 2$ for a rotation matrix $\bm{R}$ close to $I$. Let $\Delta\xi^r$ and $\Delta\xi{^l}(w^*)$ denote the increase of mis-classification errors of models $\mathcal{P}_v$ and $\mathcal{P}_v^l(w^*)$, respectively, due to such a distribution shift. We may achieve the following results. 


\begin{theorem}[Transferability]
\label{thm:3}
    Suppose that $(p, q)$ satisfies Assumption~\ref{as.1}, for $\mathcal{G}' \sim$ CSBM-G$(n,p',q',\mu',\nu')$, under the linear separable condition $\sqrt{m}\|\mu' - \nu'\|_2 = \omega_n(\sqrt{\log n / S(p', q')n})$. Suppose $\mathcal{P}_v$ and $\mathcal{P}_v^l(w^*)$ have learnt parameters from $\mathcal{G} \sim$ CSBM-G$(n,p,q,\mu,\nu)$ where the parameters of two CSBM-Gs follow the relation described above. Then, we have \vspace{-1mm}
    \begin{itemize}[leftmargin=5mm]
        \item \textbf{I. Limited Attributed Information}: When $\sqrt{m}\|\mu - \nu\|_2 = o_n(|\log(p/q)|)$, $ \Delta \xi^r/\Delta \xi^l(w^*) \to 1$. 
        \item \textbf{II. Sufficient Attributed Information}: When $\sqrt{m}\|\mu - \nu\|_2 = \omega_n(1)$ and satisfies Assumption~\ref{as.2}, $ \Delta \xi^r/ \Delta \xi{^l}(w^*) \to 0$.   
    \end{itemize}\vspace{-3mm}
\end{theorem}

Similar to Theorem~\ref{thm:2}, 
when attributed information is very limited, nonlinearity will not bring any benefit, while in the regime with informative attributes, nonlinearity  increases  model transferability. We leave the intermediate regime $\sqrt{m}\|\mu - \nu\|_2\in[\Omega_n(|\log(p/q)|), O_n(1)]$ for future study.




\vspace{-1mm}
\section{Experiments} \label{sec:exp}
\vspace{-1mm}
In this section, we verify our theoretical results based on synthetic and real datasets. In all experiments, we fix $w$ in the linear model (Eq.~\eqref{eq:linear}) as $w=1$ for the homophilic case $(p>q)$ and $w=-1$ for the heterophilic case $(p<q)$. Experiments on other $w$'s can be found in Appendix~\ref{app:exp_w}, which does not change the achieved conclusion. This is because when the node number $n$ is large, for a constant $w$, the neighbor information will dominate the results. Later, we use $\mathcal{P}_v^l=\mathcal{P}_v^l(w)$ for simplicity.

\vspace{-1mm}
\subsection{Asymptotic Experiments - Model Accuracy \& Transferability Study} \label{sec:asmp}
\begin{figure}[t]
\centering
\includegraphics[width=0.32\textwidth]{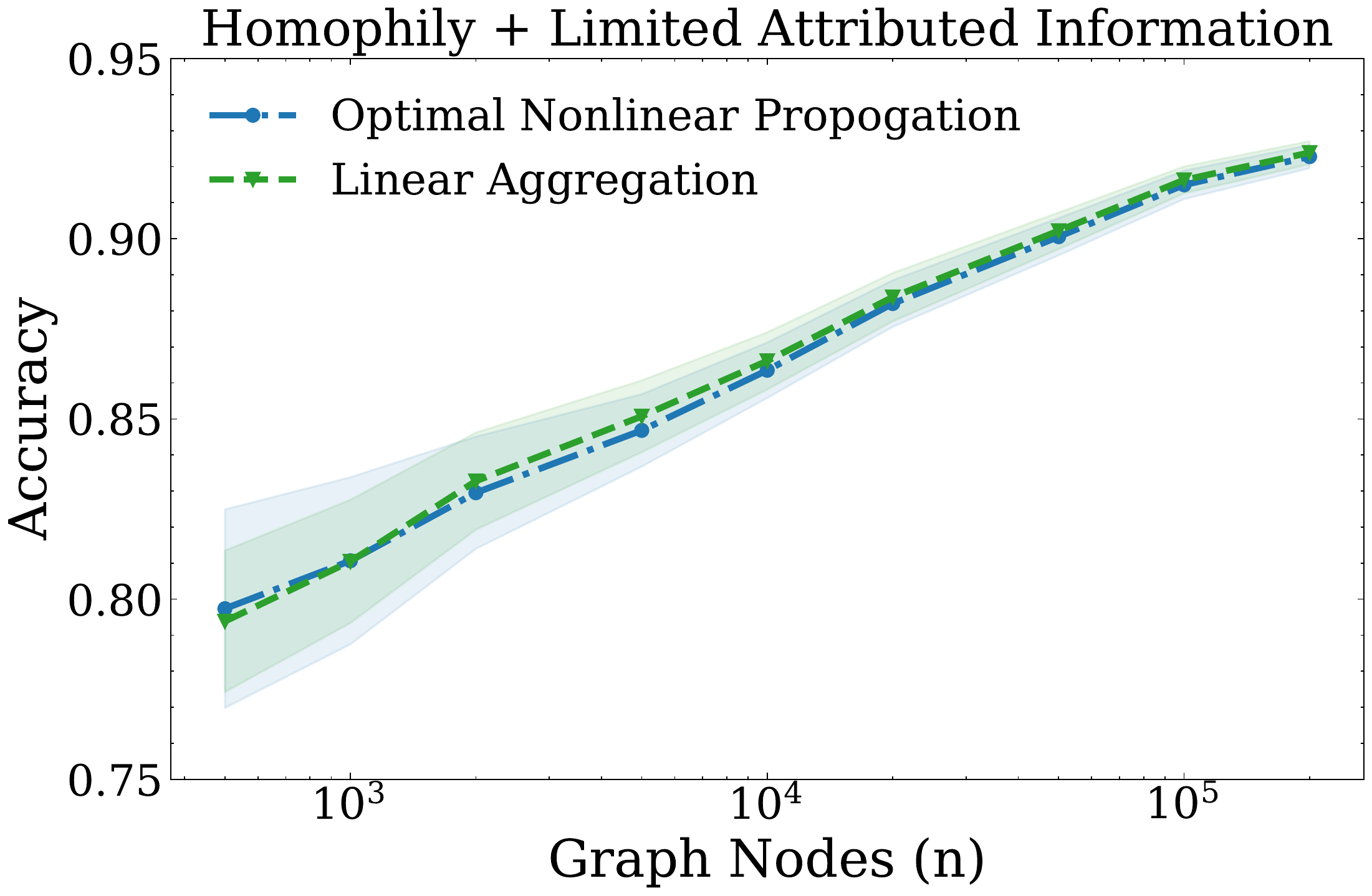}
\includegraphics[width=0.32\textwidth]{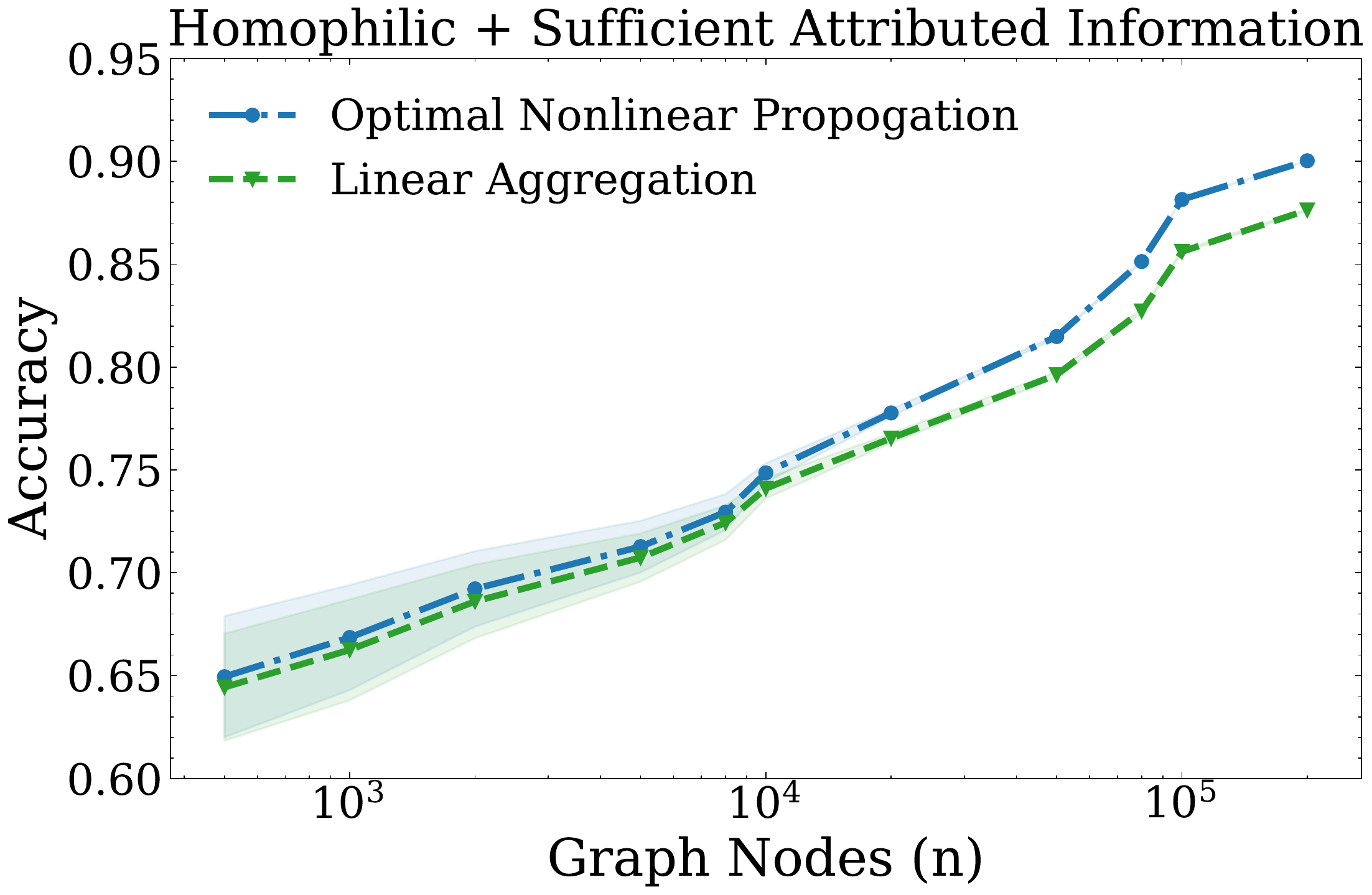}
\hfill\vline\hfill
\includegraphics[width=0.32\textwidth]{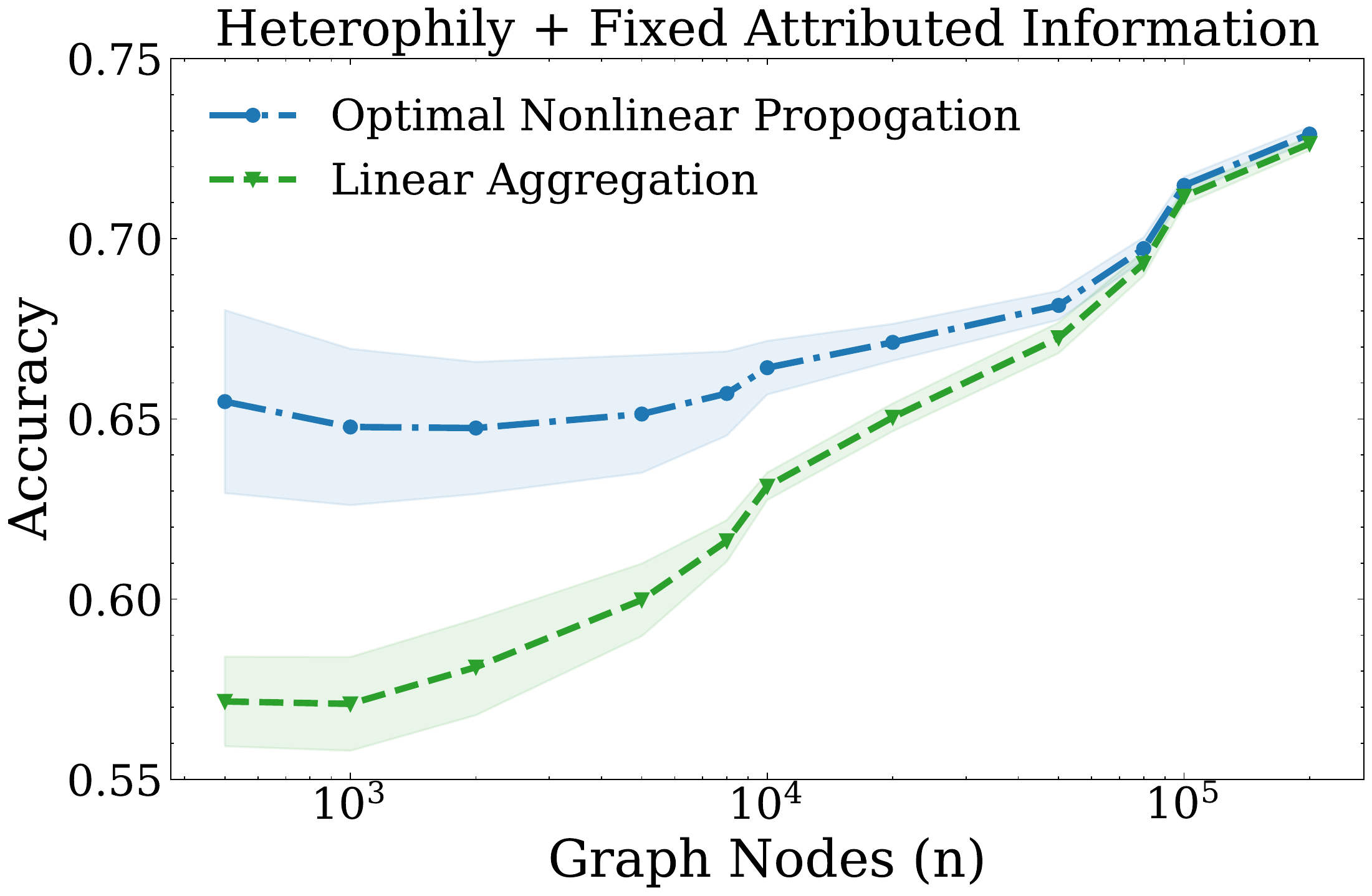}
\small{\caption{Classification Performance on Nonlinear Models v.s. Linear Models ($\mathcal{P}_v$ v.s. $\mathcal{P}_v^l$). LEFT: Homophily + Limited Attr. Info.; 
MIDDLE: Homophily + Suff. Attr. Info.; 
    RIGHT: Heterophily + Fixed Attr. Info.. $m=10$ and other parameters are listed in the figures. 
\label{fig.asym}}}
\end{figure}

\vspace{-1mm}
Our first experiments focus on evaluating the asymptotic $(n\rightarrow \infty)$ classification performance of nonlinear and linear models. 
Given a CSBM-G, we generate 5 graphs and compute the average accuracy results (\#correctly classified nodes / \#total nodes). We compare the nonlinear v.s. linear models under three different CSBM-G settings. Fig.~\ref{fig.asym} shows the results.

All three cases satisfy the separability condition in Theorem~\ref{thm.1}, so, as $n$ increases, the accuracy progressively increases to 1. Our results also match well with the implications provided by Theorem~\ref{thm:2}. In the regime with limited attribute information (Fig.~\ref{as.1} LEFT) where $\rho_r = \Theta_n(\rho_l^*)$ as proved, the nonlinear model and the linear model behave almost the same (performance gap $<0.15\%$ for $n \geq 10^5$). 
In the regime with sufficient attribute information (Fig.~\ref{as.1} MIDDLE) where $\rho_r = \omega_n(\rho_l^*)$ as proved, we may observe that the nonlinear model can significantly outperform the linear model as $n\rightarrow \infty$. Fig.~\ref{as.1} RIGHT is to show the heterophilic graph case ($p<q$). If we switch the values of $p,q$ (and also change the models correspondingly), we obtain the exactly same figure up to some experimental randomness (see Appendix~\ref{app:heter_to_homo}). Also, Fig.~\ref{as.1} RIGHT considers a boundary case of sufficient attributed information, i.e., $\sqrt{m}\|\mu-\nu\|_2 = \Theta_n(1)$. We observe that Theorem~\ref{thm:2} still well describes the asymptotic performance when $n\rightarrow\infty$. 
\vspace{-1mm}

\begin{figure}[t]
\centering
\includegraphics[width=0.32\textwidth]{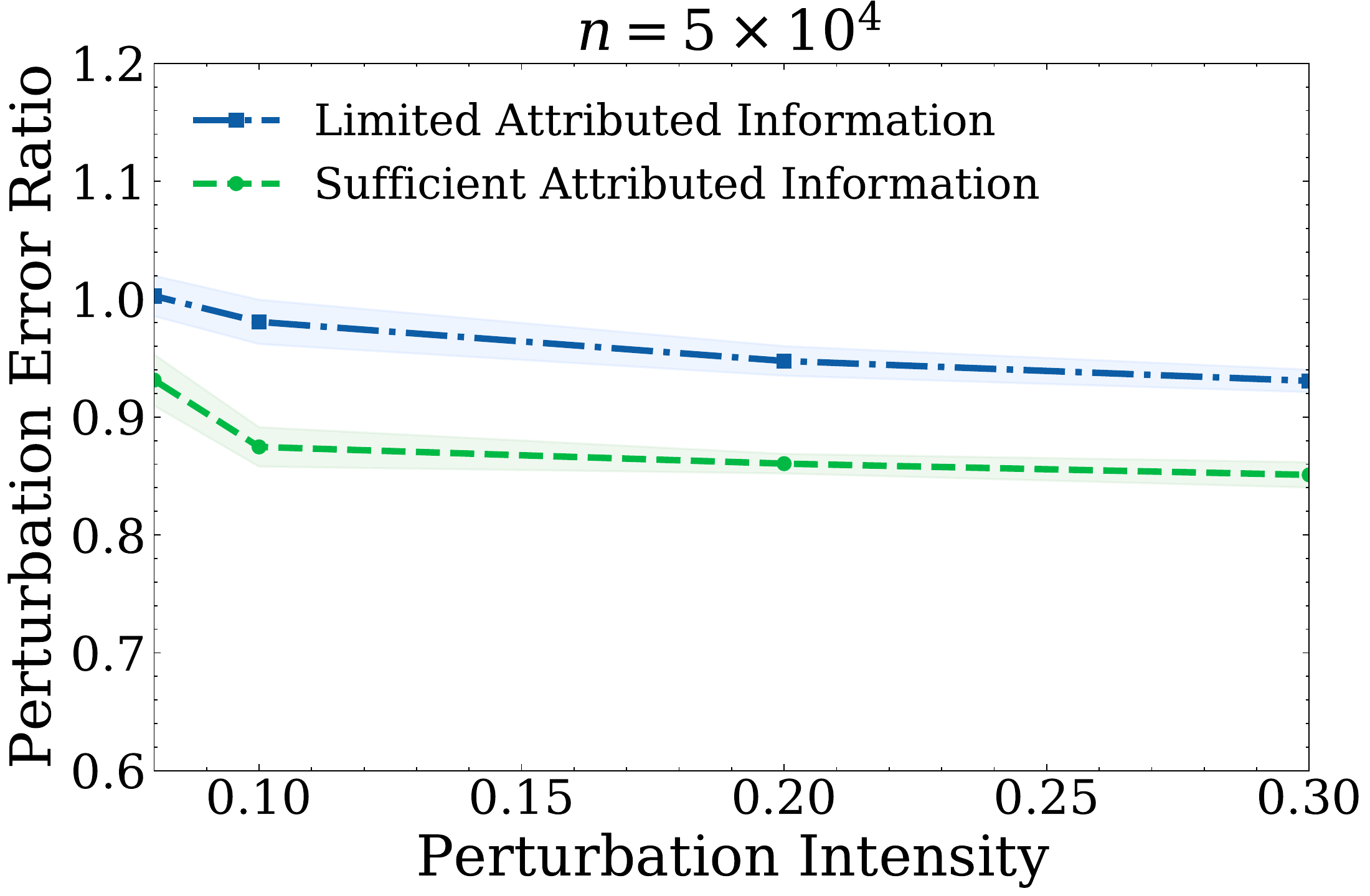}
\centering
\includegraphics[width=0.32\textwidth]{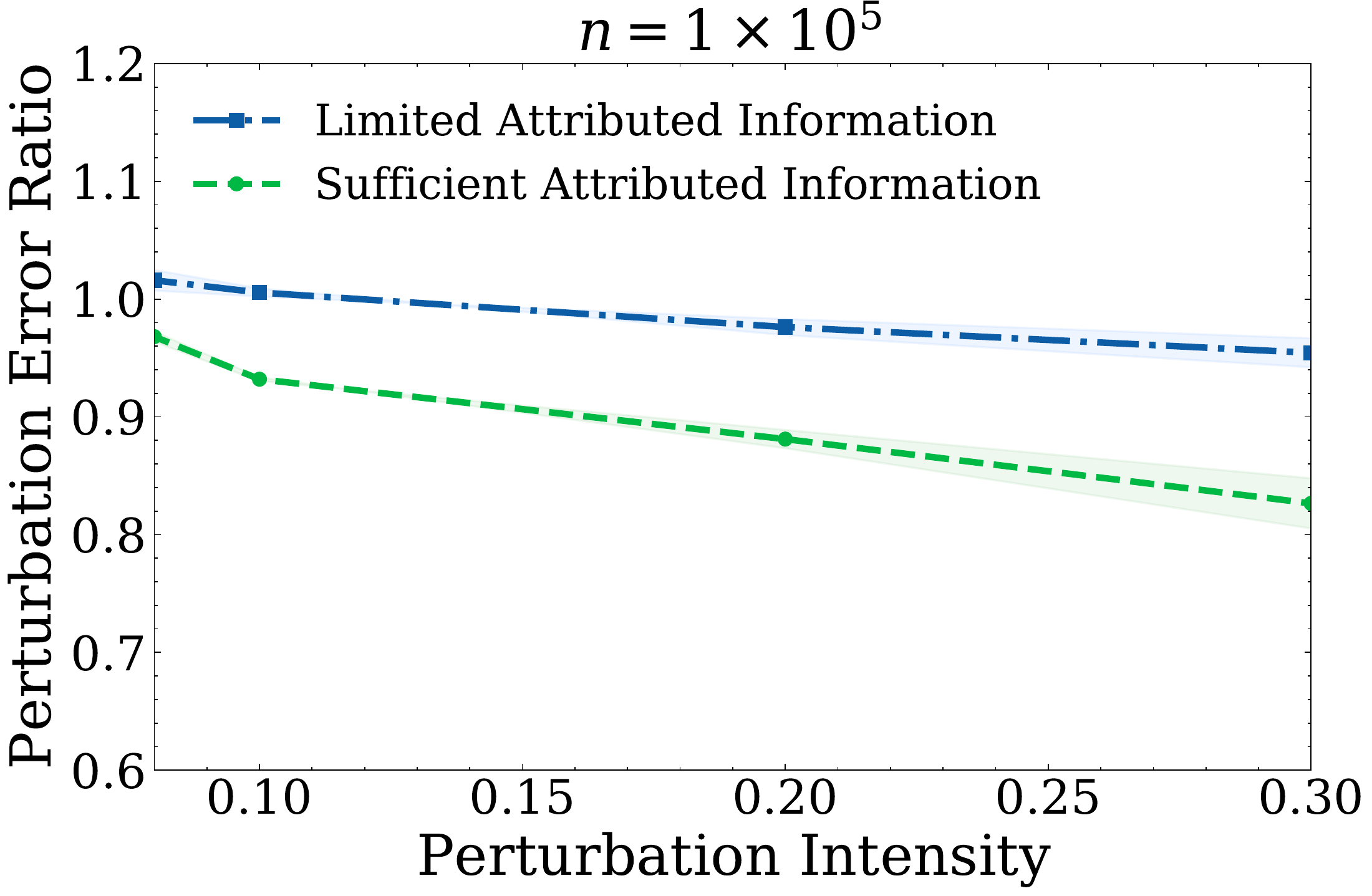}
\centering
\includegraphics[width=0.32\textwidth]{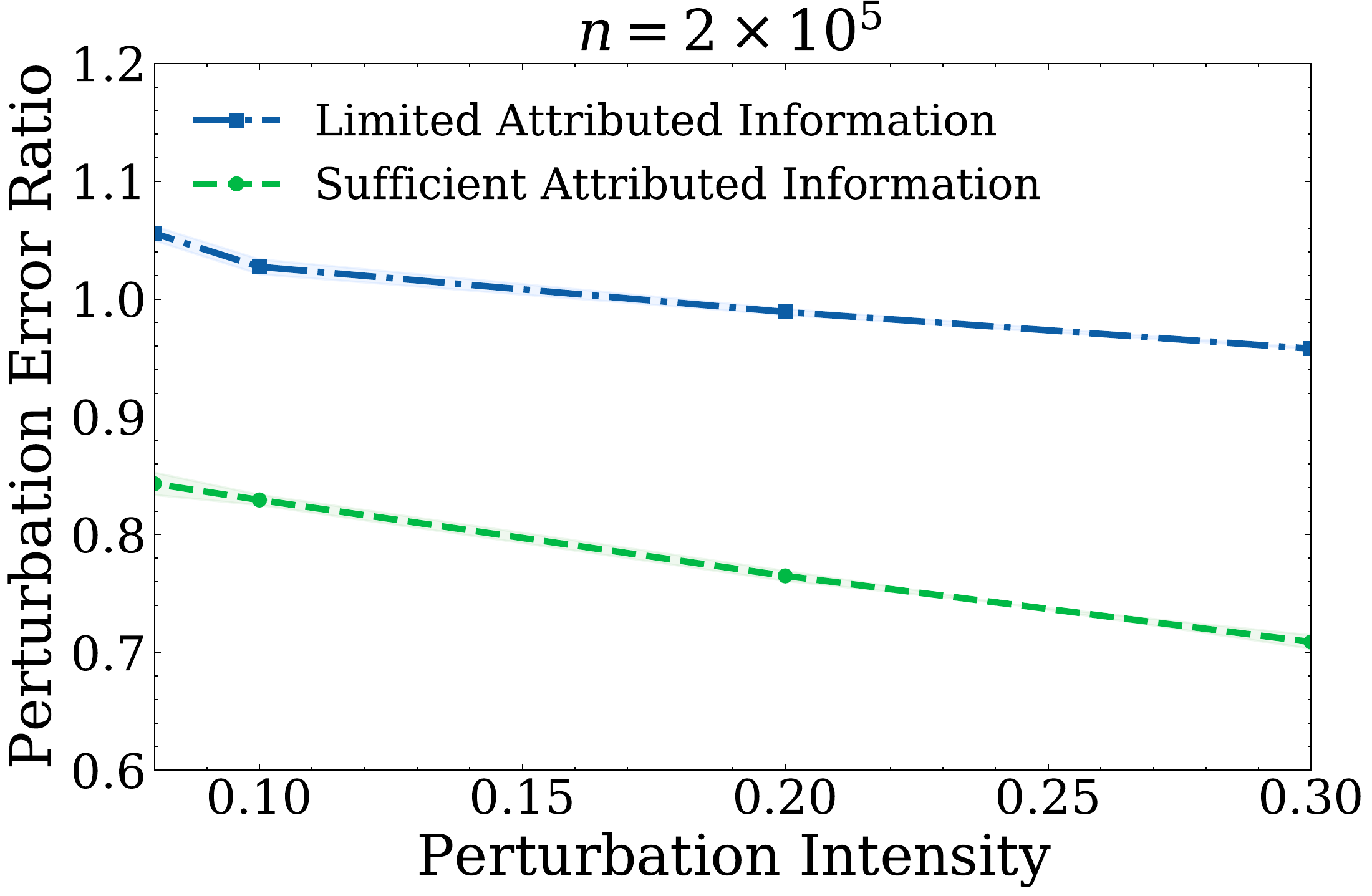}
\small{\caption{Perturbation Intensity ($1 - \langle \mu' - \nu', \mu - \nu \rangle / \|\mu - \nu\|_2^2$) v.s. Perturbation Error Ratio ($\Delta \xi^r / \Delta \xi^l$) with Different Node Numbers. Other parameters are: $p = 2 \sqrt{n} / n, q = \sqrt{n}/n$; Limited Attr. Info. $\|\mu - \nu\|_2 = 0.3\log^2 n / \sqrt{n}$; Suff.  Attr. Info. $\|\mu - \nu\|_2 = 0.1 \sqrt{\log n}$. 
}
\label{fig.perturb}}
\vspace{-2mm}
\end{figure}

We further study the transferability for the non-linear model and the linear model. We follow the setting in Theorem~\ref{thm:3} by rotating $\mu,\nu\rightarrow \mu',\nu'$. Fig.~\ref{fig.perturb} shows the result and well matches Theorem~\ref{thm:3}. In the regime of limited attributed information, the two models have the almost same transferability, i.e., the perturbation error ratio is close to 1. In contrast, with sufficient attributed information, the non-linear model is more transferrable than the linear counterpart as the ratio is smaller than 1. 

\begin{textblock*}{2.5cm}(5.6cm,3.5cm) 
   \fontsize{0.5}{4}
   \flushright
       $p=\frac{2}{\sqrt{n}}$\\
       $q=\frac{1}{\sqrt{n}}$ \\
       $\|\mu-\nu\|_2 = \frac{0.3\log^2n}{\sqrt{n}}$
\end{textblock*}
\begin{textblock*}{2.5cm}(10.2cm,3.5cm) 
   \fontsize{0.5}{4}
   \flushright
       $p=\frac{10}{\sqrt{n}}$\\
       $q=\frac{9}{\sqrt{n}}$ \\
       $\|\mu-\nu\|_2 = \sqrt{\log n}$
\end{textblock*}
\begin{textblock*}{2.5cm}(15.1cm,3.5cm) 
   \fontsize{0.5}{4}
   \flushright
       $p=\frac{9}{\sqrt{n}}$\\
       $q=\frac{10}{\sqrt{n}}$ \\
       $\|\mu-\nu\|_2 = 0.5$
\end{textblock*}


\vspace{-2mm}

\begin{figure}[t]
\includegraphics[height=2.8cm]{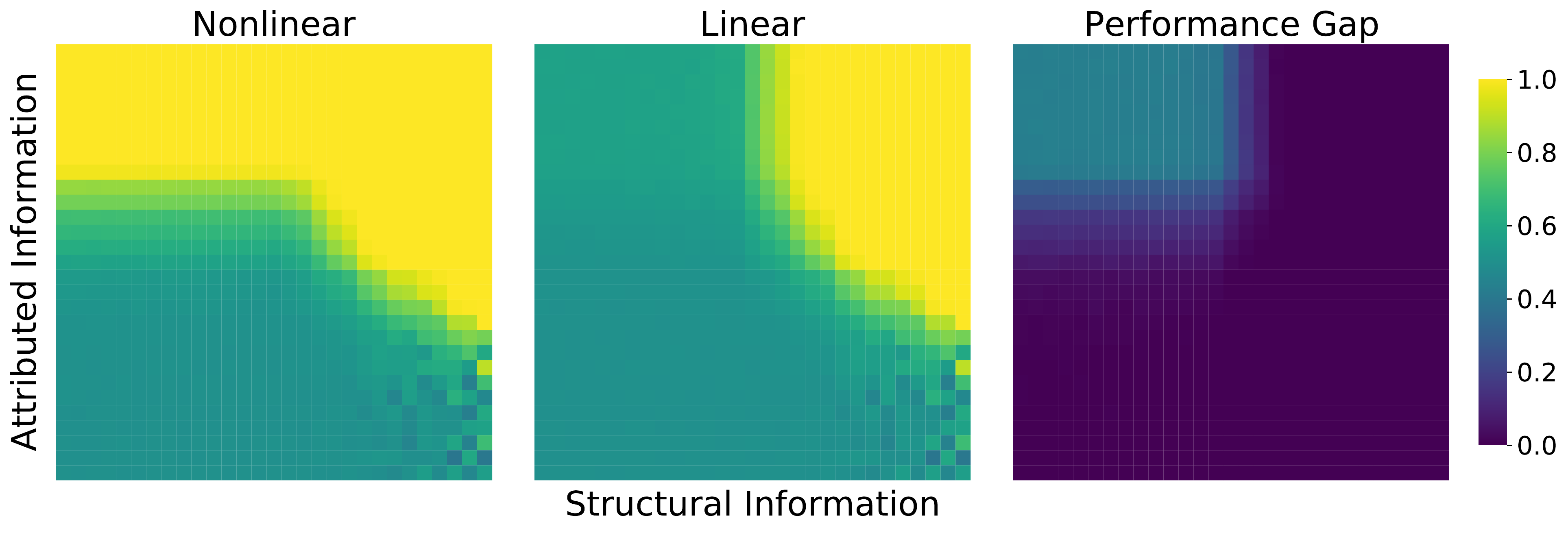}
\hfill\vline\hfill
\includegraphics[height=2.8cm]{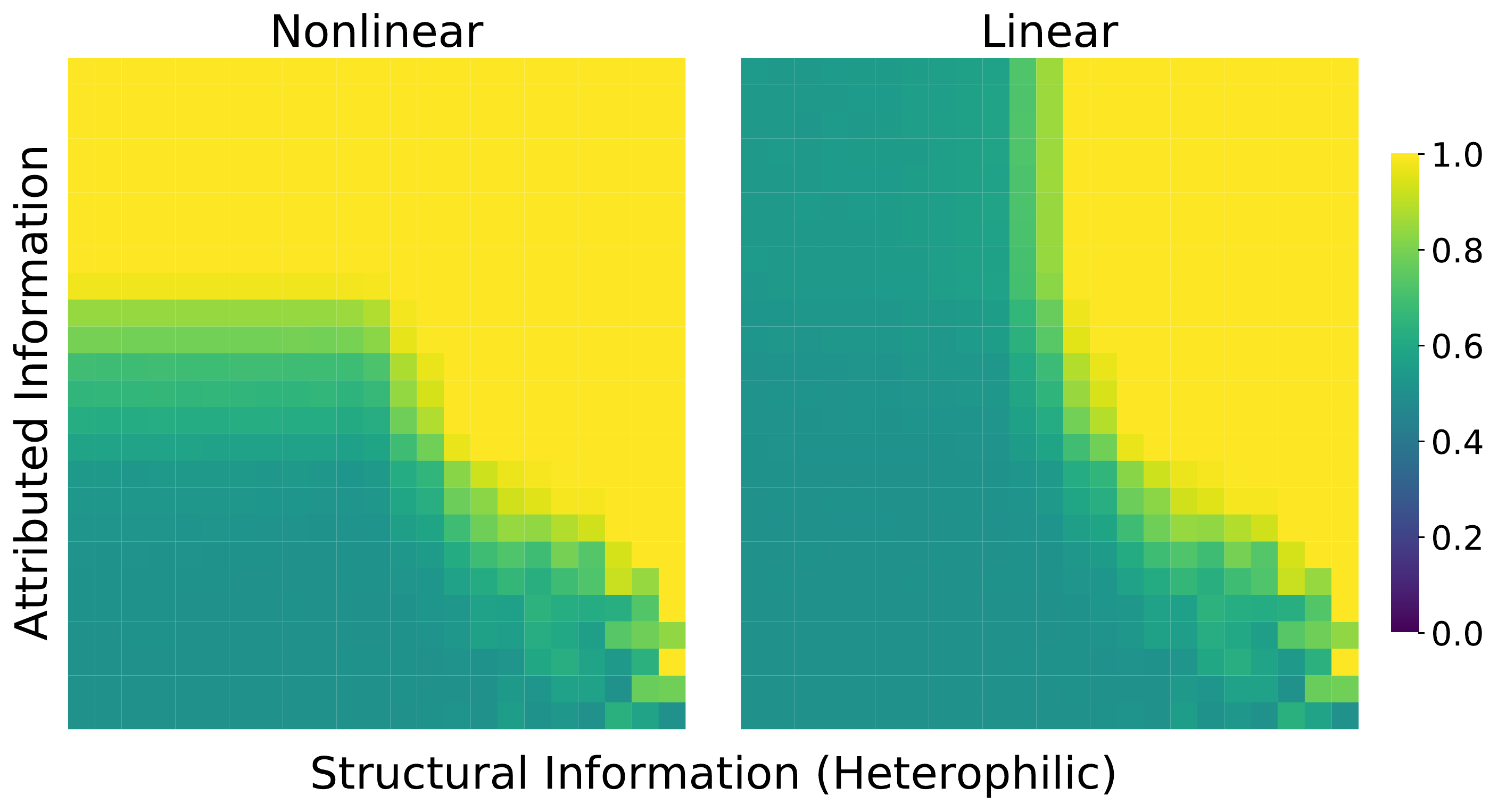}
\small{\caption{Transition Curves Attributed Information ($\sqrt{m}\|\mu - \nu\|_2$) v.s. Structural Information ($|\log (p/q)|$) for CSBM-G with Homophilic (LEFT) / Heterophilic (RIGHT) Graph Structures. The values in Performance Gap are obtained by the nonlinear case subtracting the linear case.
}\label{fig.trans}}
\end{figure}

\subsection{Transition Curve}\label{subsec:trans}
\vspace{-1mm}
Our second experiment studies the tradeoff between attributed information and structural information. We fix the graph size $n = 2 \times 10^{4}$ and get the averaged classification accuracy based on 5 generated graphs. For the homophilic case, we test different levels of attributed information ($\|\mu-\nu\|$ from $10^{-4}$ to $10$ with $m=10$) and structural information (fixing $q = 5\times10^{-3}$ and increasing $p$ from $p = q$ to $1$). The intermediate testing points are sampled in log scales. Fig.~\ref{fig.trans} LEFT shows the results. When structural information is limited and attributed information is sufficient, the non-linear model shows significant advantage over the linear model while for most other parameter settings, these two models share similar performance. Fig.~\ref{fig.trans} RIGHT shows the heterophilic case, where we observe a similar pattern. In the heterophilic case, we fixing $p = 5\times10^{-3}$ and increasing $q$ from $q = p$ to $1$. 




\vspace{-2mm}
\subsection{Real-world Network Experiments}\label{sec:exp-real}
\vspace{-1mm}
\begin{figure}[t]
\label{fig.ova}
\centering
\centering
\centering
\includegraphics[width=0.32\textwidth]{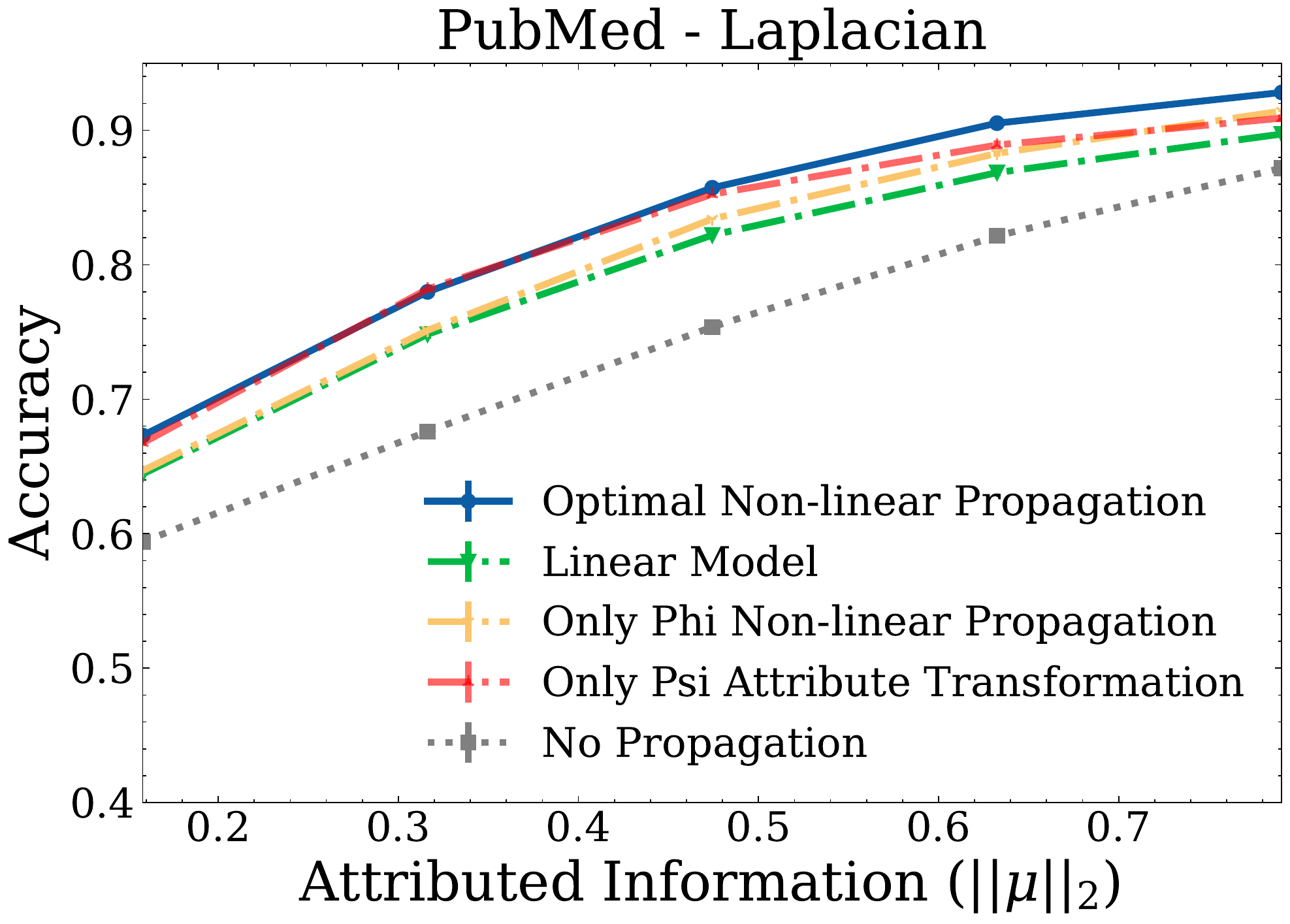}
\centering
\includegraphics[width=0.32\textwidth]{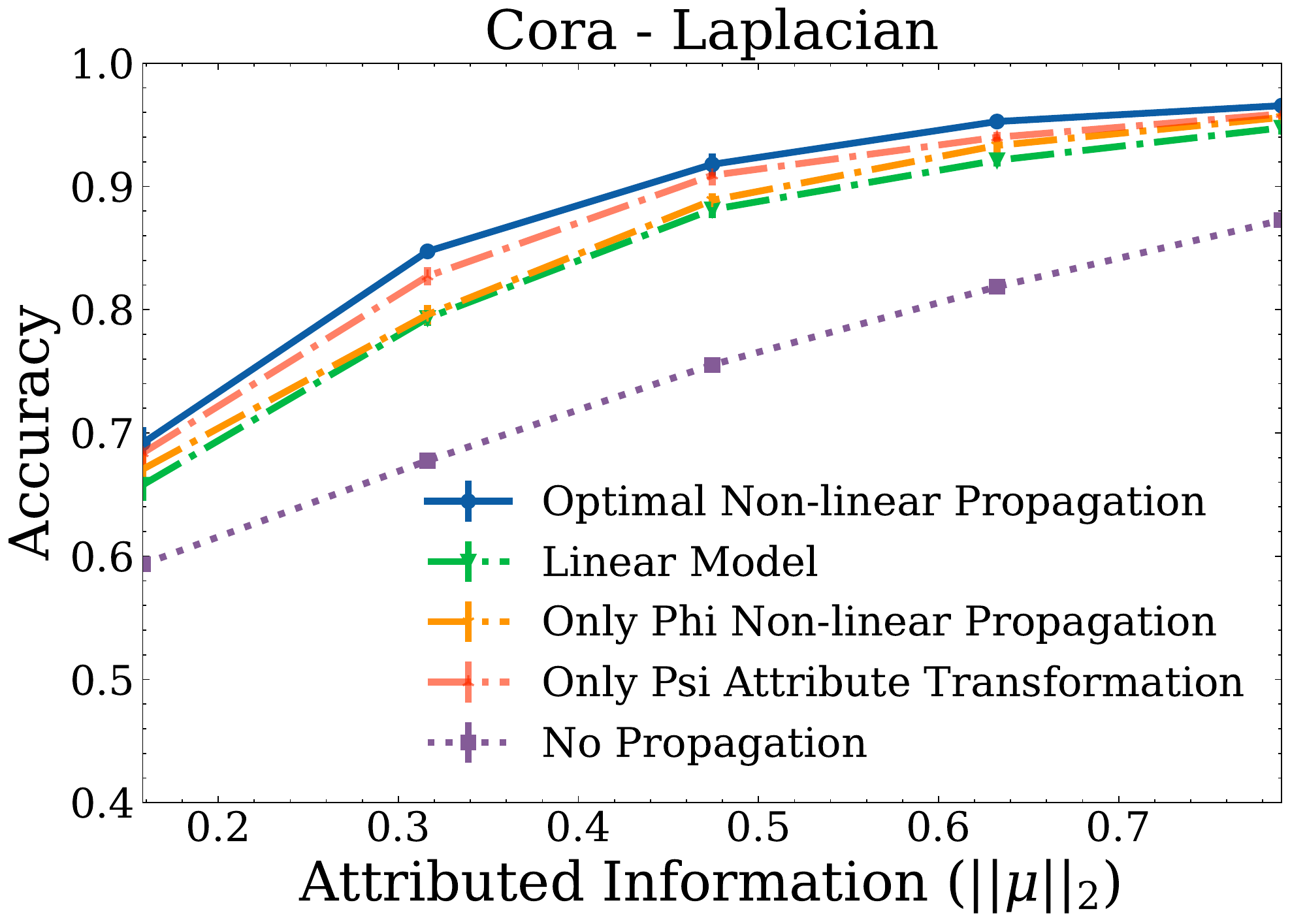}
\centering
\includegraphics[width=0.32\textwidth]{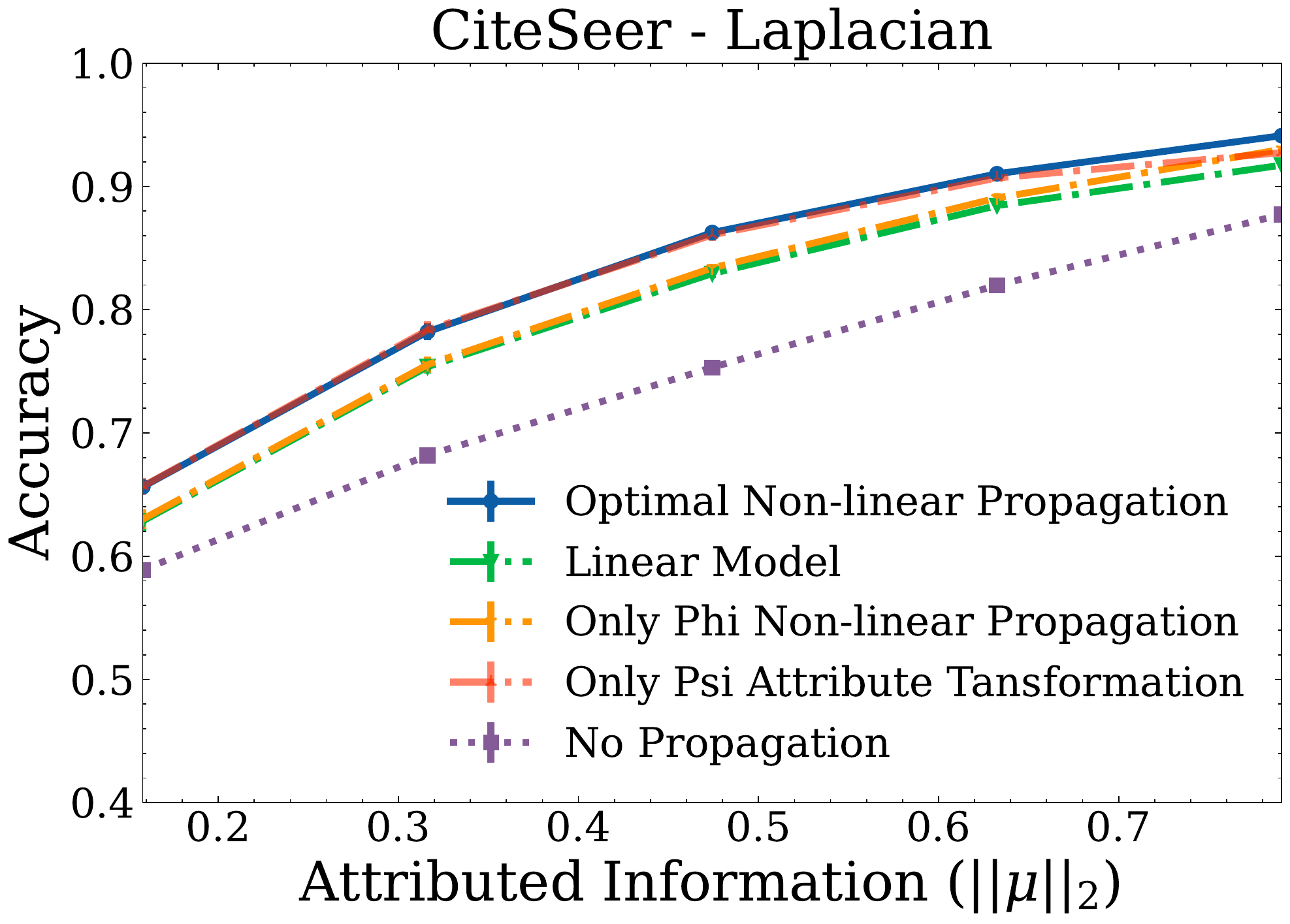}
\small{\caption{Averaged one-vs-all Classification Accuracies on Citation Networks of Different Nonlinear Models v.s. Linear Models. Node attributes in or out of the one class are generated from Laplace distributions with different means $\pm\mu$ and $b=1$ (Example~\ref{eg.lap}). The optimal non-linear model has advantage over the models with only nonlinear attribute transformation ($\psi_{\text{lap}}$), with only nonlinear information propagation ($\phi$), the linear model.
\label{fig.real}
}}
\vspace{-1mm}
\end{figure}

This experiments compare non-linear models and linear models under Gaussian and Laplacian attributes on three benchmark citation networks PubMed, Cora, and CiteSeer ~\cite{sen2008collective}. In these three networks, nodes denote papers and edges denote the citation relationships between the papers. The statistics ($\#$ nodes, $\#$ edges, $\#$ classes) of these three networks are: PubMed (19,717, 44,338, 3); Cora (2,708, 5,428, 7); CiteSeer (3,327, 4,732, 6).


\textbf{Experimental Settings.} We carry out one-v.s.-all and several-v.s.-several classification tasks. After nodes are put into two classes, we generate two graphs independently with attributes according to Gaussian (or Laplace) distributions. One graph is used for training and the other one for testing. For the Gaussian case, we use a nonlinear model by following Eq.~\ref{eq:prop} with $\psi=\psi_{\text{Gau}}$ while the parameters such as $\log(p/q)$, $\mu-\nu$ and other biases need to be learned. For the Laplacian case, we consider three nonlinear models by following the form of (a) full Eq.~\ref{eq:prop} with $\psi=\psi_{\text{lap}}$; (b) only nonlinear attribute transformation $\psi=\psi_{\text{lap}}$; (c) only nonlinear propagation $\phi$ with linear attribute transformation. Later, we call them nonlinear models (a), (b), (c), respectively. Similar to the Gaussian case, all the parameters in these functions are obtained by training. The model is trained with Adam optimizer (learning rate = $1e-2$, weight decay = $5e-4$). We give other details to Appendix~\ref{app:exp_OVA}.





\textbf{Result Analysis.} We report the averaged results over 5 trials in Fig.~\ref{fig.ova-gaussian} (Gaussian) and Fig.~\ref{fig.real} (Laplacian). Due to the space limit, we leave the results for the several-v.s.-several case in Appendix~\ref{app:exp_OVA}. The Gaussian case well matches our theory. Only when the node features are very informative, the gaps between the nonlinear model and the linear model become significant. This is true for all three networks. 

The Laplacian case is more complicated. Non-linear model (a) outperforms the two non-linear models (b) and (c). The two non-linear models both outperform the linear model. More specifically, when attributed information is not very informative, i.e., small $\|\mu\|_2$, attribute nonlinear transformation function $\psi_{\text{Lap}}$ is more crucial, because in this regime, non-linear model (a) significantly outperforms non-linear model (c) and non-linear model (b) significantly outperforms the linear model, while two non-linear models (a) and (b) perform similarly, and non-linear model (c) and the linear model perform similarly. With more informative attributed information, nonlinear propagation function $\phi$ becomes more significant, because the gaps between two non-linear models (a) and (b) (also, non-linear model (c) and the linear model) are obvious, which again matches our Theorem~\ref{thm:2} although here we have Laplacian node attributes instead of Gaussian node attributes.   

\vspace{-2mm}
\section{Conclusion}\label{sec:con}
\vspace{-2mm}
This work uses Bayesian methods to investigate the function of non-linearity in GNNs. Given a graph generated from CSBM, we observe the optimal non-linearity to estimate a node label given its own and neighbors' attributes is in twofold: attribute  non-linear transformation and non-linear propagation. We further investigate the non-linear propagation by imposing Gaussian assumptions on node attributes. We prove that non-linear propagation shares a similar performance (with or without distribution shift) with linear propagation in most cases except when node attributes become very informative. These findings explain many previous empirical observations in this domain and would help researchers and practitioners to understand their GNNs’ behaviors in practice.

\vspace{-2mm}
\section{Acknowledgement}\label{sec:con}
\vspace{-2mm}

We greatly thank all the reviewers for valuable feedback and actionable suggestions. R. Wei, H. Yin and P. Li are partially supported by 2021 JPMorgan Faculty Award and NSF award OAC-2117997.

\bibliographystyle{ACM}
\bibliography{reference}

\begin{thebibliography}{10}
\providecommand{\url}[1]{#1}
\csname url@samestyle\endcsname
\providecommand{\newblock}{\relax}
\providecommand{\bibinfo}[2]{#2}
\providecommand{\BIBentrySTDinterwordspacing}{\spaceskip=0pt\relax}
\providecommand{\BIBentryALTinterwordstretchfactor}{4}
\providecommand{\BIBentryALTinterwordspacing}{\spaceskip=\fontdimen2\font plus
\BIBentryALTinterwordstretchfactor\fontdimen3\font minus
  \fontdimen4\font\relax}
\providecommand{\BIBforeignlanguage}[2]{{%
\expandafter\ifx\csname l@#1\endcsname\relax
\typeout{** WARNING: IEEEtran.bst: No hyphenation pattern has been}%
\typeout{** loaded for the language `#1'. Using the pattern for}%
\typeout{** the default language instead.}%
\else
\language=\csname l@#1\endcsname
\fi
#2}}
\providecommand{\BIBdecl}{\relax}
\BIBdecl

\bibitem{zhu2005semi}
X.~Zhu, \emph{Semi-supervised learning with graphs}.\hskip 1em plus 0.5em minus
  0.4em\relax Carnegie Mellon University, 2005.

\bibitem{GRLbook}
W.~L. Hamilton, ``Graph representation learning,'' \emph{Synthesis Lectures on
  Artificial Intelligence and Machine Learning}, vol.~14, no.~3, pp. 1--159.

\bibitem{fortunato2010community}
S.~Fortunato, ``Community detection in graphs,'' \emph{Physics reports}, vol.
  486, no. 3-5, pp. 75--174, 2010.

\bibitem{lancichinetti2009community}
A.~Lancichinetti and S.~Fortunato, ``Community detection algorithms: a
  comparative analysis,'' \emph{Physical review E}, vol.~80, no.~5, p. 056117,
  2009.

\bibitem{chen2020supervised}
Z.~Chen, L.~Li, and J.~Bruna, ``Supervised community detection with line graph
  neural networks,'' in \emph{International Conference on Learning
  Representations}, 2020.

\bibitem{liu2021deep}
F.~Liu, S.~Xue, J.~Wu, C.~Zhou, W.~Hu, C.~Paris, S.~Nepal, J.~Yang, and P.~S.
  Yu, ``Deep learning for community detection: progress, challenges and
  opportunities,'' in \emph{Proceedings of the Twenty-Ninth International Joint
  Conferences on Artificial Intelligence}, 2021, pp. 4981--4987.

\bibitem{ma2021comprehensive}
X.~Ma, J.~Wu, S.~Xue, J.~Yang, C.~Zhou, Q.~Z. Sheng, H.~Xiong, and L.~Akoglu,
  ``A comprehensive survey on graph anomaly detection with deep learning,''
  \emph{IEEE Transactions on Knowledge and Data Engineering}, 2021.

\bibitem{wang2021bipartite}
A.~Z. Wang, R.~Ying, P.~Li, N.~Rao, K.~Subbian, and J.~Leskovec, ``Bipartite
  dynamic representations for abuse detection,'' in \emph{Proceedings of the
  27th ACM SIGKDD Conference on Knowledge Discovery \& Data Mining}, 2021, pp.
  3638--3648.

\bibitem{aittokallio2006graph}
T.~Aittokallio and B.~Schwikowski, ``Graph-based methods for analysing networks
  in cell biology,'' \emph{Briefings in bioinformatics}, vol.~7, no.~3, pp.
  243--255, 2006.

\bibitem{scott2006efficient}
J.~Scott, T.~Ideker, R.~M. Karp, and R.~Sharan, ``Efficient algorithms for
  detecting signaling pathways in protein interaction networks,'' \emph{Journal
  of Computational Biology}, vol.~13, no.~2, pp. 133--144, 2006.

\bibitem{hamilton2017inductive}
W.~Hamilton, Z.~Ying, and J.~Leskovec, ``Inductive representation learning on
  large graphs,'' in \emph{Advances in Neural Information Processing Systems},
  vol.~30, 2017.

\bibitem{kipf2016semi}
T.~N. Kipf and M.~Welling, ``Semi-supervised classification with graph
  convolutional networks,'' in \emph{International Conference on Learning
  Representations}, 2017.

\bibitem{xu2018powerful}
K.~Xu, W.~Hu, J.~Leskovec, and S.~Jegelka, ``How powerful are graph neural
  networks?'' in \emph{International Conference on Learning Representations},
  2019.

\bibitem{morris2019weisfeiler}
C.~Morris, M.~Ritzert, M.~Fey, W.~L. Hamilton, J.~E. Lenssen, G.~Rattan, and
  M.~Grohe, ``Weisfeiler and leman go neural: Higher-order graph neural
  networks,'' in \emph{Proceedings of the AAAI conference on artificial
  intelligence}, vol.~33, 2019, pp. 4602--4609.

\bibitem{keriven2019universal}
N.~Keriven and G.~Peyr{\'e}, ``Universal invariant and equivariant graph neural
  networks,'' \emph{Advances in Neural Information Processing Systems},
  vol.~32, 2019.

\bibitem{chen2019equivalence}
Z.~Chen, S.~Villar, L.~Chen, and J.~Bruna, ``On the equivalence between graph
  isomorphism testing and function approximation with gnns,'' \emph{Advances in
  Neural Information Processing Systems}, vol.~32, 2019.

\bibitem{azizian2021expressive}
W.~Azizian \emph{et~al.}, ``Expressive power of invariant and equivariant graph
  neural networks,'' in \emph{International Conference on Learning
  Representations}, 2021.

\bibitem{gilmer2017neural}
J.~Gilmer, S.~S. Schoenholz, P.~F. Riley, O.~Vinyals, and G.~E. Dahl, ``Neural
  message passing for quantum chemistry,'' in \emph{International Conference on
  Machine Learning}.\hskip 1em plus 0.5em minus 0.4em\relax PMLR, 2017, pp.
  1263--1272.

\bibitem{wu2019simplifying}
F.~Wu, A.~Souza, T.~Zhang, C.~Fifty, T.~Yu, and K.~Weinberger, ``Simplifying
  graph convolutional networks,'' in \emph{International Conference on Machine
  Learning}.\hskip 1em plus 0.5em minus 0.4em\relax PMLR, 2019, pp. 6861--6871.

\bibitem{he2020lightgcn}
X.~He, K.~Deng, X.~Wang, Y.~Li, Y.~Zhang, and M.~Wang, ``Lightgcn: Simplifying
  and powering graph convolution network for recommendation,'' in
  \emph{Proceedings of the 43rd International ACM SIGIR conference on research
  and development in Information Retrieval}, 2020, pp. 639--648.

\bibitem{klicpera2018predict}
J.~Klicpera, A.~Bojchevski, and S.~G{\"u}nnemann, ``Predict then propagate:
  Graph neural networks meet personalized pagerank,'' in \emph{International
  Conference on Learning Representations}, 2018.

\bibitem{DBLP:conf/nips/KlicperaWG19}
J.~Klicpera, S.~Wei{\ss}enberger, and S.~G{\"{u}}nnemann, ``Diffusion improves
  graph learning,'' in \emph{Advances in Neural Information Processing
  Systems}, vol.~32, 2019.

\bibitem{huang2020combining}
Q.~Huang, H.~He, A.~Singh, S.-N. Lim, and A.~Benson, ``Combining label
  propagation and simple models out-performs graph neural networks,'' in
  \emph{International Conference on Learning Representations}, 2020.

\bibitem{velivckovic2018graph}
P.~Veli{\v{c}}kovi{\'c}, G.~Cucurull, A.~Casanova, A.~Romero, P.~Li{\`o}, and
  Y.~Bengio, ``Graph attention networks,'' in \emph{International Conference on
  Learning Representations}, 2018.

\bibitem{chien2021adaptive}
E.~Chien, J.~Peng, P.~Li, and O.~Milenkovic, ``Adaptive universal generalized
  pagerank graph neural network,'' in \emph{International Conference on
  Learning Representations}, 2021.

\bibitem{wang2022powerful}
X.~Wang and M.~Zhang, ``How powerful are spectral graph neural networks,'' in
  \emph{International Conference on Machine Learning}.\hskip 1em plus 0.5em
  minus 0.4em\relax PMLR, 2022.

\bibitem{zhu2020beyond}
J.~Zhu, Y.~Yan, L.~Zhao, M.~Heimann, L.~Akoglu, and D.~Koutra, ``Beyond
  homophily in graph neural networks: Current limitations and effective
  designs,'' \emph{Advances in Neural Information Processing Systems}, vol.~33,
  2020.

\bibitem{zhu2021graph}
J.~Zhu, R.~A. Rossi, A.~B. Rao, T.~Mai, N.~Lipka, N.~K. Ahmed, and D.~Koutra,
  ``Graph neural networks with heterophily,'' in \emph{Proceedings of the AAAI
  Conference on Artificial Intelligence}, 2021.

\bibitem{chen2020graph}
L.~Chen, Z.~Chen, and J.~Bruna, ``On graph neural networks versus
  graph-augmented mlps,'' in \emph{International Conference on Learning
  Representations}, 2021.

\bibitem{cong2021provable}
W.~Cong, M.~Ramezani, and M.~Mahdavi, ``On provable benefits of depth in
  training graph convolutional networks,'' \emph{Advances in Neural Information
  Processing Systems}, vol.~34, 2021.

\bibitem{li2022expressive}
P.~Li and J.~Leskovec, ``The expressive power of graph neural networks,'' in
  \emph{Graph Neural Networks: Foundations, Frontiers, and Applications}.\hskip
  1em plus 0.5em minus 0.4em\relax Springer, 2022, pp. 63--98.

\bibitem{binkiewicz2017covariate}
N.~Binkiewicz, J.~T. Vogelstein, and K.~Rohe, ``Covariate-assisted spectral
  clustering,'' \emph{Biometrika}, vol. 104, no.~2, pp. 361--377, 2017.

\bibitem{deshpande2018contextual}
Y.~Deshpande, S.~Sen, A.~Montanari, and E.~Mossel, ``Contextual stochastic
  block models,'' \emph{Advances in Neural Information Processing Systems},
  vol.~31, 2018.

\bibitem{morris1982natural}
C.~N. Morris, ``Natural exponential families with quadratic variance
  functions,'' \emph{The Annals of Statistics}, pp. 65--80, 1982.

\bibitem{theodoridis2015machine}
S.~Theodoridis, \emph{Machine learning: a Bayesian and optimization
  perspective}.\hskip 1em plus 0.5em minus 0.4em\relax Academic press, 2015.

\bibitem{suresh2021breaking}
S.~Suresh, V.~Budde, J.~Neville, P.~Li, and J.~Ma, ``Breaking the limit of
  graph neural networks by improving the assortativity of graphs with local
  mixing patterns,'' in \emph{Proceedings of the 27th ACM SIGKDD Conference on
  Knowledge Discovery \& Data Mining}, 2021, pp. 1541--1551.

\bibitem{ma2021homophily}
Y.~Ma, X.~Liu, N.~Shah, and J.~Tang, ``Is homophily a necessity for graph
  neural networks?'' \emph{arXiv preprint arXiv:2106.06134}, 2021.

\bibitem{maron2019provably}
H.~Maron, H.~Ben-Hamu, H.~Serviansky, and Y.~Lipman, ``Provably powerful graph
  networks,'' \emph{Advances in neural information processing systems},
  vol.~32, 2019.

\bibitem{balcilar2021breaking}
M.~Balcilar, P.~H{\'e}roux, B.~Gauzere, P.~Vasseur, S.~Adam, and P.~Honeine,
  ``Breaking the limits of message passing graph neural networks,'' in
  \emph{International Conference on Machine Learning}.\hskip 1em plus 0.5em
  minus 0.4em\relax PMLR, 2021, pp. 599--608.

\bibitem{azizian2020expressive}
W.~Azizian \emph{et~al.}, ``Expressive power of invariant and equivariant graph
  neural networks,'' in \emph{International Conference on Learning
  Representations}, 2021.

\bibitem{murphy2019relational}
R.~Murphy, B.~Srinivasan, V.~Rao, and B.~Ribeiro, ``Relational pooling for
  graph representations,'' in \emph{International Conference on Machine
  Learning}.\hskip 1em plus 0.5em minus 0.4em\relax PMLR, 2019, pp. 4663--4673.

\bibitem{sato2021random}
R.~Sato, M.~Yamada, and H.~Kashima, ``Random features strengthen graph neural
  networks,'' in \emph{Proceedings of the 2021 SIAM International Conference on
  Data Mining (SDM)}.\hskip 1em plus 0.5em minus 0.4em\relax SIAM, 2021, pp.
  333--341.

\bibitem{abboud2021surprising}
R.~Abboud, I.~I. Ceylan, M.~Grohe, and T.~Lukasiewicz, ``The surprising power
  of graph neural networks with random node initialization,'' in
  \emph{Proceedings of the Thirtieth International Joint Conference on
  Artificial Intelligence}, 2021, pp. 2112--2118.

\bibitem{bodnar2021weisfeiler}
C.~Bodnar, F.~Frasca, N.~Otter, Y.~Wang, P.~Lio, G.~F. Montufar, and
  M.~Bronstein, ``Weisfeiler and lehman go cellular: Cw networks,''
  \emph{Advances in Neural Information Processing Systems}, vol.~34, 2021.

\bibitem{li2020distance}
P.~Li, Y.~Wang, H.~Wang, and J.~Leskovec, ``Distance encoding: Design provably
  more powerful neural networks for graph representation learning,''
  \emph{Advances in Neural Information Processing Systems}, vol.~33, pp.
  4465--4478, 2020.

\bibitem{loukas2020graph}
A.~Loukas, ``What graph neural networks cannot learn: depth vs width,'' in
  \emph{International Conference on Learning Representations}, 2020.

\bibitem{vignac2020building}
C.~Vignac, A.~Loukas, and P.~Frossard, ``Building powerful and equivariant
  graph neural networks with structural message-passing,'' \emph{Advances in
  Neural Information Processing Systems}, vol.~33, 2020.

\bibitem{zhang2021nested}
M.~Zhang and P.~Li, ``Nested graph neural networks,'' \emph{Advances in Neural
  Information Processing Systems}, vol.~34, 2021.

\bibitem{oono2019graph}
K.~Oono and T.~Suzuki, ``Graph neural networks exponentially lose expressive
  power for node classification,'' in \emph{International Conference on
  Learning Representations}, 2019.

\bibitem{li2018deeper}
Q.~Li, Z.~Han, and X.-M. Wu, ``Deeper insights into graph convolutional
  networks for semi-supervised learning,'' in \emph{Proceedings of the AAAI
  Conference on Artificial Intelligence}, 2018.

\bibitem{alon2020bottleneck}
U.~Alon and E.~Yahav, ``On the bottleneck of graph neural networks and its
  practical implications,'' in \emph{International Conference on Learning
  Representations}, 2021.

\bibitem{topping2021understanding}
J.~Topping, F.~Di~Giovanni, B.~P. Chamberlain, X.~Dong, and M.~M. Bronstein,
  ``Understanding over-squashing and bottlenecks on graphs via curvature,''
  \emph{arXiv preprint arXiv:2111.14522}, 2021.

\bibitem{yan2021two}
Y.~Yan, M.~Hashemi, K.~Swersky, Y.~Yang, and D.~Koutra, ``Two sides of the same
  coin: Heterophily and oversmoothing in graph convolutional neural networks,''
  \emph{arXiv preprint arXiv:2102.06462}, 2021.

\bibitem{balcilar2020analyzing}
M.~Balcilar, G.~Renton, P.~H{\'e}roux, B.~Ga{\"u}z{\`e}re, S.~Adam, and
  P.~Honeine, ``Analyzing the expressive power of graph neural networks in a
  spectral perspective,'' in \emph{International Conference on Learning
  Representations}, 2020.

\bibitem{bianchi2021graph}
F.~M. Bianchi, D.~Grattarola, L.~Livi, and C.~Alippi, ``Graph neural networks
  with convolutional arma filters,'' \emph{IEEE Transactions on Pattern
  Analysis and Machine Intelligence}, 2021.

\bibitem{du2019graph}
S.~S. Du, K.~Hou, R.~R. Salakhutdinov, B.~Poczos, R.~Wang, and K.~Xu, ``Graph
  neural tangent kernel: Fusing graph neural networks with graph kernels,''
  \emph{Advances in neural information processing systems}, vol.~32, 2019.

\bibitem{garg2020generalization}
V.~Garg, S.~Jegelka, and T.~Jaakkola, ``Generalization and representational
  limits of graph neural networks,'' in \emph{International Conference on
  Machine Learning}.\hskip 1em plus 0.5em minus 0.4em\relax PMLR, 2020, pp.
  3419--3430.

\bibitem{liao2020pac}
R.~Liao, R.~Urtasun, and R.~Zemel, ``A pac-bayesian approach to generalization
  bounds for graph neural networks,'' in \emph{International Conference on
  Learning Representations}, 2020.

\bibitem{gama2019diffusion}
F.~Gama, J.~Bruna, and A.~Ribeiro, ``Diffusion scattering transforms on
  graphs,'' in \emph{International Conference on Learning Representations},
  2019.

\bibitem{gama2019stability}
F.~Gama, A.~Ribeiro, and J.~Bruna, ``Stability of graph scattering
  transforms,'' \emph{Advances in Neural Information Processing Systems},
  vol.~32, 2019.

\bibitem{levie2021transferability}
R.~Levie, W.~Huang, L.~Bucci, M.~Bronstein, and G.~Kutyniok, ``Transferability
  of spectral graph convolutional neural networks,'' \emph{Journal of Machine
  Learning Research}, vol.~22, no. 272, pp. 1--59, 2021.

\bibitem{gama2020stability}
F.~Gama, J.~Bruna, and A.~Ribeiro, ``Stability properties of graph neural
  networks,'' \emph{IEEE Transactions on Signal Processing}, vol.~68, pp.
  5680--5695, 2020.

\bibitem{abbe2017community}
E.~Abbe, ``Community detection and stochastic block models: recent
  developments,'' \emph{The Journal of Machine Learning Research}, vol.~18,
  no.~1, pp. 6446--6531, 2017.

\bibitem{abbe2018proof}
E.~Abbe and C.~Sandon, ``Proof of the achievability conjectures for the general
  stochastic block model,'' \emph{Communications on Pure and Applied
  Mathematics}, vol.~71, no.~7, pp. 1334--1406, 2018.

\bibitem{massoulie2014community}
L.~Massouli{\'e}, ``Community detection thresholds and the weak ramanujan
  property,'' in \emph{Proceedings of the forty-sixth annual ACM symposium on
  Theory of computing}, 2014, pp. 694--703.

\bibitem{bordenave2018nonbacktracking}
C.~Bordenave, M.~Lelarge, and L.~Massouli{\'e}, ``Nonbacktracking spectrum of
  random graphs: Community detection and nonregular ramanujan graphs,''
  \emph{Annals of Probability}, vol.~46, no.~1, pp. 1--71, 2018.

\bibitem{montanari2016semidefinite}
A.~Montanari and S.~Sen, ``Semidefinite programs on sparse random graphs and
  their application to community detection,'' in \emph{Proceedings of the
  forty-eighth annual ACM symposium on Theory of Computing}, 2016, pp.
  814--827.

\bibitem{keriven2020convergence}
N.~Keriven, A.~Bietti, and S.~Vaiter, ``Convergence and stability of graph
  convolutional networks on large random graphs,'' \emph{Advances in Neural
  Information Processing Systems}, vol.~33, 2020.

\bibitem{keriven2021universality}
N.Keriven, A.~Bietti, and S.~Vaiter, ``On the universality of graph neural
  networks on large random graphs,'' \emph{Advances in Neural Information
  Processing Systems}, vol.~34, 2021.

\bibitem{ruiz2020graphon}
L.~Ruiz, L.~Chamon, and A.~Ribeiro, ``Graphon neural networks and the
  transferability of graph neural networks,'' \emph{Advances in Neural
  Information Processing Systems}, vol.~33, 2020.

\bibitem{baranwal2021graph}
A.~Baranwal, K.~Fountoulakis, and A.~Jagannath, ``Graph convolution for
  semi-supervised classification: Improved linear separability and
  out-of-distribution generalization,'' in \emph{International Conference on
  Machine Learning}.\hskip 1em plus 0.5em minus 0.4em\relax PMLR, 2021, pp.
  684--693.

\bibitem{fountoulakis2022graph}
K.~Fountoulakis, A.~Levi, S.~Yang, A.~Baranwal, and A.~Jagannath, ``Graph
  attention retrospective,'' \emph{arXiv preprint arXiv:2202.13060}, 2022.

\bibitem{jin2019graph}
D.~Jin, Z.~Liu, W.~Li, D.~He, and W.~Zhang, ``Graph convolutional networks meet
  markov random fields: Semi-supervised community detection in attribute
  networks,'' in \emph{Proceedings of the AAAI conference on artificial
  intelligence}, vol.~33, no.~01, 2019, pp. 152--159.

\bibitem{qu2019gmnn}
M.~Qu, Y.~Bengio, and J.~Tang, ``Gmnn: Graph markov neural networks,'' in
  \emph{International conference on machine learning}.\hskip 1em plus 0.5em
  minus 0.4em\relax PMLR, 2019, pp. 5241--5250.

\bibitem{kuck2020belief}
J.~Kuck, S.~Chakraborty, H.~Tang, R.~Luo, J.~Song, A.~Sabharwal, and S.~Ermon,
  ``Belief propagation neural networks,'' \emph{Advances in Neural Information
  Processing Systems}, vol.~33, 2020.

\bibitem{jia2022unifying}
J.~Jia and A.~R. Benson, ``A unifying generative model for graph learning
  algorithms: Label propagation, graph convolutions, and combinations,''
  \emph{SIAM Journal on Mathematics of Data Science}, vol.~4, no.~1, pp.
  100--125, 2022.

\bibitem{satorras2021neural}
V.~G. Satorras and M.~Welling, ``Neural enhanced belief propagation on factor
  graphs,'' in \emph{International Conference on Artificial Intelligence and
  Statistics}.\hskip 1em plus 0.5em minus 0.4em\relax PMLR, 2021, pp. 685--693.

\bibitem{jia2021graph}
J.~Jia, C.~Baykal, V.~K. Potluru, and A.~R. Benson, ``Graph belief propagation
  networks,'' \emph{arXiv preprint arXiv:2106.03033}, 2021.

\bibitem{qu2021neural}
M.~Qu, H.~Cai, and J.~Tang, ``Neural structured prediction for inductive node
  classification,'' in \emph{International Conference on Learning
  Representations}, 2021.

\bibitem{mehta2019stochastic}
N.~Mehta, L.~C. Duke, and P.~Rai, ``Stochastic blockmodels meet graph neural
  networks,'' in \emph{International Conference on Machine Learning}.\hskip 1em
  plus 0.5em minus 0.4em\relax PMLR, 2019, pp. 4466--4474.

\bibitem{poor2013introduction}
H.~V. Poor, \emph{An introduction to signal detection and estimation}.\hskip
  1em plus 0.5em minus 0.4em\relax Springer Science \& Business Media, 2013.

\bibitem{zhang2020non}
A.~R. Zhang and Y.~Zhou, ``On the non-asymptotic and sharp lower tail bounds of
  random variables,'' \emph{Stat}, vol.~9, no.~1, p. e314, 2020.

\bibitem{ding2021efficient}
J.~Ding, Z.~Ma, Y.~Wu, and J.~Xu, ``Efficient random graph matching via degree
  profiles,'' \emph{Probability Theory and Related Fields}, vol. 179, no.~1,
  pp. 29--115, 2021.

\bibitem{li2019optimizing}
P.~Li, I.~Chien, and O.~Milenkovic, ``Optimizing generalized pagerank methods
  for seed-expansion community detection,'' \emph{Advances in Neural
  Information Processing Systems}, vol.~32, 2019.

\bibitem{lin1991divergence}
J.~Lin, ``Divergence measures based on the shannon entropy,'' \emph{IEEE
  Transactions on Information Theory}, vol.~37, no.~1, pp. 145--151, 1991.

\bibitem{zeng2019graphsaint}
H.~Zeng, H.~Zhou, A.~Srivastava, R.~Kannan, and V.~Prasanna, ``Graphsaint:
  Graph sampling based inductive learning method,'' in \emph{International
  Conference on Learning Representations}, 2019.

\bibitem{yin2022algorithm}
H.~Yin, M.~Zhang, Y.~Wang, J.~Wang, and P.~Li, ``Algorithm and system co-design
  for efficient subgraph-based graph representation learning,'' \emph{arXiv
  preprint arXiv:2202.13538}, 2022.

\bibitem{weiss2001optimality}
Y.~Weiss and W.~T. Freeman, ``On the optimality of solutions of the max-product
  belief-propagation algorithm in arbitrary graphs,'' \emph{IEEE Transactions
  on Information Theory}, vol.~47, no.~2, pp. 736--744, 2001.

\bibitem{pearl1982reverend}
J.~Pearl, ``Reverend bayes on inference engines: a distributed hierarchical
  approach,'' in \emph{Proceedings of the AAAI conference on artificial
  intelligence}, 1982, pp. 133--136.

\bibitem{cybenko1989approximation}
G.~Cybenko, ``Approximation by superpositions of a sigmoidal function,''
  \emph{Mathematics of Control, Signals and Systems}, vol.~2, no.~4, pp.
  303--314, 1989.

\bibitem{hornik1989multilayer}
K.~Hornik, M.~Stinchcombe, and H.~White, ``Multilayer feedforward networks are
  universal approximators,'' \emph{Neural Networks}, vol.~2, no.~5, pp.
  359--366, 1989.

\bibitem{sen2008collective}
P.~Sen, G.~Namata, M.~Bilgic, L.~Getoor, B.~Galligher, and T.~Eliassi-Rad,
  ``Collective classification in network data,'' \emph{AI Magazine}, vol.~29,
  no.~3, pp. 93--93, 2008.

\end{thebibliography}



\newpage

\appendix




\section{Preliminaries}
\label{app:A}
\subsection{Notations}
In this section, we introduce additional notations for the convenience of presenting proofs. Let $A = (A_{ij})$ be the adjacency matrix of the graph. $D$ is the degree matrix of $A$. $a(n) \sim b(n)$ denotes $a(n) / b(n) \rightarrow 1$ as $n \rightarrow \infty$. $X \stackrel{p}{\sim} \mathbb{P}_X$ denotes random variable $X$ following distribution $\mathbb{P}_X$. Let $\mathbf{p}_X$ denote the density function for the random variable $X$. Let $\Phi(\cdot; \mu, \sigma^2)$ denote the cumulative distribution function for Gaussian distribution with mean $\mu$ and variance $\sigma^2$. Particularly, standard Gaussian distribution function is shorthand as $\Phi(\cdot)$. $\delta(\cdot)$ denote the Dirac function.

\subsection{Graph Structure Concentration Properties}


We first introduce the concept of concentration ball ($B(\delta_1, \delta_2)$), which is used to illustrate the concentration of degrees and class sizes on CSBM$(n, p, q, \mathbb{P}_1, \mathbb{P}_{-1})$ and further helps with eliminating the weak correlations of random variables after optimal non-linear propagation.
\begin{lemmax}[Concentration Ball]
\label{app.concentrate}
    We define the concentration ball $B(\delta_1, \delta_2)$ as: 
    \begin{align}
        B(\delta_1, \delta_2) =& \left\{ |\mathcal{C}_{1}|, |\mathcal{C}_{-1}| \in [\frac{n}{2}(1 - \delta_1), \frac{n}{2}(1 + \delta_1)] \right\}\\
        &\bigcap_{u \in [n]} \left\{ D_{uu} \in [\frac{(p+q)(n-1)}{2}(1 - \delta_2), \frac{(p+q)(n-1)}{2}(1 + \delta_2)] \right\} \\
        &\bigcap_{u \in [n] \atop j \in \left\{-1, 1 \right\}, j = Y_u} \left\{ \frac{|\mathcal{C}_{j}\cap\mathcal{N}_{u}|}{|D_{uu}|} \in [\frac{p}{p + q}(1 - \delta_2), \frac{p}{p + q}(1 + \delta_2)]\right\}\\
        &\bigcap_{u \in [n] \atop j \in \left\{-1, 1 \right\}, j \neq Y_u} \left\{ \frac{|C_{j}\cap\mathcal{N}_{u}|}{|D_{uu}|} \in [\frac{q}{p + q}(1 - \delta_2), \frac{q}{p + q}(1 + \delta_2)]\right\}
    \end{align}
    For any small $\epsilon>0$, choose $\delta_1 = n^{\epsilon - 1/2}$ , $\delta_2 = (np)^{\epsilon - 1/2}$, then for any $c_1>0$, there exists a constant $c_2>0$ , such that 
    \begin{align}
        \mathbb{P}(B(n^{\epsilon - 1/2}, (np)^{\epsilon - 1/2})) \geq 1 - c_1\exp(-c_2n^\epsilon)
    \end{align}
\end{lemmax}

\begin{lemmaproofx}
    Consider two equi-probable classes $\mathcal{C}=\{-1,1\}$,
    let $\left\{I_u\right\}$ be a set of $Ber(1/2)$ variables where $I_u = 1$ if $u \in \mathcal{C}_1$, otherwise, $I_u = 0$. By the Chernoff bound for bounded variables, there is a constant $\kappa_1>0$ such that
    \begin{align}
        \mathbb{P}[|\frac{1}{n}\sum_{u = 1}^{n}I_u - \frac{1}{2}| \geq \frac{\delta_1}{2}] \leq 2 \exp(-\kappa_1 n \delta_1^2)
    \end{align}
    Since $|\mathcal{C}_1| + |\mathcal{C}_{-1}| = n$, the above implies
    \begin{align}
        \mathbb{P}[\frac{1}{2}(1 - \delta_1) \leq \frac{|C_1|}{n}, \frac{|\mathcal{C}_{-1}|}{n} \leq \frac{1}{2}(1 + \delta_1)] \geq 1 - 2 \exp(-\kappa_1 n \delta_1^2)
    \end{align}
    The expected degrees for each node are:
    \begin{align}
        \mathbb{E}[D_{uu}] = \frac{(n-1)}{2}(p+q)
    \end{align}
    Since degrees of each node are sums of Bernoulli random variables, by the Chernoff bound, there exist a constant $\kappa_2>0$ such that 
    \begin{align}
        \mathbb{P}[|D_{uu} - \mathbb{E}[D_{uu}]|\geq \delta_2 \mathbb{E}[D_{uu}]] \leq 2 \exp(-\kappa_2 \mathbb{E}[D_{uu}]\delta_2^2)
    \end{align}
    Therefore, there exist constants $\kappa_3>0$ and $c>0$ such that
    \begin{align}
        \mathbb{P}[\frac{p+q}{2}(1 - \delta_2) \leq \frac{D_{uu}}{n} \leq \frac{p+q}{2}(1 + \delta_2)] \geq 1 - c\exp(-\kappa_3n(p+q)\delta_2^2)
    \end{align}
    Similarly, since
    \begin{align}
        &\mathbb{E}[|\mathcal{C}_{j}\cap\mathcal{N}_{u}|] = \frac{p(n-1)}{2}, \quad \text{when } j \in \left\{-1, 1\right\}, j = Y_u \\
        &\mathbb{E}[|\mathcal{C}_{j}\cap\mathcal{N}_{u}|] = \frac{q(n-1)}{2}, \quad \text{when } j \in \left\{-1, 1\right\}, j \neq Y_u \\
    \end{align}
    by the Chernoff bound, there exist constants $\kappa_4>0$ and $c'>0$ such that
    
    (1) when $j \in \left\{-1, 1\right\}, j = Y_u$
    \begin{align}
        \mathbb{P}[\frac{|\mathcal{C}_{j}\cap\mathcal{N}_{u}|}{|D_{uu}|} \in [\frac{p}{p+q}(1 - \delta_2), \frac{p}{p+q}(1 + \delta_2)]] \geq 1 - c'\exp(-\kappa_4pn(\delta_2 + o(\delta_2))^2),
    \end{align}
    (2) and when $j \in \left\{-1, 1\right\}, j \neq Y_u$
    \begin{align}
        \mathbb{P}[\frac{|\mathcal{C}_{j}\cap\mathcal{N}_{u}|}{|D_{uu}|} \in [\frac{q}{p+q}(1 - \delta_2), \frac{q}{p+q}(1 + \delta_2)]] \geq 1 - c'\exp(-\kappa_4pn(\delta_2 + o(\delta_2))^2)
    \end{align}
    By the union bound, since $(p, q)$ satisfies Assumption~\ref{as.1}, for any $\epsilon>0$, choose $\delta_1 = n^{\epsilon - 1/2}$, $\delta_2 = (np)^{\epsilon - 1/2}$, there exist constants $c_1>0$ and $c_2>0$ such that
    \begin{align}
        \mathbb{P}(B(n^{\epsilon - 1/2}, (np)^{\epsilon - 1/2})) \geq 1 - c_1\exp(-c_2n^\epsilon)
    \end{align}
\end{lemmaproofx}

Lemma~\ref{app.concentrate} reveals the structural concentration properties of CSBM and our following analysis will continuously using the above results.

\subsection{Understanding the Weight Parameter in the Linear Model}
\label{app:weight}
In this discussion, we will highlight the role of the weight parameter in linear model. The linear model aggregation is fundamentally the convolution of Gaussian random variables. From graph structure concentration Lemma~\ref{app.concentrate}, for a single node, there will be $np/2(1 + o_n(1))$ same class Gaussian attributes and $nq/2(1 + o_n(1))$ different class Gaussian attributes. Therefore, for large graph node $n$,
\begin{align}
    \mathcal{P}_v^l(w) \stackrel{p}{\sim} \mathcal{N}([1 + w(p - q)n(1 + o_n(1))/2]\mathbb{E}[\psi_{\text{Gau}}(X_v)], [1 + (n - 1)w^2]\text{var}[\psi_{\text{Gau}}(X_v)])
\end{align}
where $\text{sgn}[w\cdot (p - q)] = 1$. From the above, $\rho_l(w)$ can be directly calculated as: let $Y_u \neq Y_v $, if there exist a constant $c > 0$ such that $|w| > c$, and $|p-q|=\omega_{n}(\frac{\log^2n}{n})$ due to Assumption~\ref{as.1}, we have
\begin{align}
    \rho_l(w) =& \frac{[1 + w(p - q)n(1 + o_n(1))/2]|\mathbb{E}[\psi_{\text{Gau}}(X_v)] - \mathbb{E}[\psi_{\text{Gau}}(X_u)]|}{\sqrt{[1 + w^2(p + q)n(1 + o_n(1))/2]\text{var}[\psi_{\text{Gau}}(X_v)]}} \\
    \sim & \frac{|p - q|n/2|\mathbb{E}[\psi_{\text{Gau}}(X_v)] - \mathbb{E}[\psi_{\text{Gau}}(X_u)]|}{\sqrt{n(p + q)/2\text{var}[\psi_{\text{Gau}}(X_v)]}}
    \label{eq.weight_infty}
\end{align}
As we can see that no matter how we choose weight parameter $w$, unless its absolute value is greater than a positive constant, the aggregation of neighbor attributes will dominate $\rho_l(w)$. To this end, in the following analysis, we can simply let $\rho_l(w^*)$ denote the SNR of the aggregated neighbor attributes, i.e. $w \to \infty$, formulated in Equation~\ref{eq.weight_infty}.

\section{Proof of Proposition 1}
\label{app:pro}
\begin{proof}
From Section~\ref{sec:bayes}, the MAP estimation $f^*(X_v,\{X_{u}\}_{u\in\mathcal{N}_v})$ can be formulated as
\begin{align}
f^*(X_v,\{X_{u}\}_{u\in\mathcal{N}_v}) = \argmax_{Y_v\in\mathcal{C}} \; \mathbb{P}_{Y_v}\left(X_v\right)\prod_{u\in\mathcal{N}_v} \max_{Y_u\in\mathcal{C}}\;\mathbb{P}_{Y_u}\left(X_u\right)p^{(1+Y_vY_u)/2}q^{(1-Y_vY_u)/2}
\end{align}

Let $\Psi_{\text{MAP}}(Y_v, X_v,\{X_{u}\}_{u\in\mathcal{N}_v}) = \mathbb{P}_{Y_v}\left(X_v\right)\prod_{u\in\mathcal{N}_v} \max_{Y_u\in\mathcal{C}}\;\mathbb{P}_{Y_u}\left(X_u\right)p^{(1+Y_vY_u)/2}q^{(1-Y_vY_u)/2}$, and we consider the function in the log regime, the above MAP is equivalent to $\argmax_{Y_v\in\mathcal{C}} \; \log \Psi_{\text{MAP}}(Y_v, X_v,\{X_{u}\}_{u\in\mathcal{N}_v})$. Therefore, we have
\begin{align}
    f^*(X_v,\{X_{u}\}_{u\in\mathcal{N}_v}) =& \text{sgn}[\log \Psi_{\text{MAP}}(1, X_v,\{X_{u}\}_{u\in\mathcal{N}_v}) - \log \Psi_{\text{MAP}}(-1, X_v,\{X_{u}\}_{u\in\mathcal{N}_v})] \\
    =& \text{sgn}[\log \mathbb{P}_{1}(X_v) - \log \mathbb{P}_{-1}(X_v) + \sum_{u \in \mathcal{N}_v}g(X_u)]
\end{align}
where $g(X_u) = \max \{\log (p \cdot \mathbb{P}_{1}(X_u)), \log (q \cdot \mathbb{P}_{-1}(X_u)) \} - \max \{\log (q \cdot \mathbb{P}_{1}(X_u)), \log (p \cdot \mathbb{P}_{-1}(X_u)) \}$. Our next discussion assumes that $p\geq q$ and the case when $p<q$ can be derived in a similar way.

    (1) If $\max \left\{ p \cdot \mathbb{P}_{1}(X_u), q \cdot \mathbb{P}_{-1}(X_u) \right\} = p \cdot \mathbb{P}_{1}(X_u)$, $\max \left\{ q \cdot \mathbb{P}_{1}(X_u), p \cdot \mathbb{P}_{-1}(X_u) \right\} = q \cdot \mathbb{P}_{1}(X_u)$:
    \begin{align}
        &p \cdot \mathbb{P}_{1}(X_u) \geq q \cdot \mathbb{P}_{-1}(X_u) \Leftrightarrow \log \frac{p}{q} + \log \frac{\mathbb{P}_{1}(X_u)}{\mathbb{P}_{-1}(X_u)} \geq 0 \Leftrightarrow \log \frac{\mathbb{P}_{1}(X_u)}{\mathbb{P}_{-1}(X_u)} \geq -\log \frac{p}{q}\\
        &q \cdot \mathbb{P}_{1}(X_u) \geq p \cdot \mathbb{P}_{-1}(X_u) \Leftrightarrow -\log \frac{p}{q} + \log \frac{\mathbb{P}_{1}(X_u)}{\mathbb{P}_{-1}(X_u)} \geq 0 \Leftrightarrow \log \frac{\mathbb{P}_{1}(X_u)}{\mathbb{P}_{-1}(X_u)} \geq \log \frac{p}{q}
    \end{align}
    In this case, when $\log \frac{\mathbb{P}_{1}(X_u)}{\mathbb{P}_{-1}(X_u)} \geq \log \frac{p}{q}$, $g(X_i) = \log \frac{p}{q}$
    
    (2) If $\max \left\{ p \cdot \mathbb{P}_{1}(X_u), q \cdot \mathbb{P}_{-1}(X_u) \right\} = p \cdot \mathbb{P}_{1}(X_u)$, $\max \left\{ q \cdot \mathbb{P}_{1}(X_u), p \cdot \mathbb{P}_{-1}(X_u) \right\} = p \cdot \mathbb{P}_{-1}(X_u)$:
    \begin{align}
        &p \cdot \mathbb{P}_{1}(X_u) \geq q \cdot \mathbb{P}_{-1}(X_u) \Leftrightarrow \log \frac{\mathbb{P}_{1}(X_u)}{\mathbb{P}_{-1}(X_u)} \geq -\log \frac{p}{q} \\
        &p \cdot \mathbb{P}_{-1}(X_u) \geq q \cdot \mathbb{P}_{1}(X_u) \Leftrightarrow \log \frac{\mathbb{P}_{1}(X_u)}{\mathbb{P}_{-1}(X_u)} \leq \log \frac{p}{q}
    \end{align}
    In this case, when $\log \frac{\mathbb{P}_{1}(X_u)}{\mathbb{P}_{-1}(X_u)} \in (-\log \frac{p}{q}, \log \frac{p}{q})$, $g(X_u) = \log \frac{\mathbb{P}_{1}(X_u)}{\mathbb{P}_{-1}(X_u)}$.
    
    (3) If $\max \left\{ p \cdot \mathbb{P}_{1}(X_u), q \cdot \mathbb{P}_{-1}(X_u) \right\} = q \cdot \mathbb{P}_{-1}(X_u)$, $\max \left\{ q \cdot \mathbb{P}_{1}(X_u), p \cdot \mathbb{P}_{-1}(X_u) \right\} = q \cdot \mathbb{P}_{1}(X_u)$: This case does not exist.
    
    (4) If $\max \left\{ p \cdot \mathbb{P}_{1}(X_u), q \cdot \mathbb{P}_{-1}(X_u) \right\} = q \cdot \mathbb{P}_{-1}(X_u)$, $\max \left\{ q \cdot \mathbb{P}_{1}(X_u), p \cdot \mathbb{P}_{-1}(X_u) \right\} = p \cdot \mathbb{P}_{-1}(X_u)$:
    From (1)(2), we know when $\log \frac{\mathbb{P}_{1}(X_u)}{\mathbb{P}_{-1}(X_u)} \leq - \log \frac{p}{q}$, $g(X_u) = -\log \frac{p}{q}$.
    
    Now we define the optimal nonlinear propagation $\mathcal{P}_v$ as:
\begin{align}
    \mathcal{P}_v = \psi\left(X_v;\mathbb{P}_1,\mathbb{P}_{-1}\right) + \sum_{u \in \mathcal{N}_v}\phi\left(\psi\left(X_u;\mathbb{P}_1,\mathbb{P}_{-1}\right); \log(p/q)\right) 
\end{align}
where $\psi\left(a;\mathbb{P}_1,\mathbb{P}_{-1}\right) = \log\frac{\mathbb{P}_{1}\left(a\right)}{\mathbb{P}_{-1}\left(a\right)}$ and $\phi(a;\log\frac{p}{q}) = \text{ReLU}(a+\log\frac{p}{q}) - \text{ReLU}(a-\log\frac{p}{q})-\log\frac{p}{q}.$ Hence, the Bayes optimal classifier $f^*(X_v,\{X_{u}\}_{u\in\mathcal{N}_v}) = \text{sgn}[\mathcal{P}_v]$
\end{proof}

\section{Proof of Lemma~\ref{lm.1}}
\label{app:lm1}
Before getting into the proof of Lemma~\ref{lm.1}, we first embark on bridging the relationship between SNRs and error tail bounds. We will present results on the mean and variance estimation over the attributes under the nonlinear model. 

\begin{lemmaC}
\label{lm.main}
    Suppose that $(p,q)$ satisfies Assumption~\ref{as.1}, let $Z$ follow Gaussian distribution $\mathcal{N}(\mu, I / m)$. The behaviors of expectation and variance of $\tilde{Z} := \phi(\psi_{\text{Gau}}(Z; \mu, -\mu), \log \frac{p}{q})$ are given as follows:
    \begin{itemize}[leftmargin=5mm]
        \item[1.] When $\sqrt{m}\|\mu\|_2 = o_n(\log \frac{p}{q})$, we have the following
        \begin{align}
            &\mathbb{E}[\tilde{Z}] = \Theta_n(m\|\mu\|_2) \\
            &\text{var}[\tilde{Z}] = \Theta_n(m\|\mu\|_2)
        \end{align}
        \item[2.] When $\sqrt{m}\|\mu\|_2 = \Omega_n(\log \frac{p}{q})$ and $\sqrt{m}\|\mu\|_2 = o_n(1)$, we have the following
        \begin{align}
            &\mathbb{E}[\tilde{Z}] = \Theta_n(\sqrt{m}\|\mu\|_2\cdot \log\frac{p}{q}) \\
            &\text{var}[\tilde{Z}] = \Theta_n((\log\frac{p}{q})^2)
        \end{align}
        \item[3.] When $\sqrt{m}\|\mu\|_2 = \Omega_{n}(1)$
        \begin{align}
            &\mathbb{E}[\tilde{Z}] = \Theta_n(\log\frac{p}{q}) \\
            &\text{var}[\tilde{Z}] = \Theta_n(\frac{\exp(-\frac{1}{2}m\|\mu\|_2^2)}{\sqrt{m}\|\mu\|_2} \cdot (\log \frac{p}{q})^2)
        \end{align}
    \end{itemize}
\end{lemmaC}
\begin{lemmaproofC}
    We defer the long proof to Section~\ref{app.additional}.
\end{lemmaproofC}

Now in order to better characterize the behavior of attribute random variable under nonlinearities, we introduce a new concept called overlapping index to quantify the agreement between two random variables. Let $X, Y$ be two independent random variables defined over measurable space $(\mathbb{R}^n, \mathcal{F})$ with probability density function $\mathbf{p}_X, \mathbf{p}_Y$. For $X$ and $Y$, the overlapping index $\eta: \mathbb{R}^n \times \mathbb{R}^n \rightarrow [0, 1]$ is defined as follows:
    \begin{align}
        \eta(X, Y) = \int \min \left\{ \mathbf{p}_X(x), \mathbf{p}_Y(y) \right\} dxdy
    \end{align}
    
With the definition of overlapping index $\eta$, the behavior of the above  random variable $\tilde{Z}$ can be characterized in the following lemma:

\begin{lemmaC} 
\label{lm.approximate}
    Suppose that $(p,q)$ satisfies Assumption~\ref{as.1}, let $Z \stackrel{p}{\sim} \mathcal{N}(\mu, I / m)$, $\tilde{Z} := \phi(\psi_{\text{Gau}}(Z; \mu, -\mu), \log \frac{p}{q})$. A random variable $\tau$ satisfy
    \begin{align}
        \eta(\tilde{Z}, \tau) \to 1
    \end{align}
    where $\tau$ takes different forms in different settings:
    \begin{itemize}[leftmargin=5mm]
        \item \textbf{Case I}: When $\sqrt{m}\|\mu\|_2 = o_n(\log \frac{p}{q})$, $\tau = \psi_{\text{Gau}}(Z, \mu, -\mu)$.
        \item \textbf{Case II}: When $\sqrt{m}\|\mu\|_2 = \Omega_n(\log \frac{p}{q})$, $\sqrt{m}\|\mu\|_2 = \mathcal{O}_n(1)$, $\tau / \log \frac{p}{q}$ is a Rademacher random variable.
        \item \textbf{Case III}: When $\sqrt{m}\|\mu\|_2 = \omega_n(1)$, let $\tau$ be a degenerated random variable, i.e. $\mathbb{P}(\tau = \log \frac{p}{q}) = 1$.
    \end{itemize}
\end{lemmaC}
\begin{lemmaproofC}
Let $\psi(Z) = \psi_{\text{Gau}}(Z, \mu, -\mu)$ and $\mathbf{p}_{\psi(Z)}\mathbbm{1}_{[a, b]}$ be the density function constrained on interval $[a, b]$. From Lemma~\ref{lm.main}, 
    \begin{itemize}[leftmargin=5mm]
        \item \textbf{In Case I}: Most Gaussian mass lands on the linear part of non-linear propagation function $\phi(\cdot; \log \frac{p}{q})$. Therefore, the probability of landing on boundary points shrinks to zero:
        \begin{align}
            &\mathbb{P}(\tilde{Z} = -\log \frac{p}{q}) \sim \sqrt{\frac{2}{\pi}}\cdot \frac{\sqrt{m\mu^T\mu}}{\log \frac{p}{q}} \cdot \exp(-\frac{(\log \frac{p}{q})^2}{8m\mu^T\mu})\\
            &\mathbb{P}(\tilde{Z} = \log \frac{p}{q}) \sim \sqrt{\frac{2}{\pi}}\cdot \frac{\sqrt{m\mu^T\mu}}{\log \frac{p}{q}} \cdot \exp(-\frac{(\log \frac{p}{q})^2}{8m\mu^T\mu})
        \end{align}
        Hence, the overlapping index between $\tilde{Z}$ and $\psi(Z)$ is given as:
        \begin{align}
            \eta(\tilde{Z}, \psi(Z)) =& \int_{\RR} \min\left\{\mathbf{p}_{\tilde{Z}} , \mathbf{p}_{\psi(Z)}\right\}dz\\
            \sim& 1 - 4\sqrt{\frac{2}{\pi}}\cdot \frac{\sqrt{m\mu^T\mu}}{\log \frac{p}{q}} \cdot \exp(-\frac{(\log \frac{p}{q})^2}{8m\mu^T\mu}) \rightarrow 1
        \end{align}
        \item \textbf{In Case II}: When the major mass falls on the threshold part of the function (i.e. $\sqrt{m\mu^T\mu} = \Omega_n(\log \frac{p}{q})$, $\sqrt{m\mu^T\mu} = \mathcal{O}_n(1)$), the probability mass of boundary points are:
        \begin{align}
        &\mathbb{P}(\tilde{Z} = -\log \frac{p}{q}) \sim \frac{1}{2} - \frac{1}{2\sqrt{2\pi}}\cdot \frac{\log \frac{p}{q}}{\sqrt{m\mu^T\mu}} ( 1 + o_n(1))\\
        &\mathbb{P}(\tilde{Z} = \log \frac{p}{q}) \sim \frac{1}{2} - \frac{1}{2\sqrt{2\pi}} \cdot \frac{\log \frac{p}{q}}{\sqrt{m\mu^T\mu}}( 1 + o_n(1))
        \end{align}
        Hence, the density function of $\tilde{Z}$ is:
        \begin{align}
            \mathbf{p}_{\tilde{Z}} = (\frac{1}{2} - o_n(1))\delta(-\log\frac{p}{q}) + \mathbf{p}_{\psi(Z)}\mathbbm{1}_{[-\log\frac{p}{q}, \log\frac{p}{q}]} + (\frac{1}{2} - o_n(1)) \delta(\log\frac{p}{q})
        \end{align}
        Define $\tau / (\log \frac{p}{q})$ to be the Rademacher random variable under this setting. Then, the overlapping index between $\tilde{Z}$ and $\tau$ is given as:
        \begin{align}
            \eta(\tilde{Z}, \tau) =& \int_{\RR} \min\left\{f_{\tilde{Z}} , \frac{1}{2}[\delta(-\log\frac{p}{q}) + \delta(\log\frac{p}{q})]\right\} dz\\
            \sim & 1 - \sqrt{\frac{2}{\pi}} \cdot \frac{\log \frac{p}{q}}{\sqrt{m\mu^T\mu}} \rightarrow 1
        \end{align}
        \item \textbf{In Case III}: The third case is straight forward from Lemma~\ref{lm.main}.
    \end{itemize}
\end{lemmaproofC}

The above lemma characterizes the behavior of attribute random variable under the nonlinear model, and to further relate the SNR ($\rho_r$) of nonlinearized attribute feature with mis-classification error $\xi^r$, we introduce the following proposition for tail bound estimation.

\begin{propositionC}[\textbf{Reverse Chernoff-Cramer Bound}~\cite{zhang2020non}]
\label{pro.inv_Chernoff}
    Let $Z$ be a random variable with moment generating function $\phi_Z(\lambda) = \mathbb{E} \exp(\lambda Z)$ defined on a set $\Lambda \in \RR$. For any $t>0$, we have the following lower bounds:
    \begin{align}
        &\mathbb{P}(Z \geq t) \geq \nonumber \\&\sup_{\alpha, \beta > 1, \lambda > \lambda' \geq 0, \atop \lambda \alpha \in \Lambda}\{ \phi_Z(\lambda) \exp(-\beta \lambda t) - \phi_Z(\alpha \lambda) \exp(-\alpha\beta\lambda t) - \exp(-(\beta \lambda - \lambda')t)\phi_Z(\lambda - \lambda') \}, \\
        &\mathbb{P}(Z \leq -t) \geq \nonumber \\& \sup_{\alpha, \beta > 1, \lambda > \lambda' \geq 0, \atop -\lambda \alpha \in \Lambda}\{ \phi_Z(-\lambda) \exp(-\beta \lambda t) - \phi_Z(-\alpha \lambda) \exp(-\alpha\beta\lambda t) - \exp(-(\beta \lambda - \lambda')t)\phi_Z(\lambda' - \lambda) \}.
    \end{align}
\end{propositionC}

Proposition~\ref{pro.inv_Chernoff} provides a Chernoff-type lower bound that can be leveraged in the analysis of node attribute nonlinear propagation. From Lemma~\ref{lm.approximate}, the behavior of nonlinearized attribute can be interpreted as either Gaussian or Rademacher random variables in the limited attributed information setting. Follow the derivation of Theorem 9 in~\cite{zhang2020non} and leverage the (reversed) Chernoff-Cramer inequality for upper bound, we have the following results:

\begin{propositionC}[\textbf{Approximate Lower and Upper Tail Bounds for Nonlinearized Attributes}]
\label{pro.approximate}
    Suppose $Z_1, Z_2, ..., Z_n$ are i.i.d random variables with normal distribution $\mathcal{N}(\mu, \frac{I}{d})$. Let $\tilde{Z}_i := \phi(\psi_{\text{Gau}}(Z_i; \mu, -\mu); \log \frac{p}{q})$ be the nonlinearized attributes. We have the following tail bounds:
    \begin{itemize}[leftmargin=5mm]
        \item When $\sqrt{m}\|\mu\|_2 = o_n(\log \frac{p}{q})$,
        \begin{align}
            \mathbb{P}(\sum_{i}^{n}\tilde{Z}_i \leq -\lambda) \sim \mathbb{P}(\sum_{i}^{n}\tilde{Z}_i \geq \lambda) \sim \exp(-\frac{\lambda^2}{4nm\mu^T\mu}), \quad \forall \lambda \geq 0
        \end{align}
        \item When $\sqrt{m}\|\mu\|_2 = \Omega_n(\log \frac{p}{q})$, $\sqrt{m}\|\mu\|_2 = \mathcal{O}_n(1)$ and $\log \frac{p}{q} = \omega_n(\frac{1}{n})$, there exist a parameter $\zeta$ and two constants $C_{\zeta}, C_{\zeta}'$ that only depend on $\zeta$, such that for $\forall 0 \leq \lambda \leq \frac{n \log \frac{p}{q}}{\zeta}$, we have
        \begin{align}
            &C_{\zeta}' \exp(-C_{\zeta} \frac{\lambda^2}{n(\log \frac{p}{q})^2}) \leq \mathbb{P}(\sum_{i}^{n}\tilde{Z}_i \leq -\lambda) \sim \mathbb{P}(\sum_{i}^{n}\tilde{Z}_i \geq \lambda) \leq \exp(-\frac{\lambda^2}{n(\log \frac{p}{q})^2})
        \end{align}
    \end{itemize}
\end{propositionC}

As we can see from Proposition~\ref{pro.approximate}
that the propagated attributes can be lower and upper bounded by the same denominator in the exponential term by neglecting the constant, i.e. node number times single node attribute variance (from Lemma~\ref{lm.main}). Besides, under Assumption~\ref{as.1}, by leveraging the lower \& upper bounds from Proposition~\ref{pro.approximate}, we can easily show that $\rho_r \sim (\mathbb{E}[\sum_{u \in \mathcal{N}}\mathcal{M}(X_u, p, q) | Y_u = 1] - \mathbb{E}[\sum_{u \in \mathcal{N}}\mathcal{M}(X_u, p, q) | Y_u = -1]) / \text{var}(\sum_{u \in \mathcal{N}}\mathcal{M}(X_u, p, q) | Y_u = 1)$ as $n \to \infty$, i.e. the SNR of nonlinear model is dominated by the SNR of propagated attributes from neighbor nodes.

\textbf{Proof of Lemma~\ref{lm.1}}: In the proof, without loss of generality, we assume the symmetric case, i.e. $\mu = -\nu$, since we can simply construct a linearly transformation ($T$) that map two Gaussians to be symmetrical about the origin, i.e. 
    $T: x \mapsto x - (\|\mu\|_2^2 - \|\nu\|_2^2)/2$
and further analysis can be directly apply to non-symmetric case under inverse mapping $T^{-1}$. Let $n_1(u)$ define the number of nodes in the same class with node $u$, and $n_2(u)$ define the number of nodes in the opposite class of node $u$. When $n \to \infty$, 
    \begin{align}
        &n_1(u) = |\mathcal{C}_j \cap \mathcal{N}_{u}| = \frac{n-1}{2}p(1 + o_n(1)), \text{ where } j = Y_u \\
        &n_2(u) = |\mathcal{C}_j \cap \mathcal{N}_{u}| = \frac{n-1}{2}q(1 + o_n(1)), \text{ where } j \neq Y_u
    \end{align}
    $n_1(u), n_2(u)$ are abbreviated as $n_1, n_2$. Hence, $\xi^l(w)$ and $\xi^r$ can be interpreted as:
    
    \begin{align}
        &\xi^l(w) = \mathbb{P}\left\{ 
        \begin{aligned}
        &Z + w [\sum_{j=1}^{n_1}Z_{j} + \sum_{t=1}^{n_2}\tilde{Z}_{t}] \leq 0 | Z, Z_1, Z_2, ..., Z_{n_1} \stackrel{i.i.d}{\sim} \mathcal{N}(\mu^T\mu, \frac{\mu^T\mu}{d});\\ &\tilde{Z}_1, \tilde{Z}_2, ..., \tilde{Z}_{n_2} \stackrel{i.i.d}{\sim} \mathcal{N}(-\mu^T\mu, \frac{\mu^T\mu}{d}); n_1 = \frac{n-1}{2}p(1 + o_n(1)), n_2 = \frac{n-1}{2}q(1 + o_n(1))
        \end{aligned}
        \right\} \\
        &\xi^r = \mathbb{P}\left\{ 
        \begin{aligned}
        &2mZ + \sum_{j=1}^{n_1}\phi(\psi_{\text{Gau}}(Z_{j})) + \sum_{t=1}^{n_2}\phi(\psi_{\text{Gau}}(\tilde{Z}_{t})) \leq 0 | Z, Z_1, Z_2, ..., Z_{n_1} \stackrel{i.i.d}{\sim} \mathcal{N}(\mu^T\mu, \frac{\mu^T\mu}{d});\\ &\tilde{Z}_1, \tilde{Z}_2, ..., \tilde{Z}_{n_2} \stackrel{i.i.d}{\sim} \mathcal{N}(-\mu^T\mu, \frac{\mu^T\mu}{d}); n_1 = \frac{n-1}{2}p(1 + o_n(1)), n_2 = \frac{n-1}{2}q(1 + o_n(1))
        \end{aligned}
        \right\}
    \end{align}
    where $\phi, \psi$ are the shorthand for $\phi(\cdot; \log \frac{p}{q}), \psi_{\text{Gau}}(\cdot; \mu, -\mu)$. For standard Gaussian distribution, we have the well-known tail bound: for $X \stackrel{p}{\sim} \mathcal{N}(0, 1), \lambda > 0$
\begin{align}
    (\frac{1}{\lambda} - \frac{1}{\lambda^3})\frac{\exp(-\frac{\lambda^2}{2})}{\sqrt{2\pi}} \leq \mathbb{P}(X > \lambda) \leq \frac{1}{\lambda}\frac{\exp(-\frac{\lambda^2}{2})}{\sqrt{2\pi}}
\end{align}
i.e. $\mathbb{P}(X > \lambda) \sim \frac{1}{\lambda}\frac{\exp(-\frac{\lambda^2}{2})}{\sqrt{2\pi}}$ as $\lambda \to \infty$. Consider $(p, q)$ satisfies Assumption~\ref{as.1}, by Gaussian additivity, the linear model $\mathcal{P}_v^l(w) \stackrel{p}{\sim} \mathcal{N}(2m\mu^T\mu(1 + n_1w - n_2w)Y_v, 4m(1 + (n-1)w^2)\mu^T\mu)$ where $n_1 = \frac{p(n-1)}{2}(1 + o_n(1)), n_2 = \frac{q(n-1)}{2}(1 + o_n(1))$. Hence, 
\begin{align}
    \xi^l(w) = \mathbb{P}(\mathcal{P}_v^l(w) \cdot Y_v < 0) \sim & \frac{\sqrt{4m(1 + (n-1)w^2)\mu^T\mu)}}{2m\mu^T\mu(1 + n_1w - n_2w)} \cdot \frac{\exp(-\frac{[2m\mu^T\mu(1 + n_1w - n_2w)]^2}{8m(1 + (n-1)w^2)\mu^T\mu)})}{\sqrt{2\pi}} \\
    = & \frac{1}{\rho_l(w)} \frac{\exp(-\frac{\rho_l(w)^2}{2})}{\sqrt{2\pi}} = \exp(-\frac{\rho_l(w)^2(1 + o_n(1))}{2})
\end{align}
Now consider nonlinear model mis-classification error $\xi^r$. 
\begin{itemize}[leftmargin=5mm]
    \item[1.]When attributed information is very limited, i.e. $\sqrt{m\mu^T\mu} = o_n(\log \frac{p}{q})$, according to Proposition~\ref{pro.approximate}, $\rho_r \sim \rho_l(w^*)$ and $\xi^r \sim \xi^l(w^*) \sim \exp(- \rho_r^2(1 + o_n(1))/2)$.
    \item[2.]When attributed information limited, i.e. $\sqrt{m\mu^T\mu} = \Omega_n(\log \frac{p}{q}), m\mu^T\mu = \mathcal{O}_n(1)$, according to Proposition~\ref{pro.approximate}, when $n \to \infty$, there exist two constants $C > 0, C' > 0$ such that
    \begin{align}
        \xi^r\in [C\exp(-C' \rho_r^2 \frac{\text{var}(\mathcal{P}_v | Y_v = 1)}{2(\log \frac{p}{q})^2}), \exp(-\rho_r^2 \frac{\text{var}(\mathcal{P}_v | Y_v = 1)}{2(\log \frac{p}{q})^2})]
    \end{align}
    From Lemma~\ref{lm.main}, $\text{var}(\mathcal{P}_v | Y_v = 1) = (\log \frac{p}{q})^2$, we have
    \begin{align}
        \xi^r\in [C\exp(-C' \frac{\rho_r^2}{2}), \exp(-\frac{\rho_r^2}{2})]
    \end{align}
    \item[3.]When attributed information is sufficient, i.e. $\sqrt{m\mu^T\mu} = o_n(1)$, as $n \to \infty$, $\mathcal{P}_v$ will be highly concentrated on $\mathbb{E}[\mathcal{P}_v]$ with Gaussian upper tail bound. Further, we have $\xi^r \leq \exp(-\rho_r^2 / 2)$. 
\end{itemize}

\section{Proof of Theorem 1}
\label{app:thm1}
\begin{proof}
Similar to the argument made in the proof of Lemma~\ref{lm.1}, we can assume the symmetric case, i.e. $\mu = -\nu$.
    First, we consider the nonlinear model. Let $A = \left\{\exists u \in \mathcal{C}_{-1}, \mathcal{P}_u>0 \right\}$, $B = \left\{\exists u \in \mathcal{C}_{-1}, \mathcal{P}_u<0 \right\}$, by lemma~\ref{lm.1}, for any small $\epsilon > 0$ choose $\delta_1 = n^{\epsilon - 1/2}, \delta_2 = (np)^{\epsilon - 1/2}$, we have
    \begin{align}
        \mathbb{P}(\exists v \in\mathcal{V}, \mathcal{P}_v \cdot Y_v < 0) =& \mathbb{P}(A \vee B) \\
        =& \mathbb{P}(A \vee B | B(\delta_1, \delta_2)) \mathbb{P}(B(\delta_1, \delta_2)) + \mathbb{P}(A \vee B | B^c(\delta_1, \delta_2)) \mathbb{P}(B^c(\delta_1, \delta_2)) \\
        \leq& \mathbb{P}(A \vee B | B(\delta_1, \delta_2)) + c_1\exp(-c_2n^\epsilon)
    \end{align}
    Therefore, by the union bound, we have the bound
    \begin{align}
        \mathbb{P}(\exists v \in\mathcal{V}, \mathcal{P}_v \cdot Y_v < 0) \leq & n \mathbb{P}(\mathcal{P}_u \cdot Y_u < 0) + c_1\exp(-c_2n^\epsilon) \\
        = & n \xi^r + c_1\exp(-c_2n^\epsilon)
    \end{align}
    From Lemma~\ref{lm.1}, $\xi^r$ is upper bounded by $\exp(-\rho_r^2 / 2)$ under all attribute settings, and hence
    \begin{align}
        \mathbb{P}(\exists v \in\mathcal{V}, \mathcal{P}_v \cdot Y_v < 0) \leq n \exp(-\rho_r^2/2) + c_1\exp(-c_2n^\epsilon)
    \end{align}
    Letting the right hand side diminish to zero as $n \to \infty$, we need
    \begin{align}
    \label{eq.tailbound}
        n\exp(-\rho_r^2/2) = o_n(1)
    \end{align}
  
         \begin{itemize}[leftmargin=5mm]
             \item[1.] When $m\mu^T\mu = o_n(\log \frac{p}{q})$, under Assumption~\ref{as.1}, from Lemma~\ref{lm.main},
             \begin{align}
                \text{Equation~\ref{eq.tailbound}} \Leftrightarrow \frac{n^2(p - q)^2(m\mu^T\mu)^2}{ (p + q) m\mu^T\mu n\log n} = \omega_n(1) \Leftrightarrow \sqrt{m}\|\mu\|_2 = \omega_n(\sqrt{\frac{\log n}{S(p, q)n}})
             \end{align}
             \item[2.] When $m\mu^T\mu = \Omega_n(\log \frac{p}{q})$ and $m\mu^T\mu = \mathcal{O}_n(1)$, under Assumption~\ref{as.1}, from Lemma~\ref{lm.main}, we have the similar results:
             \begin{align}
                \text{Equation~\ref{eq.tailbound}} \Leftrightarrow \frac{n^2(p - q)^2(\sqrt{m \mu^T\mu}\log \frac{p}{q})^2}{(p + q) \log (\frac{p}{q})^2 n\log n } = \omega_n(1) \Leftrightarrow \sqrt{m}\|\mu\|_2 = \omega_n(\sqrt{\frac{\log n}{S(p, q)n}})
             \end{align}
             \item[3.] When $m\mu^T\mu = \omega_n(1)$, $\tilde{Z}$, Equation~\ref{eq.tailbound} is naturally satisfied.
         \end{itemize}
    Summarizing the above results, when $\sqrt{m}\|\mu\|_2 = \omega_n(\sqrt{\log n / S(p, q)n})$, $\mathbb{P}(\forall v \in \mathcal{V}, \mathcal{P}_v \cdot Y_v > 0) = 1 - \mathbb{P}(\exists v \in \mathcal{V}, \mathcal{P}_v \cdot Y_v < 0) = 1 - o_n(1)$.
    
    The above analysis can directly applies to any linear model with the given qualified weight parameter $w$, i.e., $|w|\rightarrow c>0$ and $\text{sgn}(w)=\text{sgn}(p-q)$. According to Section~\ref{app:weight}, 
    \begin{align}
        n\exp(-\rho_l^2(w^*)/2) = o_n(1) \Leftrightarrow \frac{n^2(p - q)^2(m\mu^T\mu)^2}{n\log n (p + q) m\mu^T\mu} = \omega_n(1) \Leftrightarrow \sqrt{m}\|\mu\|_2 = \omega_n(\sqrt{\frac{\log n}{S(p, q)n}})
    \end{align}
    \end{proof}

\section{Proof of Theorem 2}
\label{app:thm2}
Suppose that $(p, q)$ satisfies Assumption\ref{as.1}, and the separable condition in Theorem~\ref{thm.1} is satisfied. Proposition~\ref{pro.approximate} actually implies that Lyapunov's central limit theorem provides uniformly tight tail bounds for $\mathcal{P}_v$ and Gaussian approximation error can be neglected. Hence, by leveraging Lemma~\ref{lm.main}, and $\rho_l(w^*)$ discussed in Section~\ref{app:weight},
         \begin{itemize}[leftmargin=5mm]
             \item[1.] When $m\mu^T\mu = o_n(\log \frac{p}{q})$:
             \begin{align}
                 \rho_r / \rho_l(w^*) \sim \frac{n(p - q)(2m\mu^T\mu)}{\sqrt{n\log n (p + q) 4m\mu^T\mu}} / \frac{n(p - q)(\mu^T\mu)}{\sqrt{n\log n (p + q) \mu^T\mu/m}} = 1
             \end{align}
             \item[2.] When $m\mu^T\mu = \Omega_n(\log \frac{p}{q})$ and $m\mu^T\mu = \mathcal{O}_n(1)$:
             \begin{align}
                 \rho_r / \rho_l(w^*) = \Theta_n(\frac{n(p - q)(\sqrt{m\mu^T\mu} \log \frac{p}{q})}{\sqrt{n\log n (p + q) (\log \frac{p}{q})^2}} / \frac{n(p - q)(\mu^T\mu)}{\sqrt{n\log n (p + q) \mu^T\mu/m}}) = \Theta_n(1)
             \end{align}             
             \item[3.] When $m\mu^T\mu = \omega_n(1)$, and $m\mu^T\mu / \exp(-\frac{1}{2}m\mu^T\mu) = \mathcal{O}_n(nS(p, q))$
             \begin{align}
                 \rho_r / \rho_l(w^*) = \Theta_n(\frac{\sqrt{\exp(\frac{1}{2}m\mu^T\mu)}}{{\sqrt[4]{m\mu^T\mu}}})
            \end{align}
            \item[4.] When $m\mu^T\mu = o_n(\log^2 n)$, and $m\mu^T\mu / \exp(-\frac{1}{2}m\mu^T\mu) = \omega_n(nS(p, q))$
             \begin{align}
                 \rho_r / \rho_l(w^*) = \Theta_n(\frac{\sqrt{n(p + q)} \cdot \log \frac{p}{q}}{m\mu^T\mu})
            \end{align}
            Further, to unify the sufficient attributed setting $m\mu^T\mu = \omega_n(1)$, we have $\rho_r / \rho_l(w^*) = \omega_n(\min\{\exp(m\|\mu - \nu\|_2^2/3), nS(p, q) m^{-1}\|\mu - \nu \|_2^{-2}\})$.
         \end{itemize}

\section{Proof of Theorem 3}
\label{app:thm3}
In order to further compare the transferability of both nonlinear and linear models, we introduce several results for the proof of Theorem~\ref{thm:3}.



\begin{lemmaF}[\textbf{Error Gap Under Mean Perturbations}]
\label{lm.perturb_error}
    Let $X_1, X_2$ follow Gaussian distributions $\mathcal{N}(\mu, \sigma^2)$ and $\mathcal{N}(-\mu, \sigma^2)$ respectively, and $\mu, \sigma^2$ are all the function of node number $n$. Exerting mean perturbations on two distributions, which gives $\mathcal{N}(\mu - \Delta \mu_1, \sigma^2)$ and $\mathcal{N}(-\mu + \Delta \mu_2, \sigma^2)$. Let $\tilde{X}_1, \tilde{X}_2$ be the random variables that follow the perturbed distributions. When $\frac{\mu}{\sigma} \rightarrow \infty$, $\Delta \mu_1 \cdot \mu, \Delta \mu_2 \cdot \mu \approx 0$, we have the estimation on the following error gap:
        \begin{align}
            \Delta \xi &= \mathbb{P}(\{X_1 < 0\} \cup \{X_2 > 0\}) - \mathbb{P}(\{\tilde{X}_1 < 0\} \cup \{ \tilde{X}_2 > 0\}) \\
            &\sim \frac{\exp(-\frac{\mu^2}{2\sigma^2})}{\sqrt{2\pi}} \cdot (\frac{\Delta \mu_1}{\sigma} + \frac{\Delta \mu_2}{\sigma})
        \end{align}
\end{lemmaF}
\begin{lemmaproofF}
    First, define the tail function $f(x) = \frac{1}{x}\exp( - \frac{x^2}{2})$. When $x \to \infty$ and $\Delta x \cdot x \approx 0$, the value gap between $f(x + \Delta x)$ and $f(x)$:
    \begin{align}
       & f( x + \Delta x) - f(x) \\
    = & \frac{1}{x + \Delta x} \exp( - \frac{(x + \Delta x)^2}{2}) - \frac{1}{x} \exp( - \frac{x^2}{2}) \\
    =& \frac{1}{x + \Delta x} [\exp( - \frac{(x + \Delta x)^2}{2}) - \exp( - \frac{x^2}{2})] + \frac{ \Delta x}{x + \Delta x} \exp( - \frac{x^2}{2}) \\
    \sim&  -(\Delta x + \frac{\Delta x}{x}) \cdot \exp(-\frac{x^2}{2}) \\
    \sim & -\Delta x \cdot \exp( - \frac{x^2}{2})
\end{align}
From above, we have
    \begin{align}
        &\mathbb{P}(X_1 < 0) - \mathbb{P}(X_2 > 0) - \mathbb{P}(\tilde{X}_1 < 0 ) + \mathbb{P}(\tilde{X}_2 > 0) \\
        =& \Phi(\frac{\mu - \Delta \mu_2}{\sigma}) - 1 - \Phi(- \frac{\mu - \Delta \mu_1}{\sigma}) - [\Phi(\frac{\mu}{\sigma}) - 1] + \Phi(- \frac{\mu}{\sigma}) \\
        \sim & -\frac{1}{\sqrt{2 \pi}}[\frac{\sigma}{\mu - \Delta \mu_2} \exp(- \frac{1}{2}(\frac{\mu - \Delta \mu_2}{\sigma})^2) + \frac{\sigma}{\mu - \Delta \mu_1} \exp(- \frac{1}{2}(\frac{\mu - \Delta \mu_1}{\sigma})^2) - \frac{2\sigma}{\mu} \exp( - \frac{\mu^2}{2\sigma^2})] \\
        \sim & \frac{\exp(-\frac{\mu^2}{2\sigma^2})}{\sqrt{2\pi}} \cdot (\frac{\Delta \mu_1}{\sigma} + \frac{\Delta \mu_2}{\sigma})
    \end{align}
\end{lemmaproofF}

\textbf{Proof of Theorem~\ref{thm:3}:} Suppose that $(p, q)$ satisfies Assumption\ref{as.1}, and the separable condition in Theorem~\ref{thm.1} is satisfied. As stated in the proof of Theorem~\ref{app:thm2}, as $n \to \infty$, the nonlinear model $\mathcal{P}_v$ will approximate Gaussian distributions with similar tail bounds except high order small quantities. For any given $|w|>c$, let $\Delta \rho_r = (\mathbb{E}[\mathcal{P}_v|Y_v = 1, \mathcal{G}] - \mathbb{E}[\mathcal{P}_v|Y_v = -1, \mathcal{G}] - \mathbb{E}[\mathcal{P}_v|Y_v = 1, \mathcal{G}'] + \mathbb{E}[\mathcal{P}_v|Y_v = -1, \mathcal{G}']) / \sqrt{\text{var}(\mathcal{P}_v|Y_v = 1, \mathcal{G})}, \Delta \rho_l(w^*) = (\mathbb{E}[\mathcal{P}_v^l(w^*)|Y_v = 1, \mathcal{G}] - \mathbb{E}[\mathcal{P}_v^l(w^*)|Y_v = -1, \mathcal{G}] - \mathbb{E}[\mathcal{P}_v^l(w^*)|Y_v = 1, \mathcal{G}'] + \mathbb{E}[\mathcal{P}_v^l(w^*)|Y_v = -1, \mathcal{G}'])/ \sqrt{\text{var}(\mathcal{P}_v^l(w^*)|Y_v = 1, \mathcal{G})}$. Hence, by leveraging Lemma~\ref{lm.perturb_error}
\begin{align}
    &\Delta \xi^r \sim \frac{\exp(-\rho_r^2)}{\sqrt{2\pi}}\cdot \Delta \rho_r \label{eq.error_r}\\
    &\Delta \xi^l(w^*) \sim \frac{\exp(-\rho_l^2(w^*))}{\sqrt{2\pi}} \cdot \Delta \rho_l(w^*) \label{eq.error_l}
\end{align}
Hence, in the limited attribute setting, i.e. $m\mu^T\mu = o_n(\log \frac{p}{q})$, by Theorem~\ref{thm:2}, we have $\rho_r \sim \rho_l(w^*)$. From Lemma~\ref{lm.main}, $\phi$ will almost behave like a linear function when the mean perturbation is small, and hence $\Delta \rho_r / \Delta \rho_l(w^*) \sim [1 - \langle \mu' - \nu', \mu - \nu \rangle / \|\mu - \nu\|_2^2] / [1 - \langle \mu' - \nu', \mu - \nu \rangle / \|\mu - \nu\|_2^2] = 1$. Thus, $\Delta \xi^r / \Delta \xi^l(w^*) \to 1$. In the sufficient attribute setting, i.e. $m\mu^T\mu = \omega_n(1)$, since $\rho_r = \omega_n(\rho_l(w^*))$ and $\phi(\mathbb{E}[\mathcal{P}_v|Y_v = 1, \mathcal{G}] - \mathbb{E}[\mathcal{P}_v|Y_v = 1, \mathcal{G}']) + \phi(\mathbb{E}[\mathcal{P}_v|Y_v = -1, \mathcal{G}] - \mathbb{E}[\mathcal{P}_v|Y_v = -1, \mathcal{G}']) \sim \mathbb{E}[\mathcal{P}_v|Y_v = 1, \mathcal{G}] - \mathbb{E}[\mathcal{P}_v|Y_v = 1, \mathcal{G}'] + \mathbb{E}[\mathcal{P}_v|Y_v = -1, \mathcal{G}] - \mathbb{E}[\mathcal{P}_v|Y_v = -1, \mathcal{G}']$, which implies $\Delta \rho_r / \Delta \rho_l(w^*) \to 1$. Therefore, the exponential term will dominate the error gap, we have $\Delta \xi^r / \Delta \xi^l(w^*) \to 0$.

\section{Experimental Settings}
\label{app:exp}
\subsection{Datasets}
\label{app:exp_datasets}

As mentioned in Section~\ref{sec:exp-real}, this paper includes three real-world citation network datasets: Cora, Citeseer and Pubmed. Their details are included in Table~\ref{tab:dataset}.
\begin{table}
\centering
\caption{Datasets statistics \label{tab:dataset}}
\resizebox{0.95\columnwidth}{!}{
\begin{tabular}{l|ccc|c|ccccccc}
\toprule[1pt]
\textbf{Dataset} & \textbf{\# Nodes} & \textbf{\# Edges} & \textbf{\# Features} & \textbf{\# Class} &  \multicolumn{7}{c}{\textbf{\# Nodes in Each Class}} \\ \hline
\textbf{Cora} & 2,708 & 5,428 & 1,433 & 7 & 351 & 217 & 418 & 818 & 426 & 298 & 180\\
\textbf{CiteSeer} & 3,327 & 4,732 & 3,703 & 6 & 264 & 590 & 668 & 701 & 596 & 508 & /\\
\textbf{PubMed} & 19,717 & 44,338 & 500 & 3  & 4103 & 7739 & 7875 & / & / & / & /\\
\bottomrule
\end{tabular}}
\end{table}

\subsection{Environment}
\label{app:exp_env}
Experiments were performed on a server with four Intel 24-Core Gold 6248R CPUs, 1TB DRAM, and eight NVIDIA QUADRO RTX 6000 (24GB) GPUs.

\section{Additional Experimental Results}
\label{app:additional_results}
In this section, we introduce detailed experimental settings and present additional experiments to backup our main results in Section~\ref{sec:main}.

\subsection{Synthetic Experiments}
\label{app:exp_synthetic}

\subsubsection{Analysis on Linear Model Hyper-parameter $w$}
\label{app:exp_w}

\begin{figure}[t]
\label{fig.w}
\centering

\centering
\includegraphics[width=0.32\textwidth]{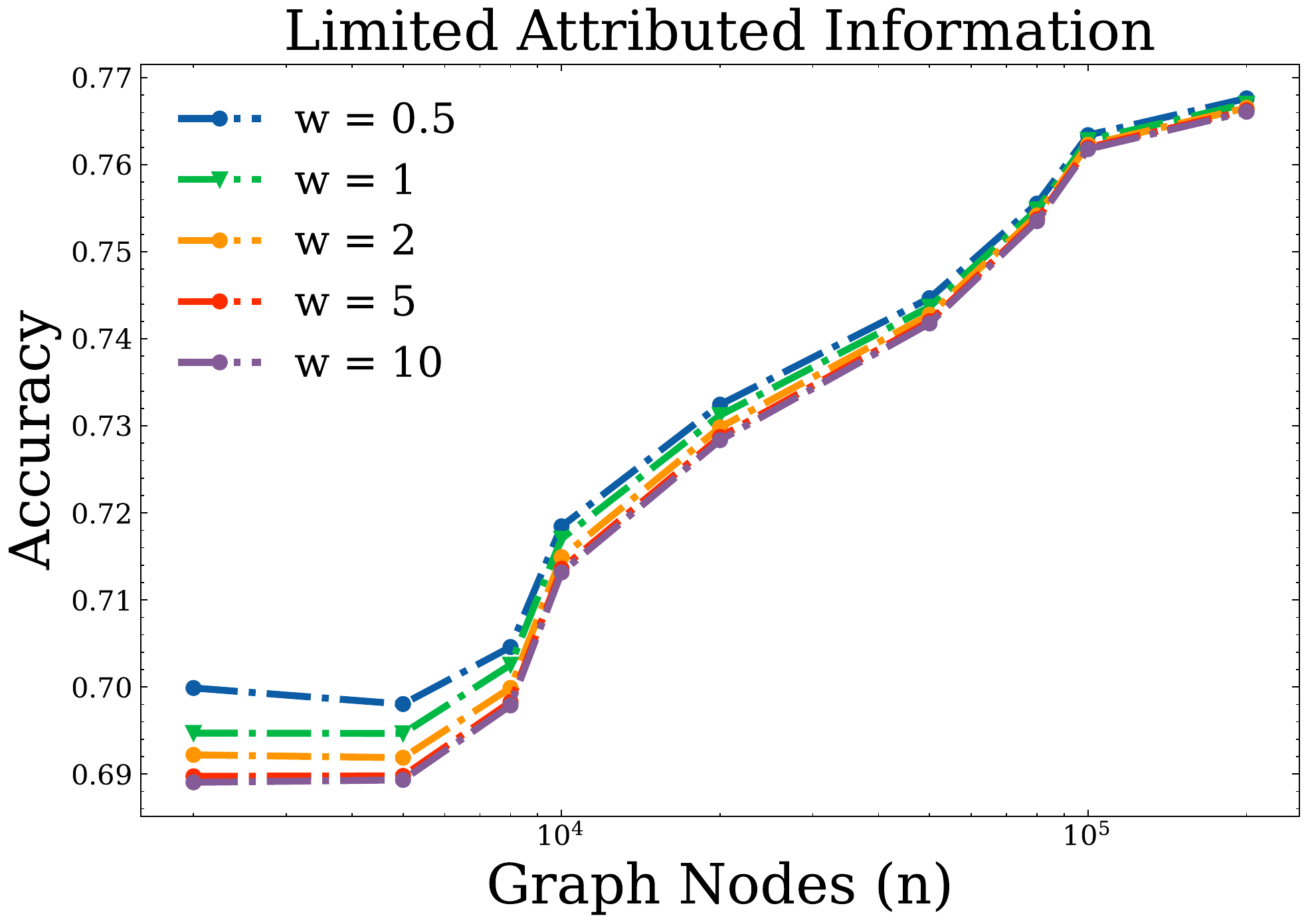}
\includegraphics[width=0.32\textwidth]{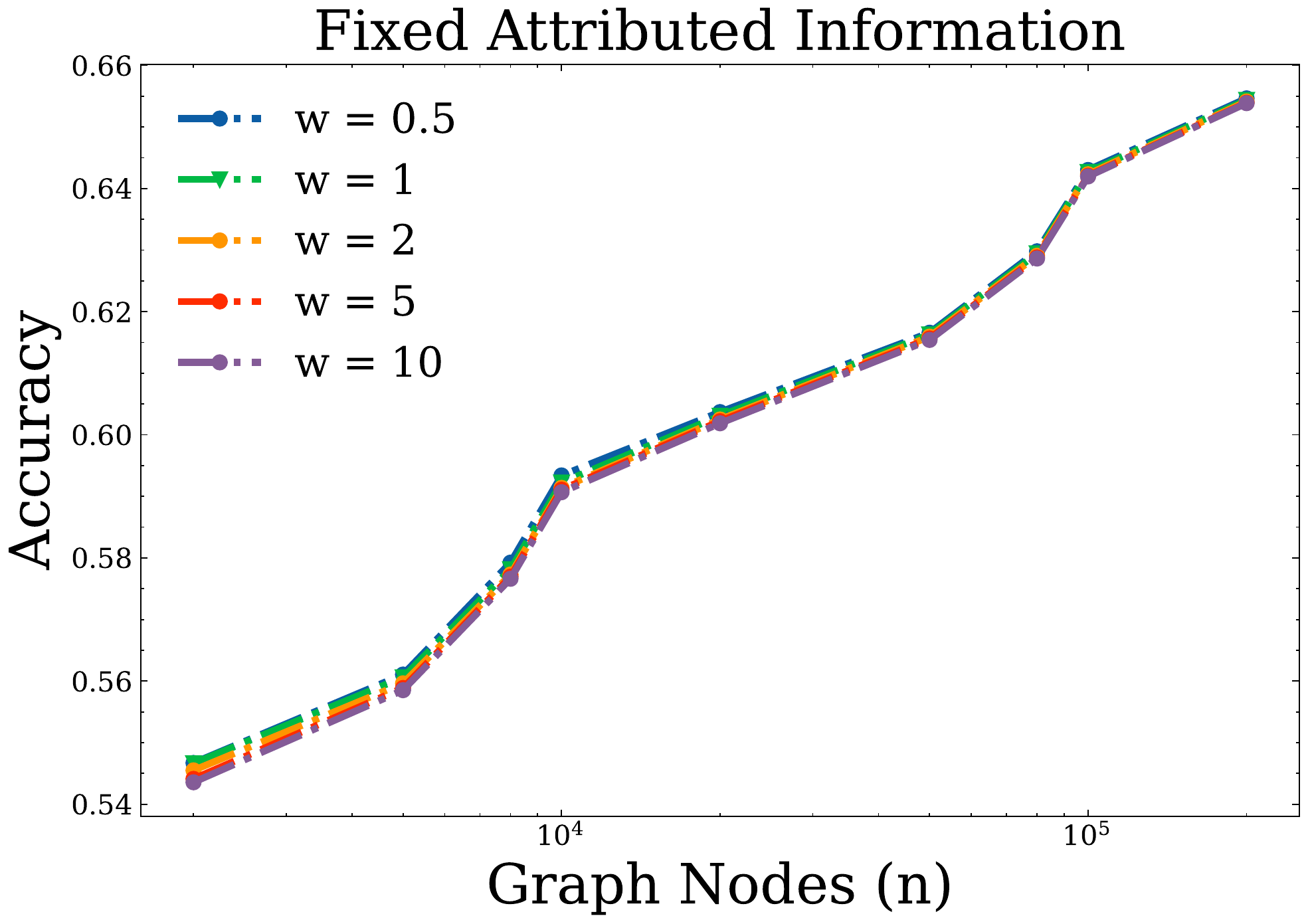}
\centering
\includegraphics[width=0.32\textwidth]{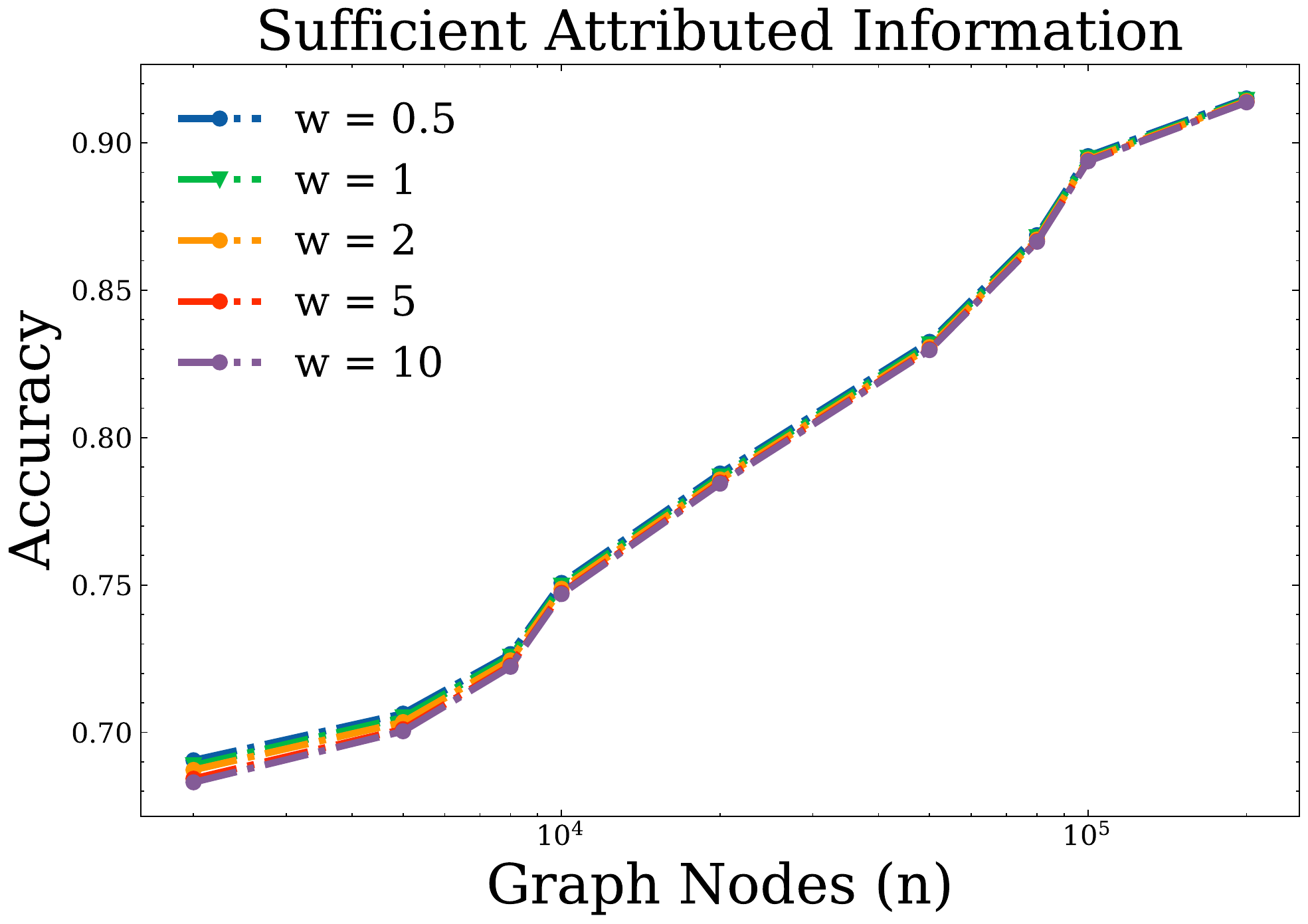}
\small{\caption{Classification Performance on the Linear Model $\mathcal{P}_v^l(w)$ with Different Weight Parameter. Structural Information: $p = 2\sqrt{n} / n, q = \sqrt{n} / n$. LEFT: Limited Attr. Info. $\|\mu - \nu\|_2 = 0.2 \log n / \sqrt{n}$; MIDDLE: Fixed Attr. Info. $\|\mu - \nu\|_2 = 0.05$; RIGHT: Sufficient Attr. Info. $\|\mu - \nu\|_2 = 0.01 \sqrt{\log n}$.}}
\end{figure}

In this experiment, we testify the choice of weight parameter $w$ in the linear model on CSBM-G with limited, fixed, and sufficient attributed information. The increase in the $w$ from $0.5$ to $10$. The results are presented in Fig~\ref{fig.w}. All three settings satisfy the separability condition given in Theorem~\ref{thm.1}. The accuracy will progressively increase with respect to n. Under three settings (Fig.~\ref{fig.w} LEFT, MIDDLE, RIGHT), the linear model with different weight parameter $w$'s behave almost the same (performance gap $<0.1\%$ for $n = 2\times10^5$). 

\subsubsection{Comparison between Heterophilic Graphs and Homophilic Graph}
\label{app:heter_to_homo}
\begin{figure}[t]
\centering
\includegraphics[height=0.25\textwidth]{fig/Classification_heter.pdf}
\includegraphics[height=0.25\textwidth]{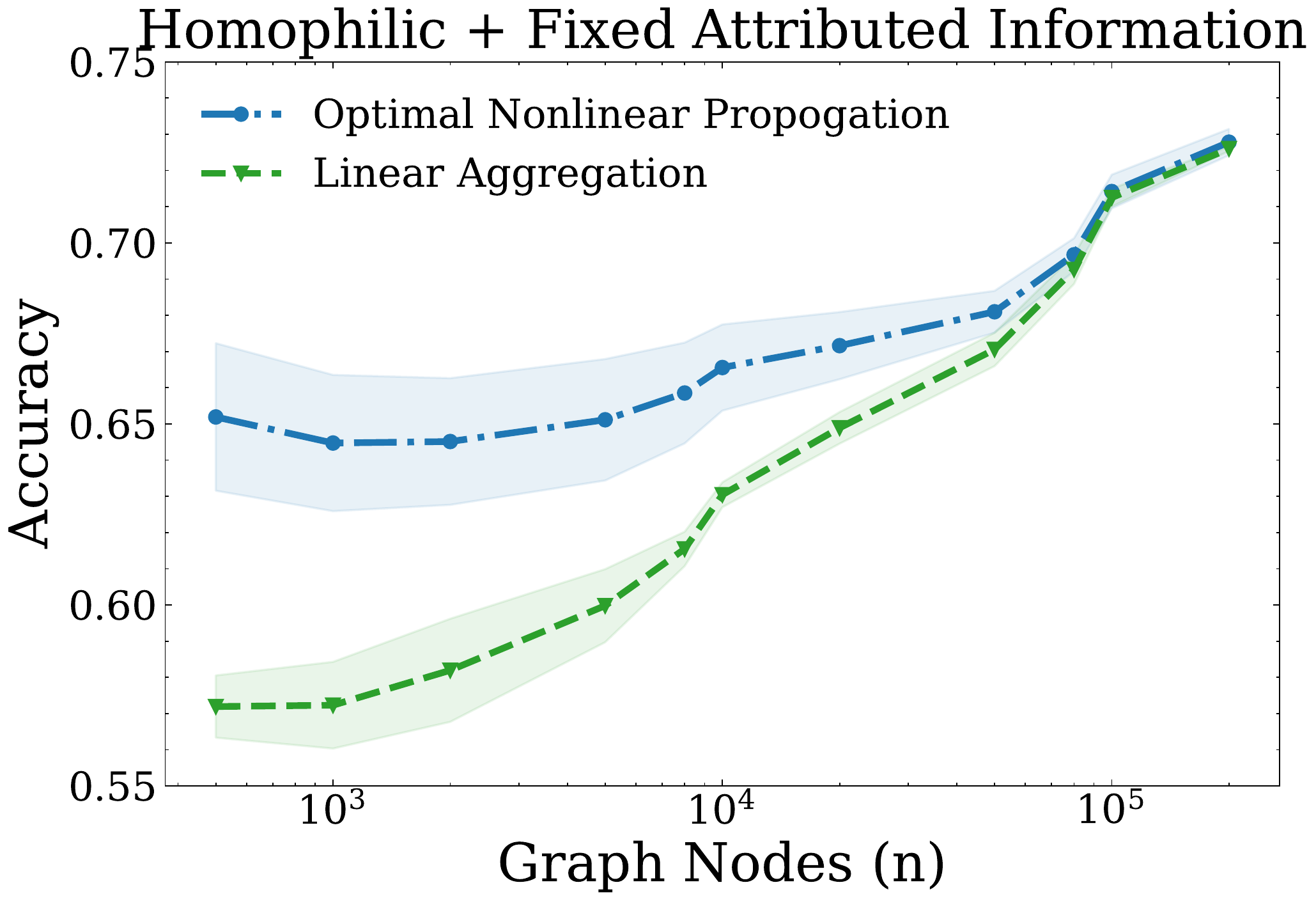}
\caption{Classification Performance on the Nonlinear Model v.s the Linear Model ($\mathcal{P}_v$ v.s. $\mathcal{P}_v^l$) with Heterophilic Graph Structure v.s. Homophilic Graph Structure. LEFT shows the heterophilic case and RIGHT shows the homophilic case when we switch the values of $p$, $q$. Both settings are with Gaussian node attributes. The parameter settings are: LEFT, $p=9/\sqrt{n},\,q=10/\sqrt{n},\,\|\mu-\nu\|_2=0.5,\,m=10$; RIGHT, $p=10/\sqrt{n},\,q=9/\sqrt{n},\,\|\mu-\nu\|_2=0.5,\,m=10$. }
\label{fig:compare-heter-homo}
\end{figure}

Here, we show the symmetric patterns between the cases with heterophilic graph structures and homophilic graphs structures by switching the structural parameters $p$ and $q$. Note that the nonlinear model $\mathcal{P}_v$ can naturally take care of the heterophilic case, because $\phi(x;\log p/q) = -\phi(x;\log q/p)$. For the linear model $\mathcal{P}_v^l(w)$, we use positive or negative $w$ to make the model fit the homophilic case or the heterophlic case, respectively. We follow the experiment setting in Sec.~\ref{sec:asmp} while testing different models on two CSBM-Gs by switching their parameters $p$ and $q$. The results are shown in Fig.~\ref{fig:compare-heter-homo}. The two curves are almost the same up-to some experimental randomness.
\subsubsection{Transition Curves under Laplacian Assumption}
\label{app:exp_trans_laplacian}
\begin{figure}[t]
\centering
\includegraphics[width=0.7\textwidth]{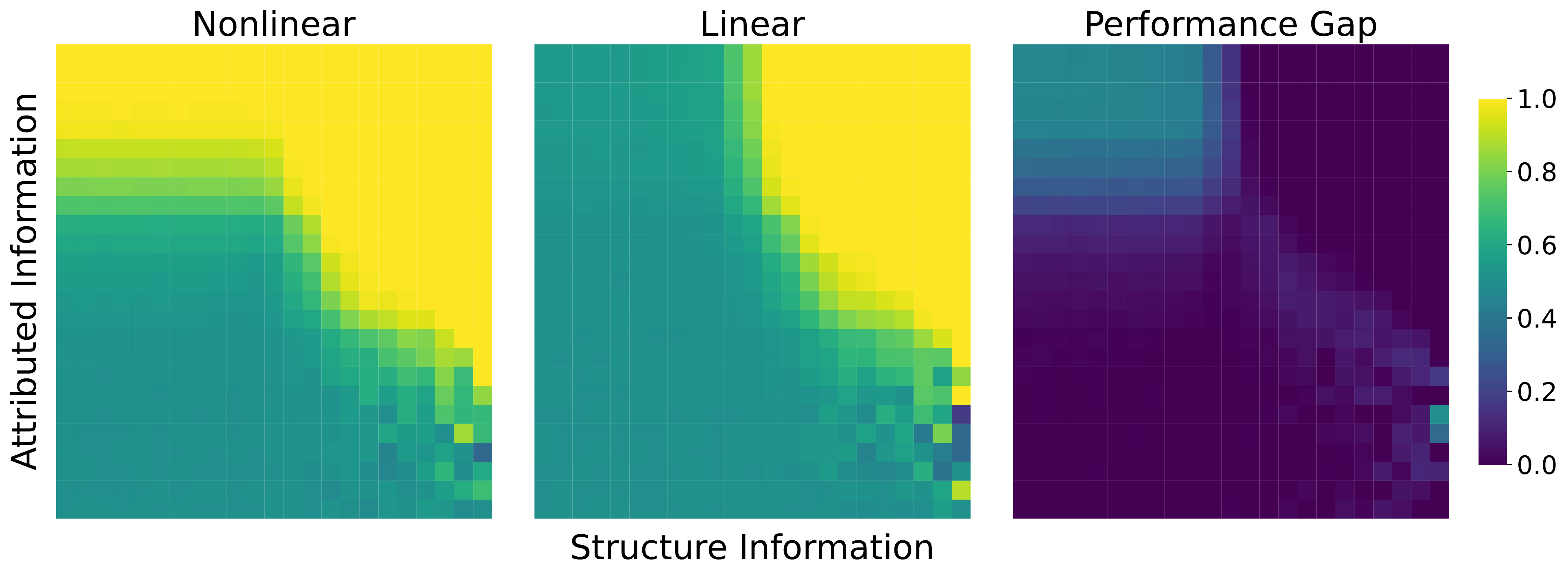}
\small{\caption{Transition Curves Attributed information ($\sqrt{m}\|\mu\|_2$) v.s. Structural Information ($|\log (p/q)|$) for CSBMs with Laplacian Node Attributions and Homophilic Graph Structures. The values in Performance Gap are obtained by the nonlinear case subtracting the linear case.
 }}
\label{fig.trans_lap}
\end{figure}

We present an additional experiment to explore the tradeoff between attributed information and structural information for CSBMs with Laplacian node attributes. Similar to the setting given in Section~\ref{subsec:trans}, the graph size is fixed to $n = 2\times10^4$, and we calculate the averaged result on 5 generated graphs. We test different levels of attributed information ($\|\mu\|_2$ from $10^{-4}$ to $10$, $m = 10$) and structural information (fixing $q = 5 \times 10^{-3}$ and increasing $p$ from $p = q$ to 1). The results are given in Fig~\ref{fig.trans_lap}, and similar to the case with Gaussian node attributes, when the attributed information is sufficient and the structural information is limited, the non-linear model outperforms the linear model. For other cases, two models perform similar. 

\subsubsection{Sparse Regime}
\begin{figure}[t]
\label{fig.sparse}
\centering
\includegraphics[width=0.32\textwidth]{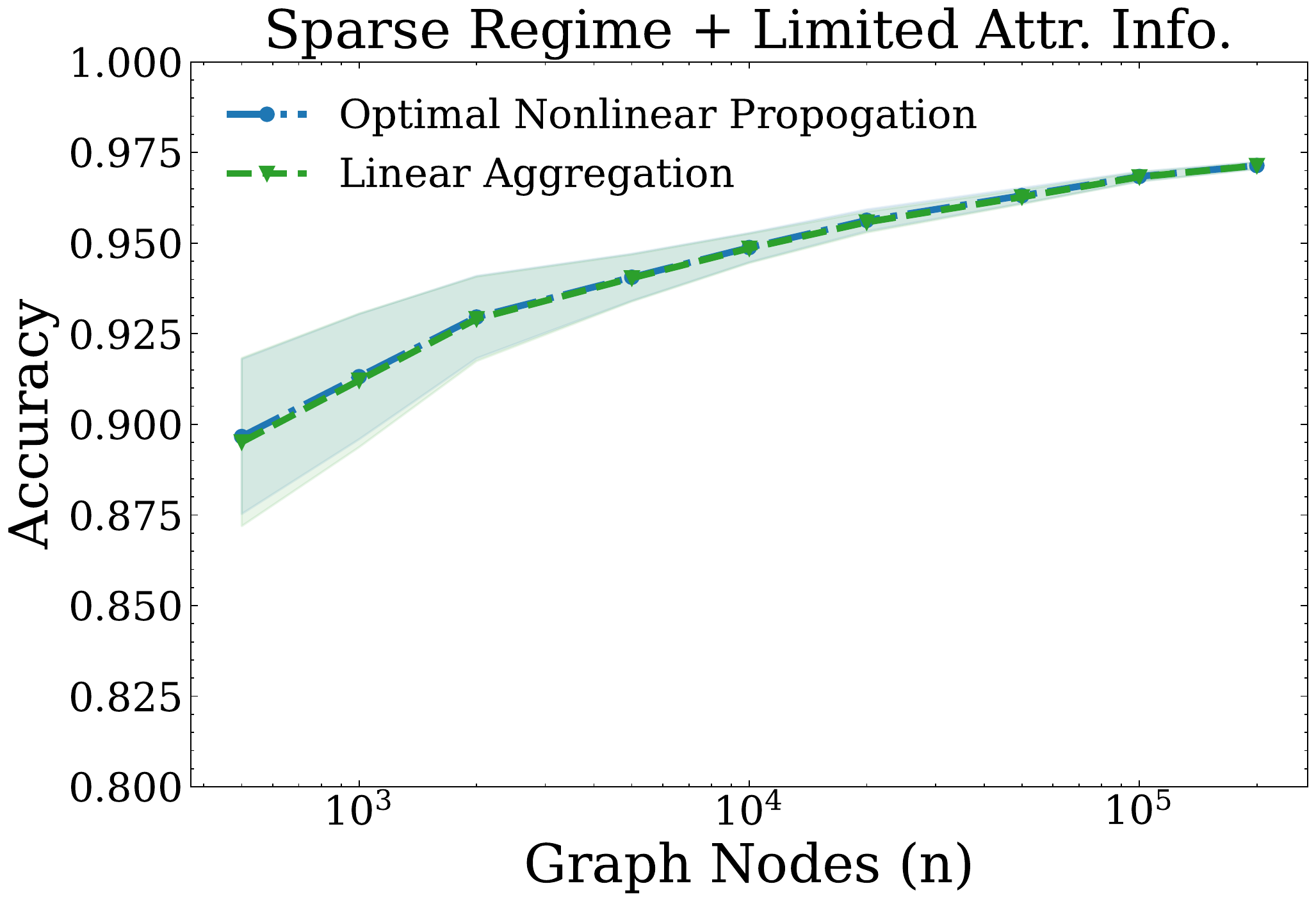}
\includegraphics[width=0.32\textwidth]{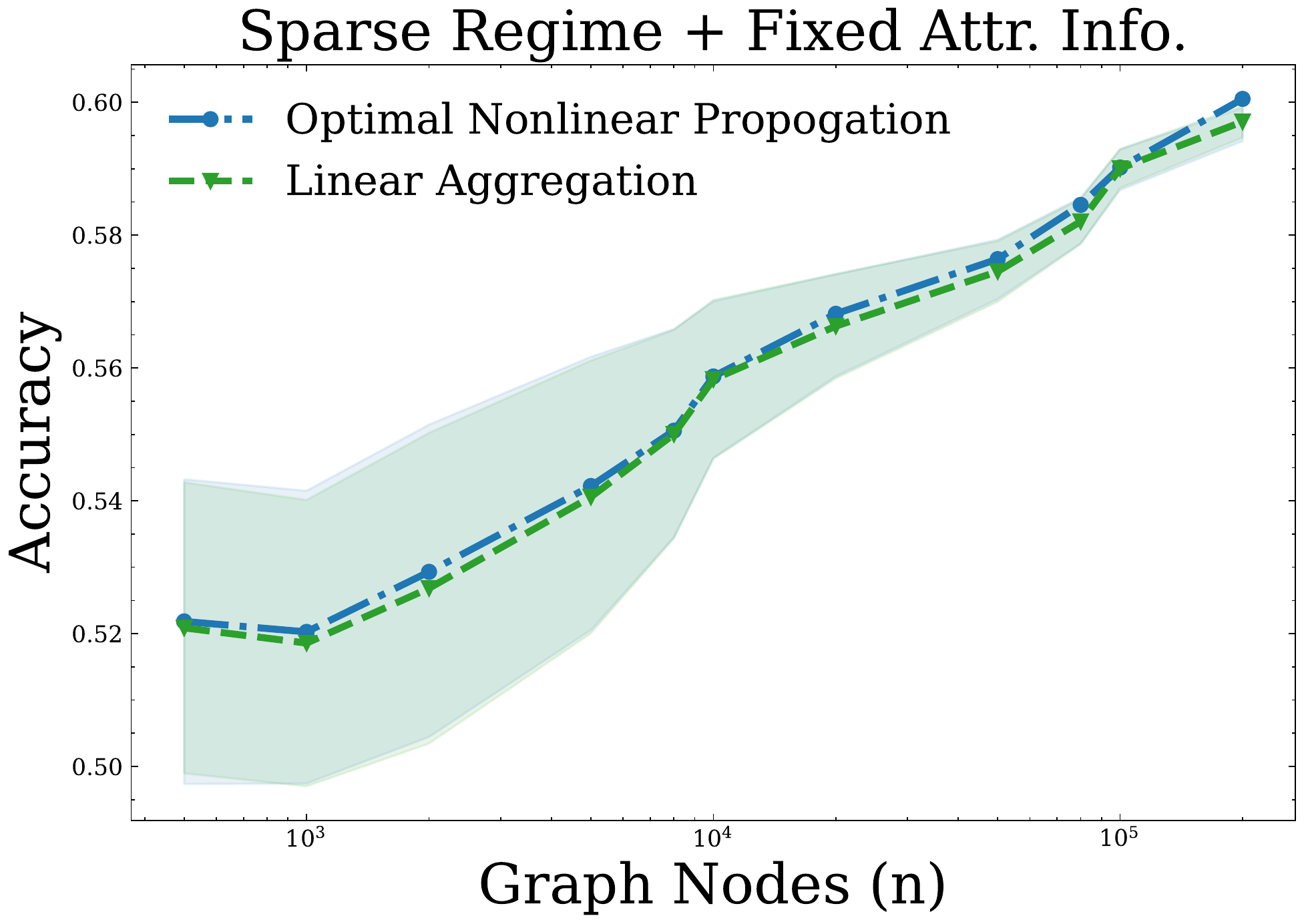}
\centering
\includegraphics[width=0.32\textwidth]{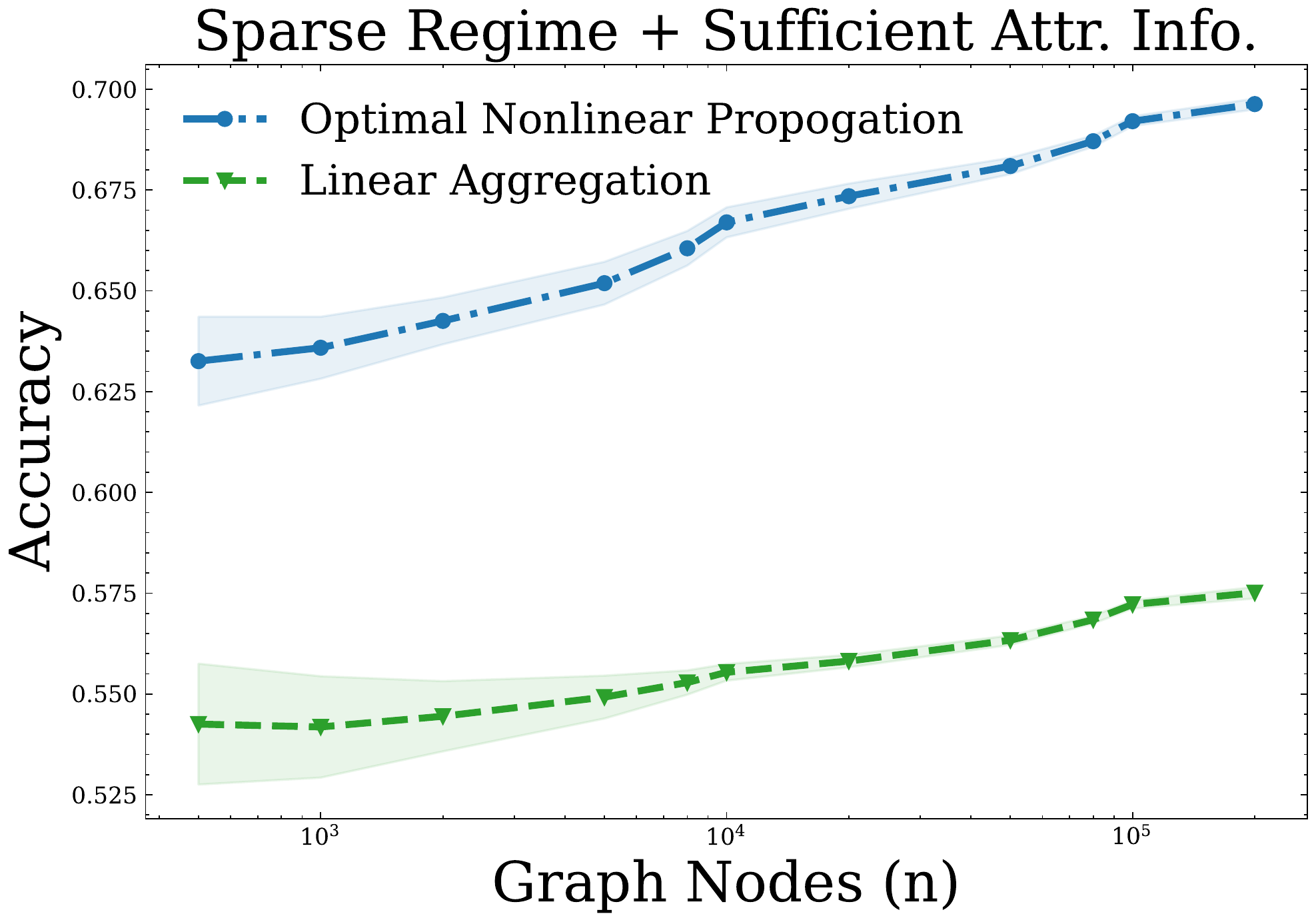}
\vspace{-2mm}
\small{\caption{Classification Performance on Nonlinear Models v.s. Linear Models ($\mathcal{P}_v$ v.s. $\mathcal{P}_v^l$) with Sparse Graph Structures. Structural Information: $p =  0.1 \log^4 n / n, q = 0.08\log^4 n / n$. LEFT: Limited Attr. Info. $\|\mu - \nu\|_2 = 0.03 \log^2 n / \sqrt{n}$; MIDDLE: Fixed Attr. Info. $\|\mu - \nu\|_2 = 0.03$; RIGHT: Sufficient Attr. Info. $\|\mu - \nu\|_2 = 0.03 \sqrt{\log n}$.}}
\end{figure}

The experimental results shown in Section~\ref{sec:exp} are on dense graphs, i.e. $p, q = \Theta_n(\sqrt{n} / n)$. In this section, we will consider more sparse graph that each node has approximately $(\log n)^4$ edges, which satisfies Assumption~\ref{as.1}. Similar to the settings presented in Section~\ref{sec:asmp}, the classification results under three settings are shown in Fig.~\ref{fig.sparse}. The results backup our Theorem~\ref{thm:2} that nonlinear model will outperform the linear model only when sufficient attributes are available.

\subsection{Real Data Experiments}
\label{app:exp_semi_synthetic}

\subsubsection{One-v.s.-all \& Several-v.s.-several Experimental Setup and Results}
\label{app:exp_OVA}
\textbf{Experimental Setup.} We provide further details for the one-v.s.-all and several-v.s.-several classification tasks. According to Table~\ref{tab:dataset}, there are 3, 7, 6 classification classes in PubMed, Cora, and CiteSeer datasets, respectively. As for the several-v.s.-several tasks, we group the original classes into two with relatively balanced node numbers in either group. The separation details are given in Table~\ref{tab:several}. Under the Gaussian attribute assumption, we generate attributes with dimension $m = 10$. In the training procedure, we initialize the projection vector ($\mu-\nu$ in the model) in $\psi_{\text{Gau}}$ with standard Gaussian, and make it trainable. We also initialize the threshold in $\phi$ as 0.2 and make it trainable. We train two functions using Adam optimizer with learning rate = 1e-2, weight decay = 5e-4 under the binary cross-entropy loss with 500 epochs. Under the Laplacian attribute assumption, we stills set $m = 10$. In the training procedure, $\psi_{\text{Lap}}, \phi$ are all nonlinear functions. We train the optimal projection vector with standard Gaussian initialization and threshold in $\psi_{\text{Lap}}$ (initialized as 0.2) and train the  threshold in $\phi$ with initial parameter 0.2. Further, we adopt Adam optimizer with learning rate = 1e-2, weight decay = 5e-4 under binary cross-entropy loss with 500 epochs. We compare the nonlinear propagation with models using only $\psi$, or $ \phi$, or the linear model.

\begin{table}
\centering
\caption{Several-v.s.several Experiments \label{tab:several}}
\begin{tabular}{l|l|l}
\toprule[1pt]
\textbf{Dataset} & \multicolumn{2}{c}{\textbf{Class Separation}}\\ \hline
\textbf{Cora} & Class 0, 1, 2 (\# node: 1804) & Class 3, 4, 5, 6 (\# node: 904)\\
\textbf{CiteSeer} & Class 0, 1, 2 (\# node: 1522) & Class 3, 4, 5 (\# node: 1805)\\
\textbf{PubMed} & Class 0, 1 (\# node: 11842) & Class 2 (\# node: 7875)\\
\bottomrule
\end{tabular}
\end{table}

\begin{figure}[h]
\centering
\includegraphics[width=0.32\textwidth]{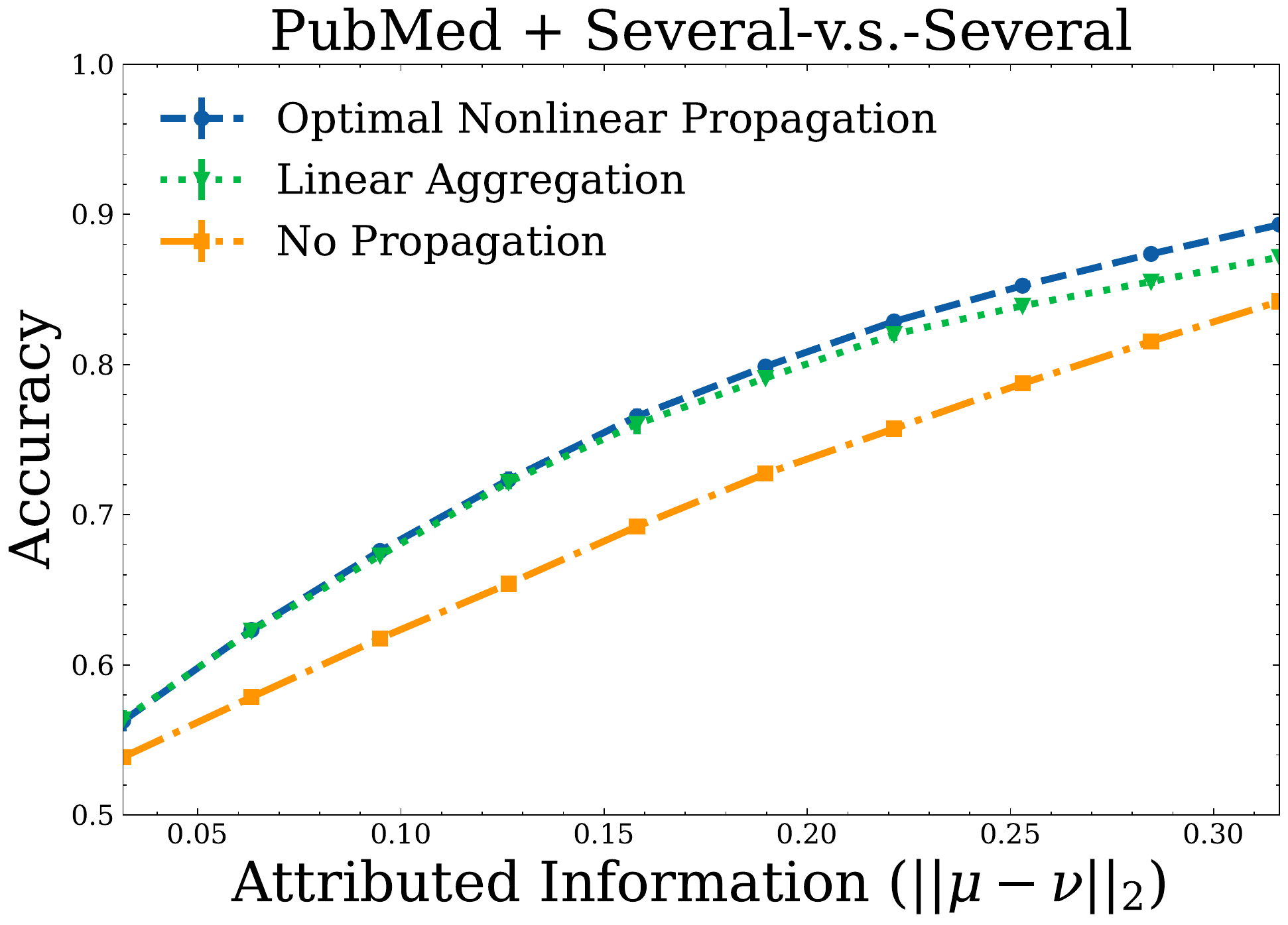}
\includegraphics[width=0.32\textwidth]{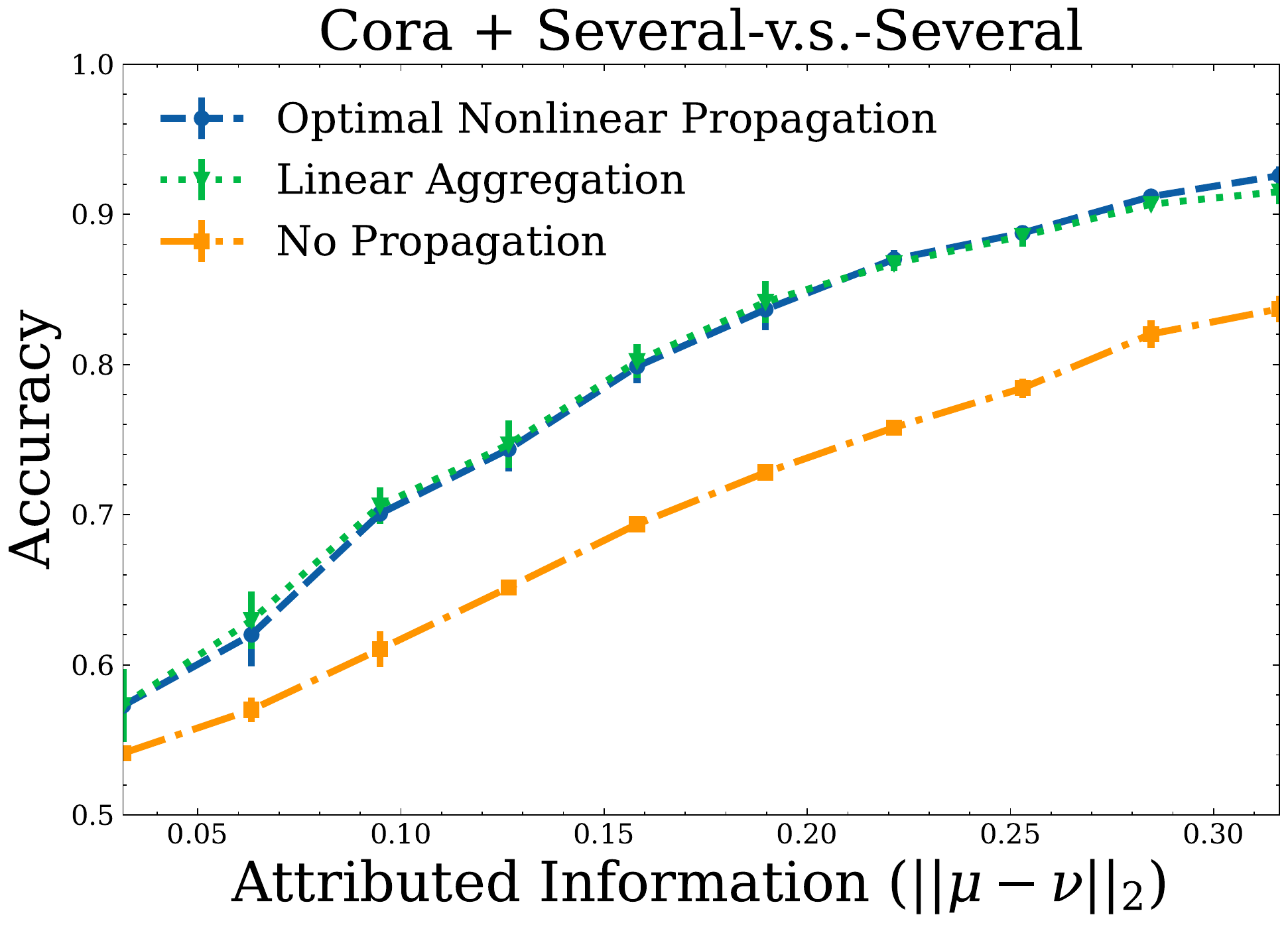}
\includegraphics[width=0.32\textwidth]{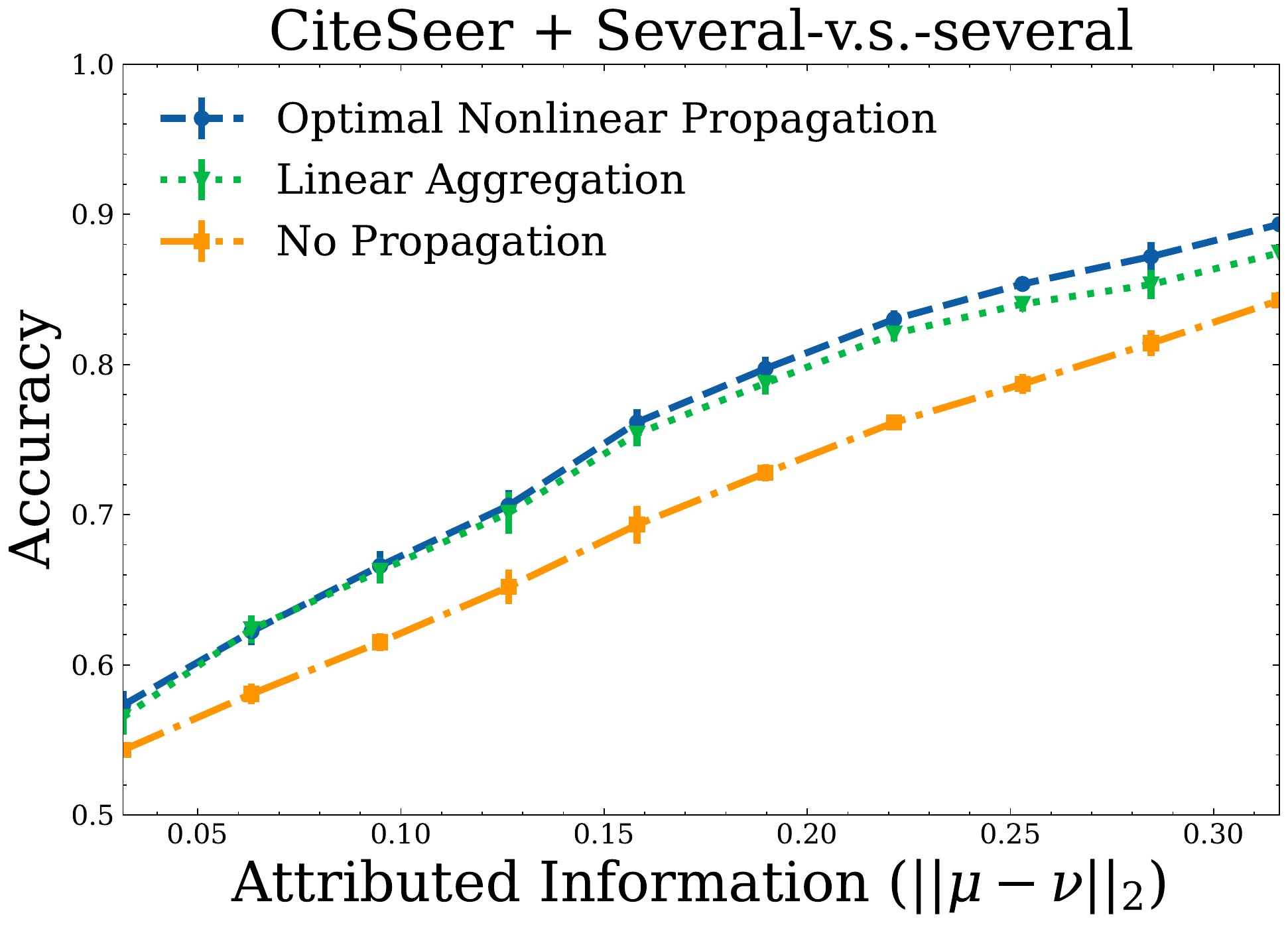}
\small{\caption{Averaged Several-vs-several Classification Performance on Citation Networks of Different Nonlinear Models v.s. Linear Models  ($\mathcal{P}_v$ v.s. $\mathcal{P}_v^l$). Node attributes in every class are generated from the Gaussian distributions with different means $\mu,\,\nu$. The optimal non-linear model has advantage over the linear model.}
\label{fig.several}}
\end{figure}

\textbf{Several-v.s.-Several Results.}
Fig~\ref{fig.several} shows the averaged performance for several-v.s.-several results over 5 trials under Gaussian node attributes. When sufficient attribute information is available, the nonlinear model outperforms the linear model.

\section{Additional Lemma Proofs}\label{app.additional}
\begin{lemmaI}
\label{lm.gaussian_density}
    Let $Z$ follow Gaussian distribution $\mathbb{P}_{Z} = \mathcal{N}(\mu, \sigma^2)$, we have the following:
    \begin{align}
        \mathbf{p}_Z'(z) = (-\frac{z - \mu}{\sigma^2})\mathbf{p}_Z(z)
    \end{align}
\end{lemmaI}
\begin{lemmaproofI}
    This above can be directly proved from definition
    \begin{align}
        \mathbf{p}_Z'(z) = \frac{1}{\sqrt{2\pi \sigma^2}}\exp(-\frac{(z - \mu)^2}{2\sigma^2})(-\frac{z - \mu}{\sigma^2}) = (-\frac{z - \mu}{\sigma^2})\mathbf{p}_Z(z)
    \end{align}
\end{lemmaproofI}

\begin{lemmaI}
    \label{lm.approx_Gaussian}
    Let $Z$ follow standard Gaussian distribution $\mathcal{N}(0, 1)$. We have the following asymptotic properties:
    \begin{align}
    \Phi(z) = 
        \left\{
            \begin{aligned}
                \frac{1}{\sqrt{2\pi}}\exp(-\frac{z^2}{2})(-\frac{1}{z} + o(\frac{1}{z})), \text{when } z \leq 0 \text{ \& } z \rightarrow -\infty \\
                1 - \frac{1}{\sqrt{2\pi}}\exp(-\frac{z^2}{2})(\frac{1}{z} + o(\frac{1}{z})), \text{when } z > 0 \text{ \& } z \rightarrow \infty \\
            \end{aligned}
        \right.
    \end{align}
\end{lemmaI}

\begin{lemmaproofI}
    For any $z > 0$,
    according to the integration by parts formula,
    \begin{align}
        \int_{z}^{+\infty} \exp(-\frac{s^2}{2}) ds
        =& (-\frac{1}{s})\exp(-\frac{s^2}{2})\bigg|_{z}^{+\infty} - \int_{z}^{+\infty} \frac{1}{s^2} \exp(-\frac{s^2}{2})ds \\
        =& (-\frac{1}{s})\exp(-\frac{s^2}{2})\bigg|_{z}^{+\infty} - (-\frac{1}{s^3}\exp(- \frac{s^2}{2})\bigg|_{z}^{+\infty} - \int_{z}^{+\infty}\frac{3}{s^4}\exp(-\frac{s^2}{2})) \\
        =& (-\frac{1}{s})\exp(-\frac{s^2}{2})\bigg|_{z}^{+\infty} + \frac{1}{s^3}\exp(- \frac{s^2}{2})\bigg|_{z}^{+\infty} - \frac{3}{s^5}\exp(- \frac{s^2}{2})\bigg|_{z}^{+\infty}  + \cdot \cdot \cdot
    \end{align}
    When $z \rightarrow +\infty$, we have
    \begin{align}
        \int_{z}^{+\infty} \exp(-\frac{s^2}{2}) ds = \exp(-\frac{z^2}{2})(\frac{1}{z} + o(\frac{1}{z}))
    \end{align}
    For $\Phi(z)$, when $z \rightarrow -\infty$, i.e., $-z \rightarrow \infty$, we have
    \begin{align}
        \Phi(z) =& \frac{1}{\sqrt{2\pi}}\int_{-\infty}^{z}\exp(-\frac{s^2}{2})ds = \frac{1}{\sqrt{2\pi}}\int_{-z}^{+\infty}\exp(-\frac{s^2}{2})ds \\
        =& \frac{1}{\sqrt{2\pi}}\exp(-\frac{z^2}{2})(-\frac{1}{z} + o(\frac{1}{z}))
    \end{align}
    When $z \rightarrow +\infty$, similarly, 
    \begin{align}
        \Phi(z) =& \frac{1}{\sqrt{2\pi}}\int_{-\infty}^{z}\exp(-\frac{s^2}{2})ds = 1 - \frac{1}{\sqrt{2\pi}}\int_{z}^{+\infty}\exp(-\frac{s^2}{2})ds \\
        =& 1 - \frac{1}{\sqrt{2\pi}} \exp(-\frac{z^2}{2})(\frac{1}{z} + o(\frac{1}{z}))
    \end{align}
\end{lemmaproofI}

    \textbf{Proof of Lemma~\ref{lm.main}}: Since $Z \stackrel{p}{\sim} \mathcal{N}(\mu, I / m)$, let $\tilde{Z} = \mu^TZ \stackrel{p}{\sim} \mathcal{N}(\mu^T\mu, \mu^T\mu / m)$ with density function $\mathbf{p}_{\tilde{Z}}(z)$.

    By the change of variables formula, 
    \begin{align}
        \mathbb{E}[\tilde{Z}] = & \int_{-\infty}^{-\frac{1}{2m}\log \frac{p}{q}}(-\log \frac{p}{q})\mathbf{p}_{\tilde{Z}}dz + \int_{-\frac{1}{2m}\log \frac{p}{q}}^{\frac{1}{2m}\log \frac{p}{q}}2mz\mathbf{p}_{\tilde{Z}}dz + \int_{\frac{1}{2m}\log \frac{p}{q}}^{+\infty}(\log \frac{p}{q})\mathbf{p}_{\tilde{Z}}dz \\
        =& (-\log \frac{p}{q})\Phi(-\frac{1}{2m}\log\frac{p}{q}; \mu^T\mu, \frac{\mu^T\mu}{m}) + \log \frac{p}{q} [1 - \Phi(\frac{1}{2m}\log\frac{p}{q}; \mu^T\mu, \frac{\mu^T\mu}{m})] \nonumber \\
        &+ 2m \int_{-\frac{1}{2m}\log\frac{p}{q}}^{\frac{1}{2m}\log\frac{p}{q}}z\mathbf{p}_{\tilde{Z}}dz
    \end{align}
    Now, consider $\int_{-\frac{1}{2m}\log\frac{p}{q}}^{\frac{1}{2m}\log\frac{p}{q}}z\mathbf{p}_{\tilde{Z}}dz$, let $t = \sqrt{\frac{m}{\mu^T\mu}}(z - \mu^T\mu)$.
    \begin{align}
        \int_{-\frac{1}{2m}\log\frac{p}{q}}^{\frac{1}{2m}\log\frac{p}{q}}z\mathbf{p}_{\tilde{Z}}dz =& \int_{-\frac{1}{2m}\log \frac{p}{q}}^{\frac{1}{2m}\log \frac{p}{q}}z\sqrt{\frac{m}{2\pi\mu^T\mu}}\exp(-\frac{m(z - \mu^T\mu)^2}{2\mu^T\mu})dz \\ 
        =& \sqrt{\frac{m}{2\pi\mu^T\mu}} \int_{N}^{M}(\sqrt{\frac{\mu^T\mu}{m}}t + \mu^T\mu)\exp(-\frac{t^2}{2})\sqrt{\frac{\mu^T\mu}{m}}dt \\
        =& \sqrt{\frac{\mu^T\mu}{2\pi m}}\int_{N}^{M}t\exp(-\frac{t^2}{2})dt + \frac{\mu^T\mu}{\sqrt{2\pi}}\int_{N}^{M}\exp(-\frac{t^2}{2})dt \\
        =& \sqrt{\frac{\mu^T\mu}{2\pi m}}(\exp(-\frac{N^2}{2}) - \exp(-\frac{M^2}{2})) + \mu^T\mu(\Phi(M) - \Phi(N))
    \end{align}
    where 
    \begin{align}
        &M = \sqrt{\frac{m}{\mu^T\mu}}(\frac{1}{2m}\log \frac{p}{q} - \mu^T\mu) = -\frac{1}{2\sqrt{m\mu^T\mu}}\log \frac{p}{q} - \sqrt{m\mu^T\mu} \\
        &N = \sqrt{\frac{m}{\mu^T\mu}}(-\frac{1}{2m}\log \frac{p}{q} - \mu^T\mu) = \frac{1}{2\sqrt{m\mu^T\mu}}\log \frac{p}{q} - \sqrt{m\mu^T\mu}
    \end{align}
    Since $\sqrt{m / (\mu^T\mu)}(Z - \mu^T\mu) \sim \mathcal{N}(0, 1)$, therefore
    \begin{align}
        &\Phi(-\frac{1}{2m}\log\frac{p}{q}; \mu^T\mu, \frac{\mu^T\mu}{m}) = \Phi(\sqrt{\frac{m}{\mu^T\mu}}(-\frac{1}{2m}\log \frac{p}{q} - \mu^T\mu)) = \Phi(N) \\
        &\Phi(\frac{1}{2m}\log\frac{p}{q}; \mu^T\mu, \frac{\mu^T\mu}{m}) = \Phi(\sqrt{\frac{m}{\mu^T\mu}}(\frac{1}{2m}\log \frac{p}{q} - \mu^T\mu)) = \Phi(M)
    \end{align}
    Hence, 
    \begin{align}
    \label{eq.2}
        \mathbb{E}[\tilde{Z}] =& (\log \frac{p}{q})\cdot[1 - \Phi(M) - \Phi(N)] + 2m\mu^T\mu[\Phi(M) - \Phi(N)] \nonumber \\
        &+ \sqrt{\frac{2m\mu^T\mu}{\pi}}[\exp(-\frac{N^2}{2}) - \exp(-\frac{M^2}{2})]
    \end{align}
    We further calculate the second order moment of $\tilde{Z}$:
    \begin{align}
        \mathbb{E}[\tilde{Z}^2]= & \int_{-\infty}^{-\frac{1}{2m}\log \frac{p}{q}}(- \log \frac{p}{q})^2\mathbf{p}_{\tilde{Z}}dz + \int_{\frac{1}{2m}\log \frac{p}{q}}^{+\infty}(\log \frac{p}{q})^2\mathbf{p}_{\tilde{Z}}dz + \int_{-\frac{1}{2m}(\log \frac{p}{q})}^{\frac{1}{2m}\log \frac{p}{q}} (2mz)^2\mathbf{p}_{\tilde{Z}}dz \\
        =& (\log \frac{p}{q})^2 [1 + \Phi(N) - \Phi(M)] +  4m^2\int_{-\frac{1}{2m}\log \frac{p}{q}}^{\frac{1}{2m}\log \frac{p}{q}} z^2\mathbf{p}_{\tilde{Z}}dz
    \end{align}
    To make the formula looks clean, let $\sigma^2 = \frac{\mu^T\mu}{m}$, $\mu' = \mu^T\mu$, and $a = \frac{1}{2m}\log \frac{p}{q}$. By applying Lemma~\ref{lm.gaussian_density}, we have
    \begin{align}
        \int_{-\frac{1}{2m}\log \frac{p}{q}}^{\frac{1}{2m}\log \frac{p}{q}} = &\int_{-a}^{a} z^2\mathbf{p}_{\tilde{Z}}dz \\
        =& (-\sigma^2)\int_{-a}^{a}z(-\frac{z - \mu'}{\sigma^2})\mathbf{p}_{\tilde{Z}}dz + \mu'\int_{-a}^{a}z\mathbf{p}_{\tilde{Z}}dz \\
        =& (-\sigma^2)\int_{-a}^{a}zf_z'(z)dz + (-\mu'\sigma^2)\int_{-a}^{a}f_z'(z)dz + \mu'^2 \int_{-a}^{a}\mathbf{p}_{\tilde{Z}}dz \\
        =& -\sigma^2[af_z(a) + af_z(-a) - \Phi_Z(a) + \Phi_Z(-a)] \nonumber \\&-\mu\sigma^2[f_z(a) - f_z(-a)] + \mu'^2[\Phi_Z(a) - \Phi_Z(-a)] \\
        =& (\mu'^2 + \sigma^2)[\Phi_Z(a) - \Phi_Z(-a)] - \mu'\sigma^2[f_z(a) - f_z(-a)] - a\sigma^2[f_z(a) + f_z(-a)] \\
        =& [(\mu^T\mu)^2 + \frac{\mu^T\mu}{m}][\Phi(M) - \Phi(N)] - \mu^T\mu \cdot \frac{\mu^T\mu}{m}[f_{z}(a) - f_z(-a)] \nonumber \\
        &-\frac{1}{2d}\log \frac{p}{q} \cdot \frac{\mu^T\mu}{m} \cdot [f_z(a) + f_z(-a)]
    \end{align}
    By combing all the terms together, we get 
    \begin{align}
        \mathbb{E}[\tilde{Z}^2] = & (\log \frac{p}{q})^2 + (\log \frac{p}{q})^2[\Phi(N) - \Phi(M)] + 4m^2[(\mu^T\mu)^2 + \frac{\mu^T\mu}{m}][\Phi(M) - \Phi(N)] \nonumber \\
        &-4m\mu^T\mu \cdot \mu^T\mu \cdot [f_z(a) - f_z(-a)] - \frac{1}{2m}\log \frac{p}{q} \cdot \frac{\mu^T\mu}{m} \cdot 4m^2 \cdot [f(a) + f(-a)] \\
        = & (\log \frac{p}{q})^2 + [4(m\mu^T\mu)^2 + 4m\mu^T\mu - (\log \frac{p}{q})^2][\Phi(M) - \Phi(N)] \nonumber \\
        &-\frac{4m\mu^T\mu \cdot \mu^T\mu}{\sqrt{2\pi} \cdot \sqrt{\frac{\mu^T\mu}{m}}} \cdot [\exp(- \frac{M^2}{2}) - \exp(- \frac{N^2}{2})] \nonumber \\ &- \frac{2\log \frac{p}{q} \cdot \mu^T\mu}{\sqrt{2\pi}\cdot \sqrt{\frac{\mu^T\mu}{m}}}[\exp(-\frac{M^2}{2}) + \exp(-\frac{N^2}{2})] \\
        =& (\log \frac{p}{q})^2 + [4(m\mu^T\mu)^2 + 4m\mu^T\mu - (\log \frac{p}{q})^2][\Phi(M) - \Phi(N)] \nonumber \\
        &-\frac{4m\mu^T\mu \cdot \sqrt{m\mu^T\mu}}{\sqrt{2\pi}} \cdot [\exp(- \frac{M^2}{2}) - \exp(- \frac{N^2}{2})] \nonumber \\&- \frac{2\log \frac{p}{q} \cdot \sqrt{m\mu^T\mu}}{\sqrt{2\pi}}[\exp(-\frac{M^2}{2}) + \exp(-\frac{N^2}{2})] \label{eq.3}
    \end{align}
    Now, we will use Equation~\ref{eq.2} and Equation~\ref{eq.3} for further discussion.
   By Lemma~\ref{lm.approx_Gaussian}, 
    
    \textbf{Case I: When $\sqrt{m\mu^T\mu} = o_n(\log \frac{p}{q})$},
    \begin{align}
        \Phi(N) =& \Phi(-\frac{1}{2\sqrt{m\mu^T\mu}}\log \frac{p}{q} - \sqrt{m\mu^T\mu}) \\
        \sim & \frac{1}{\sqrt{2\pi}} \exp(-\frac{N^2}{2})[\frac{1}{\frac{1}{2\sqrt{m\mu^T\mu}}\log \frac{p}{q} + \sqrt{m\mu^T\mu}} + o(\frac{1}{N})] \\
        \sim & \frac{1}{\sqrt{2\pi}} \exp(-\frac{N^2}{2}) [\frac{2\sqrt{m\mu^T\mu}}{\log \frac{p}{q}} + o(\frac{\sqrt{m\mu^T\mu}}{\log \frac{p}{q}})]
    \end{align}
    \begin{align}
        \Phi(M) =& \Phi(\frac{1}{2\sqrt{m\mu^T\mu}}\log \frac{p}{q} - \sqrt{m\mu^T\mu}) \\
        \sim & 1 - \frac{1}{\sqrt{2\pi}} \exp(-\frac{N^2}{2})[\frac{1}{\frac{1}{2\sqrt{m\mu^T\mu}}\log \frac{p}{q} - \sqrt{m\mu^T\mu}} + o(\frac{1}{M})] \\
        \sim & 1 - \frac{1}{\sqrt{2\pi}} \exp(-\frac{M^2}{2}) [\frac{2\sqrt{m\mu^T\mu}}{\log \frac{p}{q}} + o(\frac{\sqrt{m\mu^T\mu}}{\log \frac{p}{q}})]
    \end{align}
    \begin{align}
        \exp(-\frac{N^2}{2}) =& \exp(-\frac{(\log \frac{p}{q})^2}{8m\mu^T\mu}) \cdot \exp(-\frac{1}{2}\log \frac{p}{q}) \cdot \exp(-\frac{1}{2}m\mu^T\mu) \\
        =& \exp(-\frac{(\log \frac{p}{q})^2}{8m\mu^T\mu}) \cdot \exp(-\frac{1}{2}m\mu^T\mu) \cdot (1 - \frac{1}{2}\log \frac{p}{q} + o_n(\log \frac{p}{q}))
    \end{align}
    \begin{align}
        \exp(-\frac{M^2}{2}) =& \exp(-\frac{(\log \frac{p}{q})^2}{8m\mu^T\mu}) \cdot \exp(\frac{1}{2}\log \frac{p}{q}) \cdot \exp(-\frac{1}{2}m\mu^T\mu) \\
        =& \exp(-\frac{(\log \frac{p}{q})^2}{8m\mu^T\mu}) \cdot \exp(-\frac{1}{2}m\mu^T\mu) \cdot (1 + \frac{1}{2}\log \frac{p}{q} + o_n(\log \frac{p}{q}))
    \end{align}
    From above, we have
    \begin{align}
        &\Phi(M) - \Phi(N) \sim 1 - \frac{2\sqrt{2}}{\sqrt{\pi}} \cdot \frac{\sqrt{m\mu^T\mu}}{\log \frac{p}{q}} \exp(-\frac{(\log \frac{p}{q})^2}{8m\mu^T\mu}) \sim 1 \\
        &1 - \Phi(M) - \Phi(N) \sim \sqrt{\frac{2}{\pi}} \cdot \frac{\sqrt{m\mu^T\mu}}{\log \frac{p}{q}} \cdot \log \frac{p}{q} \cdot \exp(-\frac{(\log \frac{p}{q})^2}{8m\mu^T\mu}) \sim o_n(1) \\
        &\exp(-\frac{N^2}{2}) - \exp(-\frac{M^2}{2})\sim -\log \frac{p}{q} \cdot \exp(-\frac{(\log \frac{p}{q})^2}{8m\mu^T\mu})
    \end{align}
    
    In this case,
    \begin{align}
        \mathbb{E}[\tilde{Z}] =& \log \frac{p}{q}[1 - \Phi(M) - \Phi(N)] + 2m\mu^T\mu[\Phi(M) - \Phi(N)] \nonumber \\
        &+ \sqrt{\frac{2m\mu^T\mu}{\pi}}[\exp(-\frac{N^2}{2}) - \exp(-\frac{M^2}{2})] \\
        \sim & 2m\mu^T\mu
    \end{align}
    \begin{align}
        \mathbb{E}[\tilde{Z}^2] =& (\log \frac{p}{q})^2 + [4(m\mu^T\mu)^2 + 4m\mu^T\mu - (\log \frac{p}{q})^2][\Phi(M) - \Phi(N)] \nonumber \\
        &-\frac{4m\mu^T\mu \cdot \sqrt{m\mu^T\mu}}{\sqrt{2\pi}} \cdot [\exp(- \frac{M^2}{2}) - \exp(- \frac{N^2}{2})] \nonumber \\
        &- \frac{2\log \frac{p}{q} \cdot \sqrt{m\mu^T\mu}}{\sqrt{2\pi}}[\exp(-\frac{M^2}{2}) + \exp(-\frac{N^2}{2})] \\
        \sim & (\log \frac{p}{q})^2 + [4(m\mu^T\mu)^2 + 4m\mu^T\mu - (\log \frac{p}{q})^2][1 - \frac{2\sqrt{2}}{\sqrt{\pi}}\cdot \frac{m\mu^T\mu}{\log \frac{p}{q}}\exp(-\frac{(\log \frac{p}{q})^2}{8m\mu^T\mu})] \nonumber \\
        &- \frac{4m\mu^T\mu \cdot \sqrt{m\mu^T\mu}}{\sqrt{2\pi}} \cdot \log \frac{p}{q} \cdot \exp(-\frac{(\log \frac{p}{q})^2}{8m\mu^T\mu}) - \frac{4\log \frac{p}{q} \cdot \sqrt{m\mu^T\mu}}{\sqrt{2\pi}} \cdot \exp(-\frac{(\log \frac{p}{q})^2}{8m\mu^T\mu}) \\
        \sim & 4(m\mu^T\mu)^2 + 4m\mu^T\mu
    \end{align}
    \begin{align}
        \text{var}(\tilde{Z}) = \mathbb{E}[\tilde{Z}^2] - (\mathbb{E}[\tilde{Z}])^2 \sim 4m\mu^T\mu.
    \end{align}
    \textbf{Case II: When $\sqrt{m\mu^T\mu} = \Theta (\log \frac{p}{q})$,} We define $\lambda$ as:
    \begin{align}
        \lambda = \frac{\log \frac{p}{q}}{\sqrt{m\mu^T\mu}}
    \end{align}
    where $\lambda \in [C_1, C_2]$. By definition, we know
    \begin{align}
        &\Phi(M) = \Phi(\frac{\lambda}{2} - \sqrt{m\mu^T\mu}) = \Phi(\frac{\lambda}{2}(1 + o_n(1))) \\
        &\Phi(N) = \Phi(-\frac{\lambda}{2} - \sqrt{m\mu^T\mu}) = \Phi(-\frac{\lambda}{2}(1 + o_n(1)))\\
        &\exp(-\frac{N^2}{2}) = \exp(-\frac{1}{8}\lambda^2) \cdot \exp(-\frac{1}{2}\log \frac{p}{q}) \cdot \exp(-\frac{1}{2}m\mu^T\mu)\\
        &\exp(-\frac{M^2}{2}) = \exp(-\frac{1}{8}\lambda^2) \cdot \exp(\frac{1}{2}\log \frac{p}{q}) \cdot \exp(-\frac{1}{2}m\mu^T\mu)
    \end{align}
    Now, we go further for detailed analysis: On the one hand,
    \begin{align}
        1 - \Phi(M) - \Phi(N) =& 1 - \Phi(\frac{\lambda}{2} - \sqrt{m\mu^T\mu}) - \Phi(-\frac{\lambda}{2} - \sqrt{m\mu^T\mu}) \\
        =& \Phi(-\frac{\lambda}{2} + \sqrt{m\mu^T\mu}) - \Phi(-\frac{\lambda}{2} - \sqrt{m\mu^T\mu}) \\
        =& \frac{1}{\sqrt{2\pi}} \int_{-\frac{\lambda}{2}-\sqrt{m\mu^T\mu}}^{-\frac{1}{2}\lambda - \sqrt{m\mu^T\mu}} \exp(-\frac{t^2}{2})dt
    \end{align}
    $\forall$ small $\epsilon > 0$, $\sqrt{m\mu^T\mu} < \epsilon$ satisfies when $n$ is large enough.
    \begin{align}
        1 - \Phi(M) - \Phi(N) &\geq \frac{1}{\sqrt{2\pi}}\exp(-\frac{(1/2C_2 + \epsilon)^2}{2}) \cdot 2\sqrt{m\mu^T\mu} \\
        & = \sqrt{\frac{2}{\pi}} \sqrt{m\mu^T\mu} \exp(-\frac{(1/2C_2 + \epsilon)^2}{2})
    \end{align}
    On the other hand,
    \begin{align}
        1 - \Phi(M) - \Phi(N) \leq \sqrt{\frac{2}{\pi}} \sqrt{m\mu^T\mu} \exp(-\frac{(1/2C_1 - \epsilon)^2}{2})
    \end{align}
    Hence,
    \begin{align}
        1 - \Phi(M) - \Phi(N) = K_1 \sqrt{m\mu^T\mu}
    \end{align}
    where $K_1 \in [\sqrt{\frac{2}{\pi}}\exp(-\frac{(1/2C_2 + \epsilon)^2}{2}), \sqrt{\frac{2}{\pi}}\exp(-\frac{(1/2C_1 - \epsilon)^2}{2})]$.
    
    Similarly, we have the upper and lower bound for $\Phi(M) - \Phi(N)$:
    \begin{align}
        \frac{\lambda}{\sqrt{2\pi}} \exp(-\frac{(1/2C_2 + \epsilon)^2}{2}) \leq \Phi(M) - \Phi(N) \leq \frac{\lambda}{\sqrt{2\pi}}
    \end{align}
    Hence,
    \begin{align}
        \Phi(M) - \Phi(N) = K_2 \lambda
    \end{align}
    where $K_2 \in [\frac{1}{\sqrt{2\pi}}\exp(-\frac{(1/2C_2 + \epsilon)^2}{2}), \frac{1}{\sqrt{2\pi}}]$.
    Besides, we have
    \begin{align}
        \exp(-\frac{N^2}{2}) - \exp(-\frac{M^2}{2}) \sim -\log \frac{p}{q}\exp(-\frac{1}{8}\lambda^2)
    \end{align}
    Therefore,
    \begin{align}
        \mathbb{E}[\tilde{Z}] \sim \log \frac{p}{q} \sqrt{m\mu^T\mu} (K_1 + 2K_2 - \sqrt{\frac{2}{\pi}}\exp(-\frac{1}{8}\lambda^2))
    \end{align}
    From the proof, it is not hard to see $K_1 \sim \sqrt{\frac{2}{\pi}}\exp(-\frac{1}{8}\lambda^2)$, and further shows
    \begin{align}
        \mathbb{E}[\tilde{Z}] = \Theta (\sqrt{m\mu^T\mu} \cdot \log \frac{p}{q})
    \end{align}
    Applying the above analysis on $\mathbb{E}[\tilde{Z}^2]$ using Equation~\ref{eq.3}, and we get the variance:
    \begin{align}
        \text{var}[\tilde{Z}] = \mathbb{E}[\tilde{Z}]^2 - (\mathbb{E}[\tilde{Z}])^2 = \Theta((\log \frac{p}{q})^2)
    \end{align}
    \textbf{Case III: When $\sqrt{m\mu^T\mu} = \omega_{n}(\log \frac{p}{q})$, $\sqrt{m\mu^T\mu} = o_n(1)$,}
    \begin{align}
        &\Phi(N) = \Phi(-\frac{\log \frac{p}{q}}{2\sqrt{m\mu^T\mu}}(1 + o_n(1)))\\
        &\Phi(M) = \Phi(\frac{\log \frac{p}{q}}{2\sqrt{m\mu^T\mu}}(1 + o_n(1)))
    \end{align}
    From above, we have 
    \begin{align}
        \Phi(M) + \Phi(N) =& \Phi(\frac{\log \frac{p}{q}}{2\sqrt{m\mu^T\mu}} - \sqrt{m\mu^T\mu}) + \Phi(-\frac{\log \frac{p}{q}}{2\sqrt{m\mu^T\mu}} - \sqrt{m\mu^T\mu}) \\
        =& \Phi(\frac{\log \frac{p}{q}}{2\sqrt{m\mu^T\mu}} - \sqrt{m\mu^T\mu}) + 1 - \Phi(\frac{\log \frac{p}{q}}{2\sqrt{m\mu^T\mu}} + \sqrt{m\mu^T\mu}) \\
        =& 1 - [\Phi(\frac{\log \frac{p}{q}}{2\sqrt{m\mu^T\mu}} + \sqrt{m\mu^T\mu}) - \Phi(\frac{\log \frac{p}{q}}{2\sqrt{m\mu^T\mu}} - \sqrt{m\mu^T\mu})] \\
        \sim & 1 - \frac{1}{\sqrt{2\pi}}\cdot 2 \cdot \sqrt{m\mu^T\mu} = 1 - \sqrt{\frac{2}{\pi}} \cdot \sqrt{m\mu^T\mu}
    \end{align}
    \begin{align}
        \Phi(M) - \Phi(N)=& \Phi(\frac{\log \frac{p}{q}}{2\sqrt{m\mu^T\mu}} - \sqrt{m\mu^T\mu}) - \Phi(-\frac{\log \frac{p}{q}}{2\sqrt{m\mu^T\mu}} - \sqrt{m\mu^T\mu}) \\
        \sim& \frac{1}{\sqrt{2\pi}} \cdot \frac{\log \frac{p}{q}}{\sqrt{m\mu^T\mu}}
    \end{align}
    Besides, we can show the difference between the exponential term
    \begin{align}
        &\exp(-\frac{N^2}{2}) - \exp(-\frac{M^2}{2}) \sim -\log \frac{p}{q} \\
        &\exp(-\frac{N^2}{2}) + \exp(-\frac{M^2}{2}) \sim 2
    \end{align}
    In this case,
    \begin{align}
        \mathbb{E}[\tilde{Z}] =& \log \frac{p}{q}[1 - \Phi(M) - \Phi(N)] + 2m\mu^T\mu[\Phi(M) - \Phi(N)] \nonumber \\
        &+ \sqrt{\frac{2m\mu^T\mu}{\pi}}[\exp(-\frac{N^2}{2}) - \exp(-\frac{M^2}{2})] \\
        \sim & \sqrt{\frac{2}{\pi}} \cdot \sqrt{m\mu^T\mu} \cdot \log \frac{p}{q}
    \end{align}
    \begin{align}
        \mathbb{E}[\tilde{Z}]^2 = & (\log \frac{p}{q})^2 + [4(m\mu^T\mu)^2 + 4m\mu^T\mu - (\log \frac{p}{q})^2][\Phi(M) - \Phi(N)] \nonumber \\
        &-\frac{4m\mu^T\mu \cdot \sqrt{m\mu^T\mu}}{\sqrt{2\pi}} \cdot [\exp(- \frac{M^2}{2})- \exp(- \frac{N^2}{2})] \nonumber \\
        & - \frac{2\log \frac{p}{q} \cdot \sqrt{m\mu^T\mu}}{\sqrt{2\pi}}[\exp(-\frac{M^2}{2}) + \exp(-\frac{N^2}{2})] \\
        \sim & (\log \frac{p}{q})^2 - \frac{1}{\sqrt{2\pi}} \cdot \frac{\log \frac{p}{q}}{\sqrt{m\mu^T\mu}} \cdot (\log \frac{p}{q})^2 \sim (\log \frac{p}{q})^2
    \end{align} 
    \begin{align}
        \text{var}[\tilde{Z}] = \mathbb{E}[\tilde{Z}]^2 - (\mathbb{E}[\tilde{Z}])^2 \sim (\log \frac{p}{q})^2
    \end{align}
    \textbf{Case IV: When $m\mu^T\mu = \Theta(1)$} (suppose $m\mu^T\mu \in [\tilde{C}_1, \tilde{C}_2]$), for any small $\epsilon>0$, we have
    \begin{align}
        \sqrt{\frac{2}{\pi}} \cdot \sqrt{m\mu^T\mu} \cdot \exp(-\frac{(\tilde{C_2} + \epsilon)^2}{2}) \leq 1 - \Phi(M) - \Phi(N) \leq \sqrt{\frac{2}{\pi}} \cdot \sqrt{m\mu^T\mu}
    \end{align}
    \begin{align}
        \frac{1}{\sqrt{2\pi}} \exp(-\frac{(\tilde{C_2} + \epsilon)^2}{2}) \lambda \leq \Phi(M) - \Phi(N) \leq \frac{1}{\sqrt{2\pi}} \exp(-\frac{(\tilde{C_1} - \epsilon)^2}{2}) \lambda
    \end{align}
    \begin{align}
        \exp(-\frac{N^2}{2}) - \exp(-\frac{M^2}{2}) \sim -\log \frac{p}{q} \cdot \exp(-\frac{1}{2}m\mu^T\mu)
    \end{align}
    In this case, since $m\mu^T\mu = \Theta_n(1)$, we have
    \begin{align}
        &\mathbb{E}[\tilde{Z}] = \Theta_n(\log\frac{p}{q}) \\
        &\text{var}[\tilde{Z}] = \text{var}[\tilde{Z}] = \mathbb{E}[\tilde{Z}]^2 - (\mathbb{E}[\tilde{Z}])^2 = \Theta_n((\log \frac{p}{q})^2)
    \end{align}
     \textbf{Case IV: When $m\mu^T\mu = \omega_n(1)$},
    \begin{align}
        &N \sim -\sqrt{m\mu^T\mu} \rightarrow - \infty \\
        &M \sim -\sqrt{m\mu^T\mu} \rightarrow - \infty
    \end{align}
    Similar to the previous proof, we have
    \begin{align}
        &\Phi(M) + \Phi(N) \sim \frac{1}{\sqrt{2\pi}} \cdot \exp(-\frac{1}{2}m\mu^T\mu) \cdot \frac{2}{\sqrt{m\mu^T\mu}}\\
        &\Phi(M) - \Phi(N) \sim \frac{1}{\sqrt{2\pi}} \cdot \exp(-\frac{1}{2}m\mu^T\mu) \cdot \frac{\log \frac{p}{q}}{\sqrt{m\mu^T\mu}}
    \end{align}
    \begin{align}
        &\exp(-\frac{N^2}{2}) - \exp(-\frac{M^2}{2})
        \sim - \log \frac{p}{q} \cdot \exp(-\frac{1}{2}m\mu^T\mu) \\
        &\exp(-\frac{N^2}{2}) + \exp(-\frac{M^2}{2}) \sim 2 \exp(-\frac{1}{2}m\mu^T\mu)
    \end{align}
    In this case,
    \begin{align}
        \mathbb{E}[\tilde{Z}] =& \log \frac{p}{q}[1 - \Phi(M) - \Phi(N)] + 2m\mu^T\mu[\Phi(M) - \Phi(N)] \nonumber \\
        +& \sqrt{\frac{2m\mu^T\mu}{\pi}}[\exp(-\frac{N^2}{2}) - \exp(-\frac{M^2}{2})] \\
        =& \log \frac{p}{q} - \sqrt{\frac{2}{\pi}} \cdot \exp(-\frac{1}{2}m\mu^T\mu) \cdot \frac{\log \frac{p}{q}}{\sqrt{m\mu^T\mu}}\sim \log \frac{p}{q}
    \end{align}
    \begin{align}
        \mathbb{E}[\tilde{Z}^2] = & (\log \frac{p}{q})^2 + [4(m\mu^T\mu)^2 + 4m\mu^T\mu - (\log \frac{p}{q})^2][\Phi(M) - \Phi(N)] \nonumber \\
        &-\frac{4m\mu^T\mu \cdot \sqrt{m\mu^T\mu}}{\sqrt{2\pi}} \cdot [\exp(- \frac{M^2}{2}) - \exp(- \frac{N^2}{2})] \nonumber \\
        &- \frac{2\log \frac{p}{q} \cdot \sqrt{m\mu^T\mu}}{\sqrt{2\pi}}[\exp(-\frac{M^2}{2}) + \exp(-\frac{N^2}{2})] \\
        \sim& (\log \frac{p}{q})^2 - \frac{1}{\sqrt{2\pi}} \cdot \frac{(\log \frac{p}{q})^3}{\sqrt{m\mu^T\mu}} \cdot \exp(-\frac{1}{2}m\mu^T\mu)
    \end{align}
    Hence,
    \begin{align}
     \text{var}[\tilde{Z}] = \mathbb{E}[\tilde{Z}]^2 - (\mathbb{E}[\tilde{Z}])^2 \sim \sqrt{\frac{8}{\pi}} \cdot \exp(-\frac{1}{2}m\mu^T\mu) \cdot \frac{(\log \frac{p}{q})^2}{\sqrt{m\mu^T\mu}}
    \end{align}

\end{document}